\documentclass[11pt, letterpaper]{article}

\usepackage[left=1in, right=1in, top=1in, bottom=1in]{geometry}
\usepackage[utf8]{inputenc}
\usepackage[T1]{fontenc}
\usepackage{bm}
\usepackage{type1cm}
\usepackage{lettrine}
\usepackage{amsmath,amssymb,amsthm}
\usepackage{moreverb}
\usepackage{mathtools}
\usepackage{amsmath}
\usepackage{amssymb}
\usepackage{algorithmic}
\usepackage{graphics}
\usepackage{graphicx}
\usepackage{subfigure}
\usepackage{caption}
\usepackage{enumitem}
\usepackage{extarrows}
\usepackage{color}
\usepackage{framed}
\usepackage{wrapfig}
\usepackage{bm}
\usepackage{mathrsfs}
\usepackage{mathabx}
\usepackage{multirow}
\usepackage{longtable}
\usepackage{hyperref}
\usepackage{paralist}
\usepackage{indentfirst}
\usepackage{relsize}
\usepackage{extarrows}
\usepackage{lineno,xcolor}
\usepackage{upgreek}
\usepackage{bm}
\usepackage{mwe}
\usepackage{xcolor}
\usepackage{booktabs}
\usepackage{authblk}
\usepackage{lettrine}
\usepackage{type1cm}
\usepackage{threeparttable}
\usepackage[sort&compress,numbers]{natbib}
\usepackage[figurename=Fig.]{caption}
\usepackage[normalem]{ulem}

\usepackage[font=footnotesize,labelfont=bf]{caption}
\newcommand{\eref}[1]{(\ref{#1})}
\providecommand{\keywords}[1]{\textbf{\textit{Keywords: }} #1}

\hypersetup{
bookmarks=true,
bookmarksopen=true,
bookmarksnumbered=true,
unicode=false,
pdftoolbar=true,
pdfmenubar=true,
pdffitwindow=false,
pdfstartview={FitH},
pdftitle={My title},
pdfauthor={Author},
pdfsubject={Subject},
pdfcreator={Creator},
pdfproducer={Producer},
pdfkeywords={keywords},
pdfnewwindow=true,
colorlinks=true,
linkcolor=blue,
citecolor=blue,
filecolor=magenta,
urlcolor=blue
}

\begin{document}

\title{\textbf{Physics-informed learning of governing equations from scarce data}}

\author[1]{Zhao Chen}
\author[2,*]{Yang Liu}
\author[1,3,$\ddag$]{Hao Sun}

\affil[1]{\small Department of Civil and Environmental Engineering, Northeastern University, Boston, MA 02115, USA}
\affil[2]{Department of Mechanical and Industrial Engineering, Northeastern University, Boston, MA 02115, USA}
\affil[3]{Department of Civil and Environmental Engineering, MIT, Cambridge, MA 02139, USA \vspace{12pt}} 
\affil[*]{Corresponding author. E-mail: yang1.liu@northeastern.edu}
\affil[$\ddag$]{Corresponding author. E-mail: h.sun@northeastern.edu \vspace{12pt}}
\affil[$\bowtie$]{Author contributions: Y.L. and H.S. contributed to the ideation and design of the research; Z.C. and H.S. performed the research; Z.C., Y.L. and H.S wrote the paper.}

\date{}

\maketitle

\normalsize

\vspace{-28pt} 
\begin{abstract}
	\small
    Harnessing data to discover the underlying governing laws or equations that describe the behavior of complex physical systems can significantly advance our modeling, simulation and understanding of such systems in various science and engineering disciplines. This work introduces a novel physics-informed deep learning framework to discover governing partial differential equations (PDEs) from scarce and noisy data for nonlinear spatiotemporal systems. In particular, this approach seamlessly integrates the strengths of deep neural networks for rich representation learning, physics embedding, automatic differentiation and sparse regression to (1) approximate the solution of system variables, (2) compute essential derivatives, as well as (3) identify the key derivative terms and parameters that form the structure and explicit expression of the PDEs. The efficacy and robustness of this method are demonstrated, both numerically and experimentally, on discovering a variety of PDE systems with different levels of data scarcity and noise accounting for different initial/boundary conditions. The resulting computational framework shows the potential for closed-form model discovery in practical applications where large and accurate datasets are intractable to capture.
\end{abstract}


\vspace{24pt} 
\lettrine[lines=2, loversize=0.1, findent=3pt, nindent=0pt]{\textbf{C}}{}urrent practices on modeling of complex dynamical systems have been mostly rooted in the use of ordinary and/or partial differential equations (ODEs, PDEs) that govern the system behaviors. These governing equations are conventionally obtained from rigorous first principles such as the conservation laws or knowledge-based phenomenological derivations. However, there remain many real-world complex systems underexplored, whose analytical descriptions are undiscovered and parsimonious closed forms of governing equations are unclear or partially unknown. Luckily, observational datasets become increasingly rich and offer an alternative of distilling the underlying equations from data. Harnessing data to uncover the governing laws or equations can significantly advance and transform our modeling, simulation and understanding of complex physical systems in various science and engineering disciplines. For example, obtaining mathematical equations that govern the evolution of sea ice from observational data (e.g., satellite remote sensing images) brings distinct benefits for better understanding and predicting the growth, melt and movement of the Arctic ice pack. Distilling an explicit formulation from field sensing data (e.g., Doppler radar recordings) will accelerate more accurate prediction of weather and climate patterns. Recently, advances in machine learning theories, computational capacity and data availability kindle significant enthusiasm and efforts towards data-driven discovery of physical laws and governing equations \cite{Bongard9943, Schmidt2009, Schaeffer6634, Daniels2015, Brunton3932, Rudye1602614, Schaeffer2017, lusch2018deep, WANG201944, Champion22445, Pfister25405, HUANG2020JMPS}.

Pioneering contributions by Bongard and Lipson \cite{Bongard9943} and Schmidt and Lipson \cite{Schmidt2009} leveraged stratified symbolic regression and genetic programming to successfully distil the underlying differential equations that govern nonlinear system dynamics from data. However, this elegant approach doesn't scale up well with the dimensionality of the system, is computationally expensive, and might suffer from overfitting issues. Recently, an impressive breakthrough made by Brunton \emph{et al.} \cite{Brunton3932} leads to an innovative sparsity-promoting approach called sparse identification of nonlinear dynamics (SINDy), which selects dominant candidate functions from a high-dimensional nonlinear function space based on sparse regression to uncover parsimonious governing equations, ODEs in particular. The sparsity was achieved by a sequential threshold ridge regression (STRidge) algorithm which recursively determines the sparse solution subjected to hard thresholds \cite{Brunton3932, Rudye1602614}. Such an approach is capable of balancing the complexity and accuracy of identified models and thus results in parsimony. SINDy has drawn tremendous attention in the past few years, leading to variant algorithms with applications to identify projected low-dimensional surrogate models in the form of first-order ODEs, alternatively with linear embedding \cite{lusch2018deep, Champion22445}, for a wide range of nonlinear dynamical systems, such as fluid flows \cite{loiseau2018constrained, loiseau2018sparse}, structural systems \cite{LAI2019813, Li-PRE2220}, biological and chemical systems \cite{Mangan2016, hoffmann2019reactive, BHADRIRAJU2019372}, active matter \cite{cichos2020machine}, predictive control of nonlinear dynamics \cite{Kaiser2018}, multi-time-scale systems \cite{Champion2019SIAM}, a predator-prey system \cite{Dam2017}, and stochastic processes \cite{Boninsegna2018}, just naming a few among many others. There are also a number of other extensions of SINDy that discover implicit dynamics \cite{Mangan2016, Kaheman2020Implicit}, incorporate physics constraints \cite{loiseau2018constrained}, and embed random sampling to improve the robustness to noise for sparse discovery of high-dimensional dynamics \cite{schaeffer2018extracting}. The convergence and error estimate analyses \cite{zhang2019convergence} theoretically sustain the family of SINDy approaches. 

The sparsity-promoting paradigm has been later extended for data-driven discovery of spatiotemporal systems governed by PDEs, e.g., the PDE-FIND algorithm \cite{Rudye1602614, Schaeffer2017}, where the library of candidate functions is augmented by incorporating spatial partial derivative terms. This method has been further investigated or improved to, for example, obtain parametric PDEs from data \cite{Rudy2019SIAM}, discover PDEs enhanced by Bayesian inference \cite{ZhangPRS2018} and gene expression programming \cite{Vaddireddy2020}, identify diffusion and Navier-Stokes equations based on molecular simulation \cite{zhang_ma_2020}, and learn PDEs for biological transport models \cite{lagergren2020learning}. Nevertheless, a critical bottleneck of the SINDy framework, especially for data-driven discovery of PDEs, lies in its strong dependence on both quality and quantity of the measurement data, since numerical differentiation is required to compute the derivatives in order to construct governing equation(s). Especially, the use of finite difference or filtering to calculate derivatives leads to a pivotal challenge that reduces the algorithm robustness. This specially limits the applicability of SINDy in its present form to scenarios given highly incomplete, scarce and noisy data. It is notable that variational system identification \cite{WANG201944} shows satisfactory robustness of calculating derivatives based on isogeometric analysis for discovering the weak form of PDEs. However, such an approach doesn't scale down well with respect to the fidelity of available data. Another work \cite{Gurevich2019} shows that weak formulation can significantly improve the discovery robustness against noise, but requires careful design of test functions, which is intractable for high-dimensional spatiotemporal systems.

Graph-based automatic differentiation \cite{baydin2017automatic} is well posed to address the above issue, which has been proven successful in deep learning for solving nonlinear PDEs \cite{Han8505, SIRIGNANO20181339, RAISSI2019686, YANG2019136, Bar-Sinai15344, SUN2020112732, Raissi1026}. In particular, the deep neural network (DNN) is used to approximate the solution constrained by both the PDE(s) and available data. Latest studies \cite{BERG2019239, Both2019} show the potential of using DNNs and automatic differentiation to obtain PDEs from noisy data; yet, false positive identification occurs due to the use of less rigorous sparse regression along with DNN training. Simultaneously optimizing the DNN parameters and sparse PDE coefficients poses a significant challenge in finding the global optimum. In this work, we present a novel Physics-informed Deep Learning (PiDL) framework, possessing salient features of interpretability and generalizability, to discover governing PDEs of nonlinear spatiotemporal systems from scarce and noisy data accounting for different initial/boundary conditions. Our methodology integrates the strengths of DNNs for rich representation learning, automatic differentiation for accurate derivative calculation as well as $\ell_0$ sparse regression to tackle the fundamental limitation of existing methods that scale poorly with data noise and scarcity. The efficacy and robustness of our method are demonstrated on a variety of PDE systems, both numerically and experimentally.

\section*{RESULTS}
\subsection*{PiDL with Sparse Regression for PDE Discovery}
We consider a multi-dimensional spatiotemporal system whose governing equations can be described by a set of nonlinear, coupled, parameterized PDEs in the general form given by
\begin{equation}\label{eq:PDE}
    \mathbf{u}_t + \mathcal{F} \big[\mathbf{u}, \mathbf{u}^2, ..., \nabla_\mathbf{x}\mathbf{u}, \nabla_\mathbf{x}^2\mathbf{u}, \nabla_\mathbf{x}\mathbf{u}\cdot\mathbf{u}, ...; \boldsymbol{\lambda} \big] = \mathbf{p}
\end{equation}
where $\mathbf{u} = \mathbf{u}(\mathbf{x}, t) \in\mathbb{R}^{1\times n}$ is the multi-dimensional latent solution (dimension $=n$) while $\mathbf{u}_t$ is the first-order time derivative term; $t\in [0, T]$ denotes time and $\mathbf{x}\in\Omega$ specifies the space; $\mathcal{F}[\cdot]$ is a complex nonlinear functional of $\mathbf{u}$ and its spatial derivatives, parameterized by $\boldsymbol{\lambda}$; $\nabla$ is the gradient operator with respect to $\mathbf{x}$; $\mathbf{p}=\mathbf{p}(\mathbf{x}, t)$ is the source term (note that, in many common cases, $\mathbf{p = 0}$ represents no source input to the system). The PDEs are also subjected to initial and boundary conditions (IBCs), if known, denoted by $\mathcal{I}[\mathbf{x}\in \Omega, t=0; \mathbf{u}, \mathbf{u}_t] = 0$ and $\mathcal{B}[ \mathbf{x}\in \partial\Omega; \mathbf{u}, \nabla_\mathbf{x}\mathbf{u}] = 0$. For systems that obey Newton's second law of motion (e.g., $\mathbf{u}_{tt}$ in wave equations), the governing PDEs can be written in a state-space form of Eq. \eref{eq:PDE} by defining $\mathbf{v}=\{\mathbf{u}~\mathbf{u}_t\}$ as the solution variable. Our objective is to find the closed form of $\mathcal{F}[\cdot]$ from available spatiotemporal measurements which are assumed to be incomplete, scarce and noisy commonly seen in real-world applications (e.g., when data capture is very costly or the data itself is sparse in nature). We assume that the physical law is governed by only a few important terms which can be selected from a large-space library of candidate functions, where sparse regression can be applied \cite{Brunton3932, Rudye1602614, Schaeffer2017}. Inherent in this assumption leads to reformulation of Eq. \eref{eq:PDE} in the following (assuming zero or unknown source for simplicity):
\begin{equation}\label{eq:PDE-Sparse}
    \mathbf{u}_t = \boldsymbol{\phi}\boldsymbol{\Lambda}
\end{equation}
Here, $\boldsymbol{\phi}=\boldsymbol{\phi}(\mathbf{u})\in \mathbb{R}^{1\times s}$ is an extensive library of symbolic functions consisting of many candidate terms, e.g., constant, polynomial, and trigonometric terms with respect to each spatial dimension \cite{Rudye1602614, Schaeffer2017}, assembled in a row vector given by $\boldsymbol{\phi} = \left\{ 1, \mathbf{u}, \mathbf{u}^2, ..., \mathbf{u}_x, \mathbf{u}_y, ..., \mathbf{u}^3\odot\mathbf{u}_{xy}, ..., \sin(\mathbf{u}), ... \right\}$, where $\odot$ represents the element-wise Hadamard product; $s$ denotes the total number of candidate terms in the library; the subscripts in the context of $\{x, y, z\}$ depict the derivatives; $\boldsymbol{\Lambda}\in \mathbb{R}^{s\times n}$ is the sparse coefficient matrix (only the active candidate terms in $\boldsymbol{\phi}$ have non-zero values), e.g., $\boldsymbol{\Lambda} = \left[\boldsymbol{\lambda}^{u}~\boldsymbol{\lambda}^{v}~\boldsymbol{\lambda}^{w} \right]\in \mathbb{R}^{s\times 3}$ for $\mathbf{u} = \{u, v, w\}$. If there is an unknown source input, the candidate functions for $\mathbf{p}$ can also be incorporated into $\boldsymbol{\phi}$ for discovery (see \textcolor{blue}{\text{Supplementary} Note C3}). Thus, the discovery problem can then be stated as: given the spatiotemporal measurement data $\mathcal{D}_u$, find sparse $\boldsymbol{\Lambda}$ such that Eq. \eref{eq:PDE-Sparse} holds.

\begin{figure}[t!]
\centering
\includegraphics[width=0.995\linewidth]{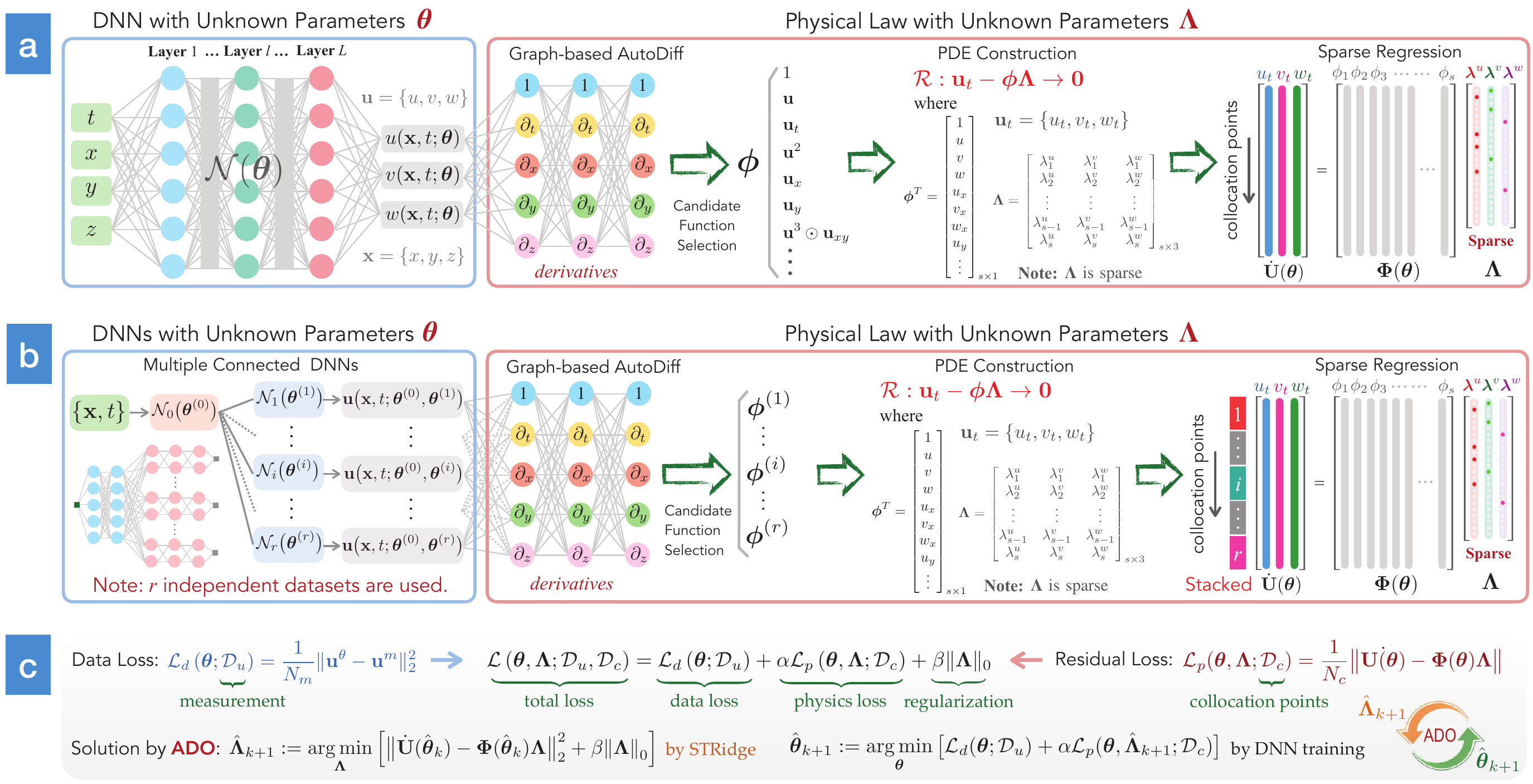}
\caption{Schematic architecture of the framework of PiDL with sparse regression for data-driven discovery of PDE(s). (a) the network for one dataset from a single IBC, (b) the ``root-branch'' network for $r\geq2$ independent datasets from multiple IBCs, and (c) schematic for training the networks based on alternating direction optimization. The network consists of two components: a DNN governed by the trainable parameters $\boldsymbol{\theta}$, which maps the spatiotemporal coordinates $\{\mathbf{x}, t\}$ to the latent solution $\mathbf{u}=\{u, v, w\}$, and the physical law described by a set of nonlinear PDEs, which are formed by the derivative candidate functions $\boldsymbol{\phi}$ parameterized by the unknown sparse coefficients $\boldsymbol{\Lambda}$. Note that, for the case of multiple independent datasets, the libraries $\boldsymbol{\phi}^{(i)}$ are concatenated to build $\boldsymbol{\phi}$ for constructing the unified governing PDE(s). The total loss function $\mathcal{L}(\boldsymbol{\theta}, \boldsymbol{\Lambda}; \mathcal{D}_u, \mathcal{D}_c)$ is composed of the data loss $\mathcal{L}_d(\boldsymbol{\theta}, \mathcal{D}_u)$, the physics loss $\alpha\mathcal{L}_p(\boldsymbol{\theta}, \boldsymbol{\Lambda}; \mathcal{D}_c)$, and the $\ell_0$ regularization term $\beta\|\boldsymbol{\Lambda}\|_0$ that promotes the sparsity. Here, $\alpha$ and $\beta$ denote the relative weighting of the loss functions, while $\mathcal{D}_u$ and $\mathcal{D}_c$ represent the measurement data and collocation samples respectively. Note that the physics loss, in a residual form, is only evaluated on the spatiotemporal collocation samples. The colored dots in the sparse coefficients matrix (or vector) on the right denote non-zero values. Simultaneous optimization of the unknown parameters $\{\boldsymbol{\theta}, \boldsymbol{\Lambda}\}$ leads to both the trained DNN for inference of the data-driven full-field solution and the discovered parsimonious closed-form PDEs.}
\label{fig:PINN}
\end{figure}

We present an interpretable PiDL paradigm with sparse regression to simultaneously model the system response and identify the parsimonious closed form of the governing PDE(s). The innovative algorithm architecture of this method is shown in Fig. \ref{fig:PINN}, where datasets sampled from two different IBC scenarios are considered: (1) one dataset from a single IBC and (2) $r\geq2$ independent datasets from multiple IBCs. For the case of single dataset, we interpret the latent solution $\mathbf{u}$ by a DNN (denoted by $\mathcal{N}$), namely, $\mathbf{u}^\theta = \mathbf{u}(\mathbf{x}, t; \boldsymbol{\theta})$, where $\boldsymbol{\theta}$ represents the DNN trainable parameters including weights and biases, as shown in Fig. \ref{fig:PINN}a. When multiple independent datasets are available, a ``root-branch'' DNN depicted in Fig. \ref{fig:PINN}b is designed to approximate the latent solutions $\mathbf{u}_i$ ($i=1, ..., r$) corresponding to different IBCs, viz., $\mathbf{u}^\theta_i =\mathbf{u}\big(\mathbf{x}, t; \boldsymbol{\theta}^{(0)}, \boldsymbol{\theta}^{(i)}\big)$, where $\boldsymbol{\theta}^{(0)}$ and $\boldsymbol{\theta}^{(i)}$ denote the trainable parameters of the root layers $\mathcal{N}^{(0)}$ and the branch layers $\mathcal{N}^{(i)}$, respectively. Noteworthy, the IBCs are unnecessarily  either known \textit{a priori} or measured since the measurement data already reflects the specific IBC (e.g., there exists a one-to-one mapping between the IBC and the PDE solution). The DNN essentially plays a role as a nonlinear functional to approximate the latent solution with the data loss function $\mathcal{L}_d(\boldsymbol{\theta}; \mathcal{D}_u)$. With graph-based automatic differentiation where derivatives on $\mathbf{u}$ are evaluated at machine precision, the library of candidate functions $\boldsymbol{\phi}^\theta$ can be computed from the DNN. For the case of multiple independent datasets, the libraries $\boldsymbol{\phi}^{(i)}$ resulted from the branch nets are concatenated to build $\boldsymbol{\phi}^\theta$ for constructing the unified governing PDE(s). Thus, the sparse representation of the reconstructed PDE(s) can be written in a residual form, namely, $\boldsymbol{\mathcal{R}}^{\theta}:=\mathbf{u}_t^{\theta} - \boldsymbol{\phi}^{\theta}\boldsymbol{\Lambda} \rightarrow \mathbf{0}$, where $\boldsymbol{\mathcal{R}}^{\theta}\in \mathbb{R}^{1\times n}$ denotes the PDE residuals. The basic concept is to adapt both the DNN trainable parameters $\boldsymbol{\theta}$ and the PDE coefficients $\boldsymbol{\Lambda}$ such that the neural network can fit the measurement data while satisfying the constraints defined by the underlying PDE(s). The PDE residuals will be evaluated on a large number of collocation points $\mathcal{D}_c=\{\mathbf{x}_i, t_i\}_{i=1}^{N_c}$, randomly sampled in the spatiotemporal space, leading to the residual physics loss function $\mathcal{L}_p(\boldsymbol{\theta}, \boldsymbol{\Lambda}; \mathcal{D}_c)$. When multiple IBCs are considered, the measurement data and the collocation points will be stacked when calculating the data loss and the physics loss (based on a unified physics residual formulation $\boldsymbol{\mathcal{R}}^{\theta} \rightarrow \mathbf{0}$).

The total loss function for training the overall PiDL network is thus composed of the data loss $\mathcal{L}_d$, the residual physics loss $\mathcal{L}_p$ and a regularization term, expressed as:
\begin{equation}\label{eq:totalloss}
    \mathcal{L}(\boldsymbol{\theta}, \boldsymbol{\Lambda}; \mathcal{D}_u, \mathcal{D}_c) = \mathcal{L}_d(\boldsymbol{\theta}; \mathcal{D}_u) + \alpha\mathcal{L}_p(\boldsymbol{\theta}, \boldsymbol{\Lambda}; \mathcal{D}_c) + \beta\|\boldsymbol{\Lambda}\|_0
\end{equation}
where $\alpha$ is the relative weighting of the residual physics loss function; $\beta$ is the regularization parameter; $\|\cdot\|_0$ represents the $\ell_0$ norm. Optimizing the total loss function can produce a DNN that can not only predict the data-driven full-field system response, but also uncover the parsimonious closed-form PDE(s), i.e., $\{\boldsymbol{\theta}^\star, \boldsymbol{\Lambda}^\star\}: = {\arg\min}_{\{\boldsymbol{\theta}, \boldsymbol{\Lambda}\}} \big[\mathcal{L}(\boldsymbol{\theta}, \boldsymbol{\Lambda}; \mathcal{D}_u, \mathcal{D}_c)\big]$, where $\{\boldsymbol{\theta}^\star, \boldsymbol{\Lambda}^\star\}$ denote the optimal set of parameters. Noteworthy, the total loss function has an implicit complex form, and thus, directly solving the optimization problem is highly intractable since the $\ell_0$ regularization makes this problem $np$-hard. To address this challenge, we present an alternating direction optimization (ADO) algorithm that divides the overall optimization problem into a set of tractable subproblems to sequentially optimize the trainable parameters, as shown in Fig. \ref{fig:PINN}c. Pre-training of PiDL is conducted before running the ADO algorithm for discovery, by simply replacing $\|\boldsymbol{\Lambda}\|_0$ in Eq. \eref{eq:totalloss} with $\|\boldsymbol{\Lambda}\|_1$ where brute-force gradient-based optimization for both $\boldsymbol{\theta}$ and $\boldsymbol{\Lambda}$ becomes applicable. The $\ell_1$-regularized pre-training can accelerate the convergence of ADO by providing an admissible ``initial guess''. More detailed formulation and algorithm description are found in \textcolor{blue}{Method} and \textcolor{blue}{\text{Supplementary} Note A}.

The synergy of DNN and sparse regression results in the following outcome: the DNN provides accurate modeling of the latent solution, its derivatives and possible candidate function terms as a basis for constructing the governing PDE(s), while the sparsely represented PDE(s) in turn constraints the DNN modeling and projects correct candidate functions, eventually turning the measured system into closed-form PDE(s).

\begin{table}[t!]
\begin{threeparttable}
\caption{Summary of the PiDL discovery results in the context of accuracy for a range of canonical models.}
{\scriptsize
\begin{tabular}{lrrrr}
\toprule
PDE name & Err. (N-0\%) & Err. (N-1\%) & Err. (N-10\%) & Description of data discretization  \\
\midrule
Burgers' & 0.01$\pm$0.01\% & 0.19$\pm$0.11\% & 1.15$\pm$1.20\% & $x\in[-8,8]_{d=256}$, $t\in[0,10]_{d=101}$, sub. 1.95\% \\
KS & 0.07$\pm$0.01\% & 0.61$\pm$0.04\% & 0.71$\pm$0.06\% & $x\in[0,100]_{d=1024}$, $t\in[0,100]_{d=251}$, sub. 12.3\% \\
Schr\"{o}dinger & 0.09$\pm$0.04\% & 0.65$\pm$0.29\% & 2.31$\pm$0.28\% & $x\in[-4.5, 4.5]_{d=512}$, $t\in[0,\pi]_{d=501}$, sub. 37.5\% \\
NS & 0.66$\pm$0.72\% & 0.86$\pm$0.63\% & 1.40$\pm$1.83\% & $x\in[0, 9]_{d=449}$, $y\in[-2, 2]_{d=199}$, $t\in[0,30]_{d=151}$, sub. 0.22\% \\
$\lambda$-$\omega$ RD & 0.07$\pm$0.08\% & 0.25$\pm$0.30\% & 4.78$\pm$3.66\% & $x,y\in[-10,10]_{d=256}$, $t\in[0,10]_{d=201}$, sub. 0.29\% \\
\bottomrule
\end{tabular}\label{discovery_result}
\begin{tablenotes}
\item Note: The error is defined as the average relative error of the identified non-zero coefficients w.r.t. the ground truth. The percentage values in the parentheses denote the noise levels (e.g., noise free 0\%, 1\% and 10\%) and the subscript $d$ represents the number of discretization. Our method is also compared with SINDy (the PDE-FIND approach presented in \cite{Rudye1602614}) as illustrated in \textcolor{blue}{\text{Supplementary} Table S1}. It is noted that much less measurement data polluted with a higher level of noise are used in our discovery. Gaussian white noise is added to the synthetic response with the noise level defined as the root-mean-square ratio between the noise and the exact solution.
\end{tablenotes}
}
\end{threeparttable}
\end{table}

\begin{figure}[t!]
\centering
\includegraphics[width=0.99\linewidth]{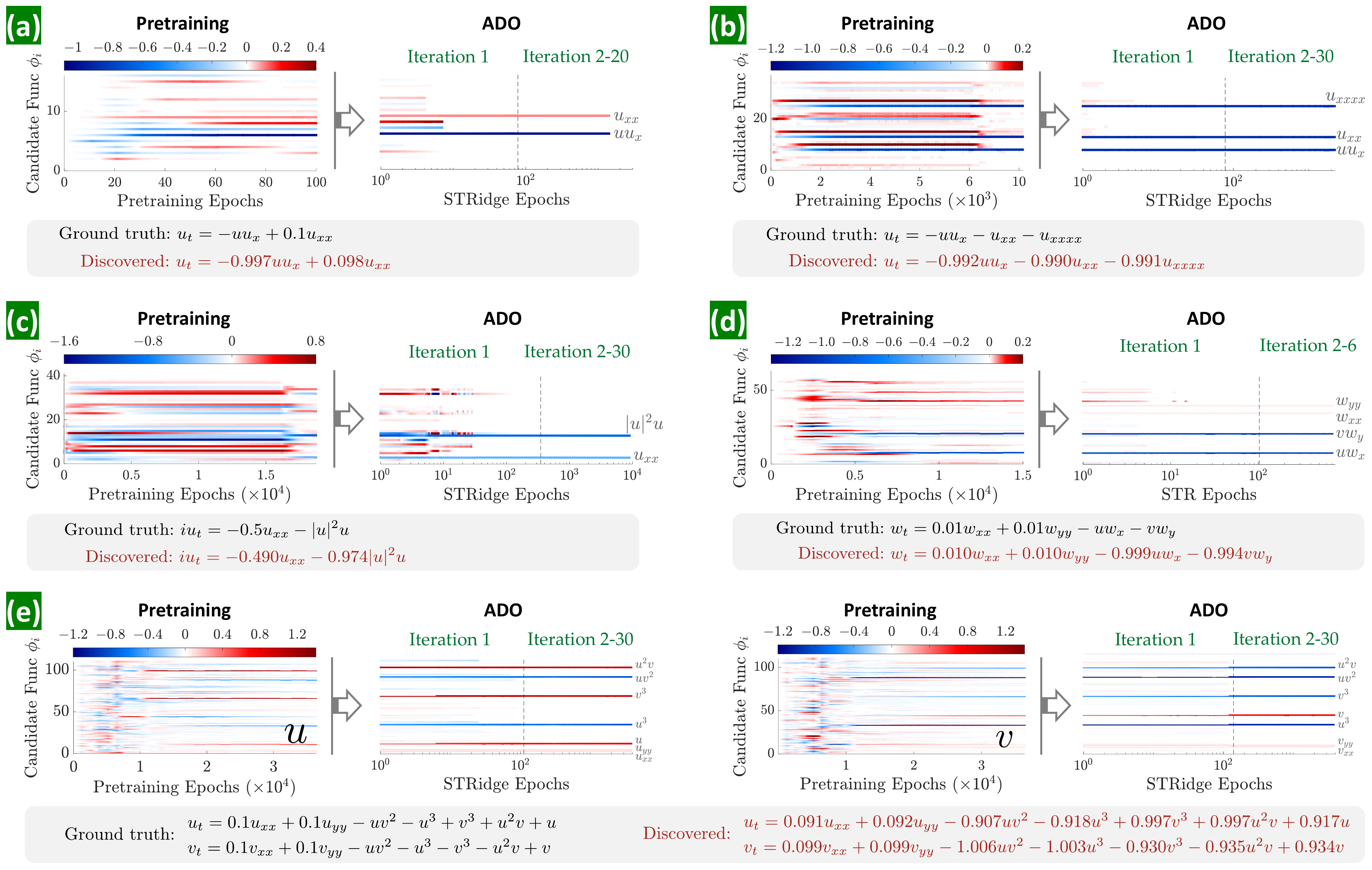}
\caption{Discovery of selected benchmark PDEs for sparsely sampled measurement data with 10\% noise. (a) Discovered Burgers' equation: evolution of the sparse coefficients $\boldsymbol{\Lambda}\in\mathbb{R}^{16\times1}$ for 16 candidate functions $\boldsymbol{\phi}\in\mathbb{R}^{1\times16}$ used to form the PDE, where the color represents the coefficient value. (b) Discovered KS equation: Evolution of the sparse coefficients $\boldsymbol{\Lambda}\in\mathbb{R}^{36\times1}$ for 36 candidate functions $\boldsymbol{\phi}\in\mathbb{R}^{1\times36}$. (c) Discovered nonlinear Schr\"{o}dinger equation: evolution of the sparse coefficients $\boldsymbol{\Lambda}\in\mathbb{R}^{40\times1}$ for the candidate functions $\boldsymbol{\phi}\in\mathbb{R}^{1\times40}$. (d) Discovered NS equation: evolution of the sparse coefficients $\boldsymbol{\Lambda}\in\mathbb{R}^{60\times1}$ for 60 candidate functions $\boldsymbol{\phi}\in\mathbb{R}^{1\times60}$. (e) Discovered RD equations: evolution of the sparse coefficients $\boldsymbol{\lambda}^u\in\mathbb{R}^{110\times1}$ and $\boldsymbol{\lambda}^v\in\mathbb{R}^{110\times1}$ ($\boldsymbol{\Lambda}=[\boldsymbol{\lambda}^u~\boldsymbol{\lambda}^v]$) for 110 candidate functions $\boldsymbol{\phi}\in\mathbb{R}^{1\times110}$ used to reconstruct the $u$-equation and the $v$-equation, respectively.}
\label{fig:coeff_benchmark}
\end{figure}

\subsection*{Discovery of Benchmark PDEs with Single Dataset}
\begin{figure}[t!]
\centering
\includegraphics[width=0.99\linewidth]{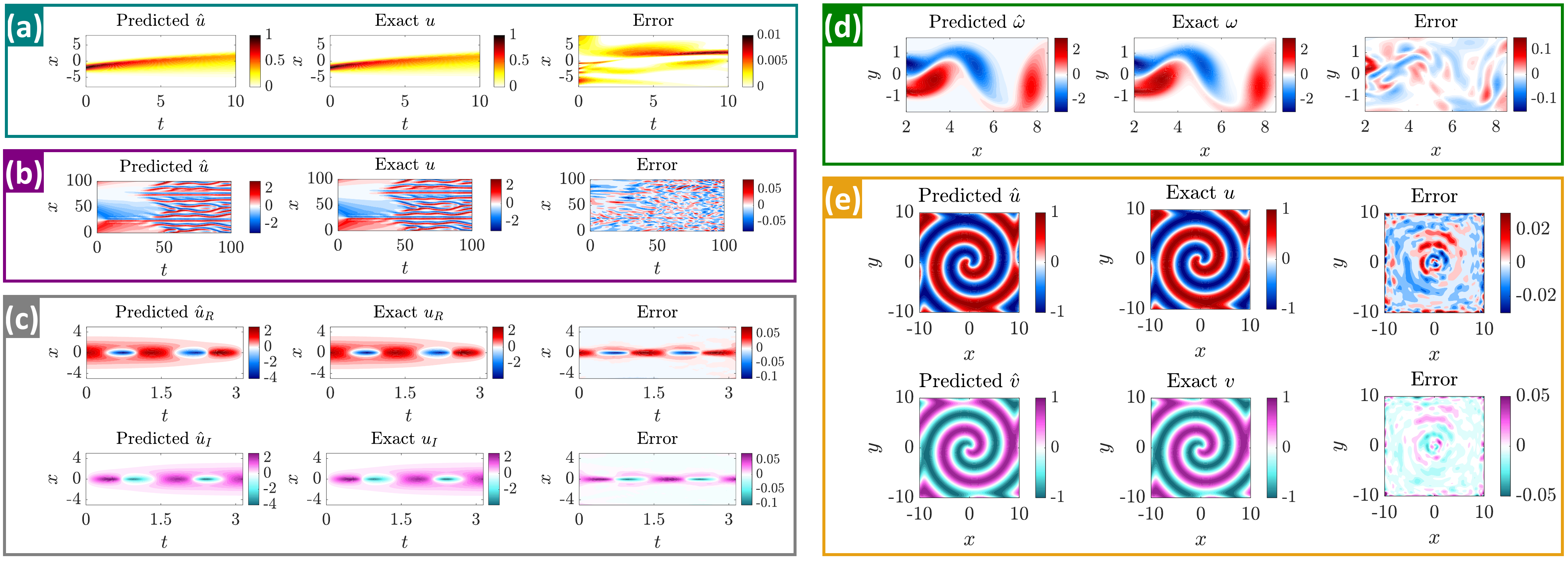}
\caption{Predicted responses compared with the exact solutions for selected canonical PDEs. (a) Burgers' equation, (b) KS equation, (c) nonlinear Schr\"{o}dinger equation, (d) NS equation, and (e) $\lambda$-$\omega$ RD equations. Note that the sparsely sampled measurement data has 10\% noise.}
\label{fig:resp_benchmark}
\end{figure}

We observe the efficacy and robustness of our methodology on a group of canonical PDEs used to represent a wide range of physical systems with nonlinear, periodic and/or chaotic behaviors. In particular, we discover the closed forms of Burgers', Kuramoto-Sivashinsky (KS), nonlinear Schr\"{o}dinger, Navier-Stokes (NS), and $\lambda$-$\omega$ Reaction-Diffusion (RD) equations from scarce and noisy time-series measurements recorded by a number of sensors at fixed locations (data are polluted with Gaussian white noise) from a single IBC. Results are presented in Table \ref{discovery_result}, Fig. \ref{fig:coeff_benchmark} and Fig. \ref{fig:resp_benchmark}, which show quite accurate discovery and demonstrate satisfactory performance of the proposed method and its robustness to measurement data scarcity and noise. We also compare our method with SINDy considering different levels of data scarcity and noise (summarized in \textcolor{blue}{\text{Supplementary} Note B6} and \textcolor{blue}{Table S1}).

\textit{\textbf{Burgers' Equation:}}
We first consider a dissipative system with the dynamics governed by a 1D viscous Burgers' equation expressed as $u_t = - uu_x + \nu u_{xx}$, where $\nu$ (equal to 0.1) denotes the diffusion coefficient. The equation describes the decaying stationary viscous shock of a system after a finite period of time, commonly found in simplified fluid mechanics, nonlinear acoustics and gas dynamics. We test the PiDL approach on the recorded traveling shock waves from the solution to Burgers' equation subjected to a Gaussian initial condition. In particular, 5 sensors are randomly placed at fixed locations among the 256 spatial grids and record the wave for 101 time steps, leading to 1.95\% of the dataset used in \cite{Rudye1602614}. A full description of the dataset, design of the library of candidate functions (16 terms) and model training is given in \textcolor{blue}{\text{Supplementary} Note B.1.1}. Fig. \ref{fig:coeff_benchmark}a shows the discovered Burgers' equation for a dataset with 10\% noise. The evolution of the coefficients $\boldsymbol{\Lambda}\in\mathbb{R}^{16\times1}$ illustrates robust convergence to the ground truth (error about 1.2\%), resulting in accurate discovery. The trained PiDL properly reproduces the dynamical response from noisy measurements (e.g., the full-field $\ell_2$ prediction error is 2.02\%) as shown in Fig. \ref{fig:resp_benchmark}a. The ADO algorithm converges only after the first alternating iteration and shows capacity to recover the correct sparsity pattern of the PDE. We also discover the Burgers' equation with an unknown/unmeasured source $\sin(x)\sin(t)$, given scarce $u$-measurement with 10\% noise. When discovering the underlying governing equation, the source should be considered and reconstructed concurrently. In this case, we incorporate 14 source candidate functions, composed of $\{\sin(t), \sin(x), \cos(t), \cos(x)\}$ and their combination, into the aforementioned library, resulting in a total of 30 candidate terms for simultaneous discovery of the PDE and reconstruction of the unknown source. The corresponding discovery result is summarized in Extended Data Fig. \ref{Fig:Source}, which includes the discovered equation and source function, the evolution of sparse coefficients $\boldsymbol{\Lambda}\in\mathbb{R}^{30\times1}$, and the predicted full-field response. It turns out that both PDE and source terms along with their coefficients are well identified. Nevertheless, if the source is very complex with its general expression or form completely unknown, distinct challenges arise when designing the source candidate functions. This may require an extraordinarily large-space library to retain diversifying representations, and thus pose additional computational complexity for accurate discovery of the PDEs. More discussions are presented in \textcolor{blue}{\text{Supplementary} Note C3}.

\textit{\textbf{Kuramoto-Sivashinsky (KS) Equation:}} 
Another dissipative system with intrinsic instabilities is considered, governed by the 1D Kuramoto-Sivashinsky (KS) equation $u_t = - uu_x - u_{xx} - u_{xxxx}$, where the reverse diffusion term $-u_{xx}$ leads to the disruptive behavior while the fourth-order derivative $u_{xxxx}$ introduces chaotic patterns as shown in Fig. \ref{fig:resp_benchmark}b, making an ideal  test problem for equation discovery. The KS equation is widely used to model the instabilities in laminar flame fronts and dissipative trapped-ion modes among others. We randomly choose 320 points as fixed sensors and record the wave response for 101 time steps, resulting in 12.3\% of the dataset used in \cite{Rudye1602614}. A total of 36 candidate functions are employed to construct the underlying PDE. Detail description of this example is found in \textcolor{blue}{\text{Supplementary} Note B.1.2}. It is notable that the chaotic behavior poses significant challenges in approximating the full-field spatiotemporal derivatives, especially the high-order $u_{xxxx}$, from poorly measured data for discovery of such a PDE. Existing methods (e.g., the family of SINDy methods \cite{Rudye1602614, Schaeffer2017}) eventually fail in this case given very coarse and noisy measurements. Nevertheless, PiDL successfully distils the closed form of the KS equation from subsampled sparse data with 10\% noise, shown in Fig. \ref{fig:coeff_benchmark}b. The evolution of the coefficients $\boldsymbol{\Lambda}\in\mathbb{R}^{36\times1}$ in Fig. \ref{fig:coeff_benchmark}b illustrates that both the candidate terms and the corresponding coefficients are correctly identified (close to the original parameters; error around 0.7\%) within a few ADO iterations. The predicted full-field wave by the trained PiDL also coincides with the exact solution at a relative $\ell_2$ error of 1.87\% (Fig. \ref{fig:resp_benchmark}b).

\textit{\textbf{Nonlinear Schr\"{o}dinger Equation:}} 
In the third example, we discover the nonlinear Schr\"{o}dinger equation, $i u_t = -0.5u_{xx}-|u|^2u$, where $u$ is a complex field variable. This well-known equation is widely used in modeling the propagation of light in nonlinear optical fibers, Bose-Einstein condensates, Langmuir waves in hot plasmas, and so on. We take 37.5\% subsamples (e.g., randomly selected from the spatial grids) of the dataset as shown in Table \ref{discovery_result} to construct the PDE using 40 candidate functions $\boldsymbol{\phi}\in\mathbb{R}^{1\times40}$. Since the function is complex-valued, we model separately the real part ($u_R$) and the imaginary part ($u_I$) of the solution in the output of the DNN, assemble them to obtain the complex solution $u = u_R+iu_I$, and construct the complex-valued candidate functions for PDE discovery. To avoid complex gradients in optimization, we use the modulus $|u|$, instead of the $\ell_2$ norm shown in Eq. \eref{eq:residualloss}, for the residual physics loss $\mathcal{L}_p$ (see \textcolor{blue}{\text{Supplementary} Note B.1.3} for more details). Fig. \ref{fig:coeff_benchmark}c shows the discovered Schr\"{o}dinger equation for the case of 10\% noise. The evolution history of the sparse coefficients $\boldsymbol{\Lambda}\in\mathbb{R}^{40\times1}$ clearly shows the convergence to the actual values (Fig. \ref{fig:coeff_benchmark}c; error about 4.14\%) resulting in accurate closed-form identification of the PDE, while the reconstructed full-field response, for both real and imaginary parts, matches well the exact solution with a slight relative $\ell_2$ error of 1\% (Fig. \ref{fig:resp_benchmark}c).

\textit{\textbf{Navier-Stokes (NS) Equation:}} 
We consider a 2D fluid flow passing a circular cylinder with the local rotation dynamics governed by the well-known Navier-Stokes vorticity equation $w_t = -(\mathbf{u} \cdot \nabla) w + \nu \nabla^2 w $, where $w$ is the spatiotemporally variant vorticity, $\mathbf{u}=\{u, v\}$ denotes the fluid velocities, and $\nu$ is the kinematic viscosity ($\nu = 0.01$ at Reynolds number 100). We leverage the open simulation data \cite{Rudye1602614} and subsample a dataset of the flow response $\{u, v, w\}$ at 500 spatial locations randomly picked within the indicated region in \textcolor{blue}{\text{Supplementary} Fig. S4}, which record time series for 60 time steps. The resulting dataset is only 10\% of that used in \cite{Rudye1602614}. A comprehensive discussion of this example is found in \textcolor{blue}{\text{Supplementary} Note B.1.4}. Fig. \ref{fig:coeff_benchmark}d summarizes the result of the discovered NS equation for a dataset with 10\% noise. It is encouraging that the uncovered PDE expression is almost identical to the ground truth, for both the derivative terms and their coefficients, even under 10\% noise corruption. The coefficients $\boldsymbol{\Lambda}\in\mathbb{R}^{60\times1}$, corresponding to 60 candidate functions $\boldsymbol{\phi}\in\mathbb{R}^{1\times60}$, converge very quickly to the correct values with precise sparsity right after the first ADO iteration (Fig. \ref{fig:coeff_benchmark}d). The vorticity patterns and magnitudes are also well predicted as indicated by the snapshot (at $t=23.8$) shown in Fig. \ref{fig:resp_benchmark}d (the full-field $\ell_2$ error for all snapshots is about 2.57\%). This example provides a compelling test case for the proposed PiDL approach which is capable of discovering the closed-form NS equation with scarce and noisy data.

\textit{\textbf{Reaction-Diffusion (RD) Equations:}}
The examples above are mostly low-dimensional models with limited complexity. We herein consider a $\lambda$-$\omega$ reaction-diffusion (RD) system in a 2D domain with the pattern forming behavior governed by two coupled PDEs: $u_t=0.1\nabla^2u+\lambda(g)u-\omega(g)v$ and $v_t=0.1\nabla^2v+\omega(g)u+\lambda(g)v$, where $u$ and $v$ are the two field variables, $g=u^2+v^2$, $\omega=-g^2$, and $\lambda=1-g^2$. The RD equations exhibit a wide range of behaviors including wave-like phenomena and self-organized patterns found in chemical and biological systems. The particular RD equations considered here display spiral waves subjected to periodic boundary conditions. Full details on the dataset, selection of candidate functions and hyperparameter setup of the PiDL model are given in \textcolor{blue}{\text{Supplementary} Note B.1.5}. Fig. \ref{fig:coeff_benchmark}e shows the evolution of the sparse coefficients $\boldsymbol{\lambda}^u, \boldsymbol{\lambda}^v\in\mathbb{R}^{110\times1}$ for 110 candidate functions $\boldsymbol{\phi}\in\mathbb{R}^{1\times110}$, given a dataset with 10\% noise. Both the sparse terms and the associated coefficients are precisely identified to form the the closed-form equations (as depicted in Fig. \ref{fig:coeff_benchmark}e). Due to the complexity of the PDEs and the high dimension, slightly more epochs are required in ADO to retain reliable convergence. The predicted response snapshots (e.g., at $t=2.95$) by the trained PiDL in Fig. \ref{fig:resp_benchmark}e are close to the ground truth. This example shows especially the great ability and robustness of our method for discovering governing PDEs for high-dimensional systems from highly noisy data.

\subsection*{Discovery of PDEs with Multiple Independent Datasets}
To demonstrate the ``root-branch'' network presented in Fig. \ref{fig:PINN}b for discovery of PDE(s) based on multiple independent datasets sampled under different IBCs, we consider (1) the 1D Burgers' equation with light viscosity that exhibits a shock behavior, and (2) a 2D Fitzhugh-Nagumo (FN) type reaction-diffusion system that describes activator-inhibitor neuron activities excited by external stimulus. The measurement data are sparsely sampled (e.g., time series or snapshots) with 10\% noise under three different IBCs. Note that the IBCs are unnecessarily either measured or known \textit{a priori} since the measurements already reflect the specific IBC which holds uniquely one-to-one mapping to the system response. The discovery results are discussed as follows.

\begin{figure}[t!]
\centering
\includegraphics[width=0.75\linewidth]{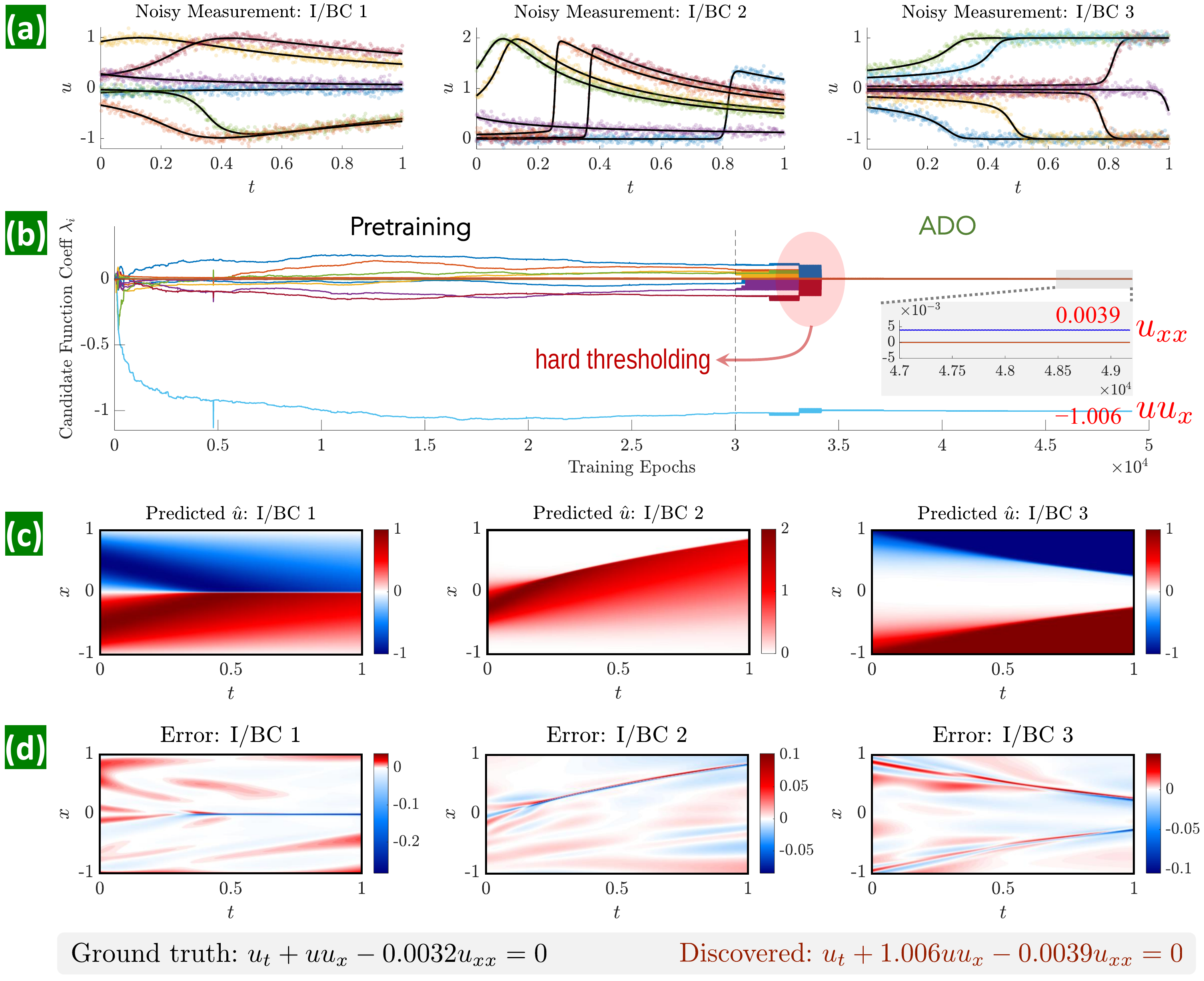}
\caption{Discovered Burgers' equation with small viscosity based on datasets sampled under three IBCs with 10\% noise. (a) Visualization of noisy measurements for the three datasets. Note that there are 30 sensors and only a few are illustrated in this figure. (b) Evolution of the sparse coefficients $\boldsymbol{\Lambda}\in\mathbb{R}^{16\times1}$ for 16 candidate functions $\boldsymbol{\phi}\in\mathbb{R}^{1\times16}$ used to construct the PDE, where the color represents the coefficient value. The correct terms ($uu_{x}$ and $u_{xx}$) and their coefficients are successfully identified while other redundant terms are eliminated by ADO. (c-d) The predicted responses and errors for three IBC cases. The ground truth is not listed herein since the visualization is almost indistinguishable from the prediction (see \textcolor{blue}{\text{Supplementary} Fig. S.7}). The relative full-field $\ell_2$ error of the stacked prediction is 2.24\%.}
\label{fig:Burgers_MultiIBC}
\end{figure}

\textit{\textbf{Burgers' Equation with Shock Behavior:}} 
In this example, we test the previously discussed Burgers' equation with a small diffusion/viscosity parameter ($\nu = 0.01/\pi\approx0.0032$) based on datasets generated by imposing three different IBCs. Such a small coefficient creates shock formation in a compact area with sharp gradient (see Fig. \ref{fig:Burgers_MultiIBC}c) that could challenge the DNN's approximation ability and thus affect the discovery. The three initial and Dirichlet boundary conditions include:
\begin{align*}
&\text{IBC 1:~} u(x, 0) = -\sin(\pi x), u(-1, t) = u(1, t) = 0 \\
&\text{IBC 2:~} u(x, 0) = \mathcal{G}(x), u(-1, t) = u(1, t) = 0 \\
&\text{IBC 3:~} u(x, 0) = -x^3, u(-1, t) = 1, u(1, t) = -1
\end{align*}
where $\mathcal{G}$ denotes a Gaussian function. Although the measurement datasets for different IBCs exhibit completely distinct system responses, they obey the same underlying PDE, namely, $u_t = - uu_x + 0.0032 u_{xx}$. For all IBCs, we assume that there are 30 sensors randomly deployed in space ($x\in[-1, 1]$) measuring the wave traveling (e.g., $u$) for 500 time instants ($t\in[0, 1]$). A denser sensor grid is needed herein, compared with the previous Burgers' example, in order to capture the shock behaviors. Fig. \ref{fig:Burgers_MultiIBC}a shows some of the measurements recorded by 6 typical sensors under 10\% noise. A three-branch network ($r=3$) shown in Fig. \ref{fig:PINN}b is used for discovery. The full description of the dataset, the library of candidate functions (16 terms) and model training is given in \textcolor{blue}{\text{Supplementary} Note B.3.1}. Fig. \ref{fig:Burgers_MultiIBC}b depicts the evolution of the coefficients ($\boldsymbol{\Lambda}\in\mathbb{R}^{16\times1}$) of candidate functions, where the correct terms in the library ($uu_{x}$ and $u_{xx}$) are successfully distilled while other redundant terms are eliminated (e.g., hardly thresholded to zero) by ADO. The coefficients of the active terms are accurately identified as well (in particular the small viscosity parameter that leads to shock formation, e.g., 0.0039). The discovered PDE reads $u_t = -1.006 uu_x + 0.0039 u_{xx}$. Fig. \ref{fig:Burgers_MultiIBC}c-d show the predicted responses and errors for three IBC cases, with a stacked full-field $\ell_2$ error of 2.24\%.

\begin{figure}[t!]
\centering
\includegraphics[width=0.9\linewidth]{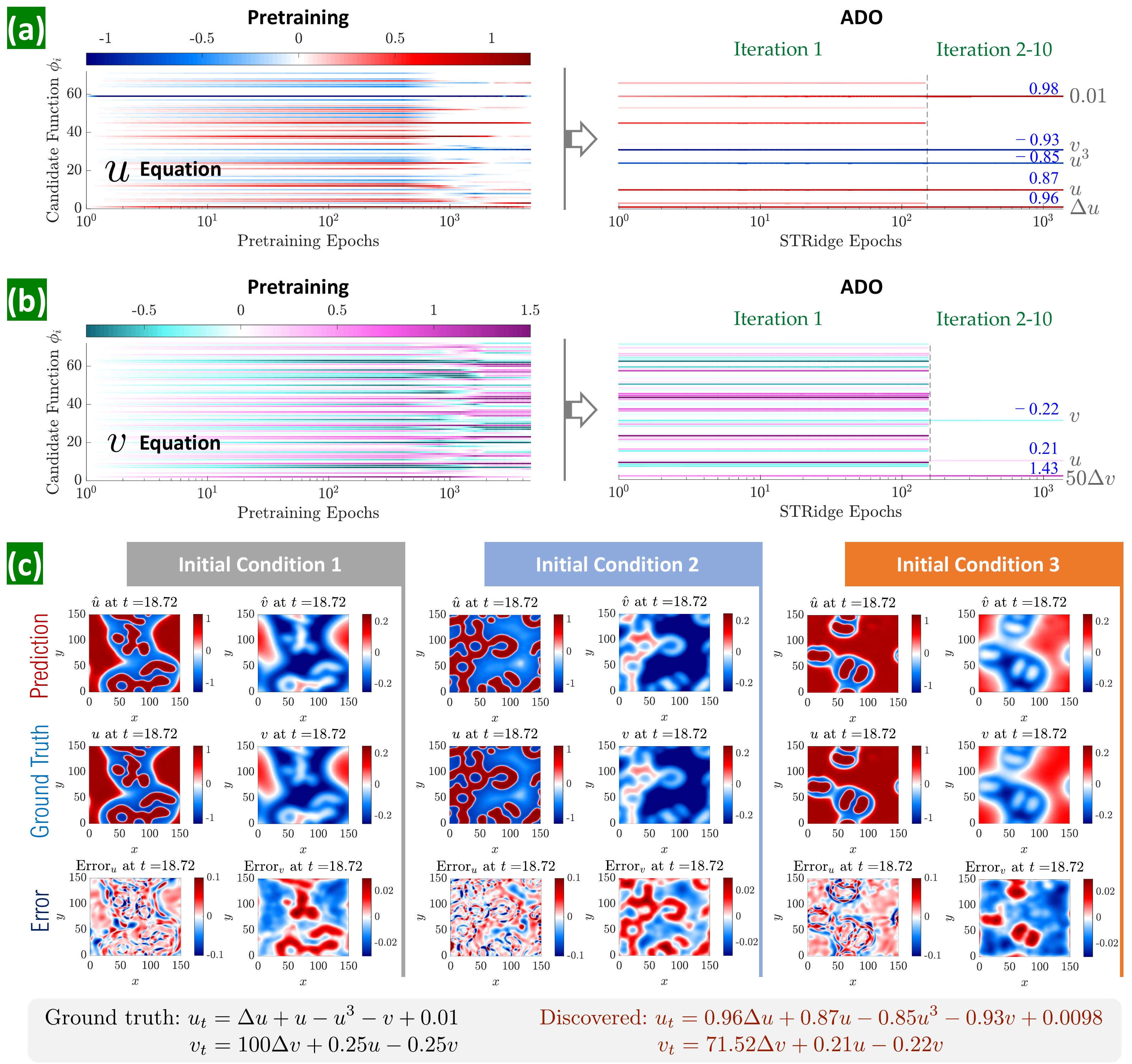}
\caption{Discovered Fitzhugh-Nagumo equations based on data sampled under three initial conditions (ICs) with 10\% noise. (a) Evolution of the sparse coefficients $\boldsymbol{\lambda}_u\in\mathbb{R}^{72\times1}$ for 72 candidate functions used to construct the first PDE ($u$-equation), where the color represents the coefficient value. (b) Evolution of the sparse coefficients $\boldsymbol{\lambda}_v\in\mathbb{R}^{72\times1}$ for the second PDE ($v$-equation). For visualization purpose, we re-scale the identified coefficients of the constant stimulus term ``1'' in the $u$-equation by multiplying 100 and of the diffusion term $\Delta v$ in the $v$-equation by dividing 50. (c) Snapshots of predicted response, ground truth and error distributions for all three ICs at an unmeasured time instance ($t=18.72$). The relative $\ell_2$ error for the predicted full-field response (stacked $u$ and $v$) is 5.04\%.}
\label{fig:FNeq}
\end{figure}

\textit{\textbf{Fitzhugh-Nagumo (FN) Reaction-Diffusion System:}} 
We consider the Fitzhugh-Nagumo (FN) type reaction-diffusion system, in a 2D domain $\Omega = [0, 150]\times [0, 150]$ with periodic boundary conditions, whose governing equations are expressed by two coupled PDEs: $u_t = \gamma_u \Delta u + u - u^3 - v + \alpha$ and $v_t = \gamma_v \Delta v + \beta(u - v)$. Here, $u$ and $v$ represent two interactive components/matters (e.g., biological), $\gamma_u = 1$ and $\gamma_v = 100$ are diffusion coefficients, $\alpha = 0.01$ and $\beta = 0.25$ are the coefficients for reaction terms, and $\Delta$ is the Laplacian operator. The FN equations are commonly used to describe biological neuron activities excited by external stimulus ($\alpha$), which exhibit an activator-inhibitor system because one equation boosts the production of both components while the other equation dissipates their new growth. Three random fields are taken as initial conditions to generate three independent datasets for discovery, each of which consists of 31 low-resolution snapshots (projected into a $31\times31$ grid) down-sampled from the high-fidelity simulation under a 10\% noise condition (see Extended Data Fig. \ref{fig:FN_MultiIC_Meas}). We assume the diffusion terms ($\Delta u$ and $\Delta v$) are known in the PDEs, whose coefficients ($\gamma_u$ and $\gamma_v$) yet need to be identified. A library with 72 candidate functions ($\boldsymbol{\phi}\in\mathbb{R}^{1\times72}$) is designed for discovery of the coupled PDEs (in particular, the nonlinear reaction terms). Similar to the previous example, a root-branch network shown in Fig. \ref{fig:PINN}b is employed for discovery. More description of the data generation, the specific candidate functions and model training can be found in \textcolor{blue}{\text{Supplementary} Note B.3.2}. Fig. \ref{fig:FNeq}a-b depict the evolution of the sparse coefficients $\boldsymbol{\lambda}^u, \boldsymbol{\lambda}^v\in\mathbb{R}^{72\times1}$ for 72 candidate functions. The pretraining step provides a redundant projection of the system onto 72 candidates; however, minor candidates are pruned out right after the first ADO iteration. The rest ADO iterations continue to refine all the trainable parameters including $\boldsymbol{\theta}$, $\boldsymbol{\lambda}^u$ and $\boldsymbol{\lambda}^v$. The finally discovered PDEs are listed in Fig. \ref{fig:FNeq} in comparison with the ground truth. It is seen that the form of the PDEs is precisely uncovered with all correct active terms (including the unknown external stimulus in the first equation). The corresponding identified coefficients are generally close to the ground truth except the diffusion coefficient for $v$ (i.e., $\gamma_v$) which seems to be a less sensitive parameter according to our test. It should be noted that, given very scarce and noisy measurement datasets in this example, the ``root-branch'' DNN is faced with challenges to accurately model the solutions with sharp propagating fronts (see Fig. \ref{fig:FNeq}c). The less accurate solution approximation by DNN then affects the discovery precision. This issue can be naturally alleviated by increasing the spatiotemporal measurement resolution (even still under fairly large noise pollution, e.g., 10\%). Nevertheless, the exact form of the PDEs is successfully discovered in this challenging example, which is deemed more important since the coefficients can be further tuned/calibrated when additional data arrives. Fig. \ref{fig:FNeq}c shows typical snapshots of the predicted $u$ and $v$ components, the ground truth reference and the error distributions for one unmeasured time instance ($t=18.72$). The stacked full-field $\ell_2$ error is 5.04\%.

\subsection*{Experimental Discovery of Cell Migration and Proliferation}
The last example is placed to demonstrate the proposed approach for discovering a governing PDE that describes cell migration and proliferation, based on the sparse and noisy experimental data collected from \textit{in vitro} cell migration (scratch) assays \cite{jin2016reproducibility}. The 1D cell density distributions at different time instants (0h, 12h, 24h, 36h, 48h) were extracted from high-resolution imaging via image segmentation and cell counting. A series of assays were performed under different initial cell densities (e.g., the total number of cells spans from 10,000 to 20,000 following the designated initial distribution in the test well shown in Extended Data Fig. \ref{fig:Cell_meas}a at $t=0$h). More detailed description of the experiment setup and datasets can be found in \cite{jin2016reproducibility}. Our objective herein is to uncover a parsimonious PDE for modeling the dynamics of cell density $\rho(x, t)$. Here, we consider four scenarios with the initial number of cells ranging from 14,000, 16,000, 18,000 to 20,000. We take the mean of the test data from three identically-prepared experimental replicates for each scenario (see Extended Data Fig. \ref{fig:Cell_meas}b-e) to train our model shown in Fig. \ref{fig:PINN}a for PDE discovery. Given our prior knowledge that the cell dynamics can be described by a diffusion (migration) and reaction (proliferation) process, we assume the PDE holds the form of $\rho_t = \gamma\rho_{xx} + \mathcal{F}(\rho)$, where $\gamma$ is the unknown diffusion coefficient and $\mathcal{F}$ denotes the underlying nonlinear reaction functional. We use 8 additional candidate terms (e.g., $\{1, \rho, \rho^2, \rho^3, \rho_x, \rho\rho_x, \rho^2\rho_x, \rho^3\rho_x\}$) to reconstruct $\mathcal{F}$, whose coefficients are sparse. Hence, the total number of trainable coefficients remains 9 (e.g., $\boldsymbol{\Lambda}\in\mathbb{R}^{9\times1}$). Other details on the PiDL model setting and training can be found in \textcolor{blue}{\text{Supplementary} Note B.4}. 

\begin{figure}[t!]
\centering
\includegraphics[width=0.9\linewidth]{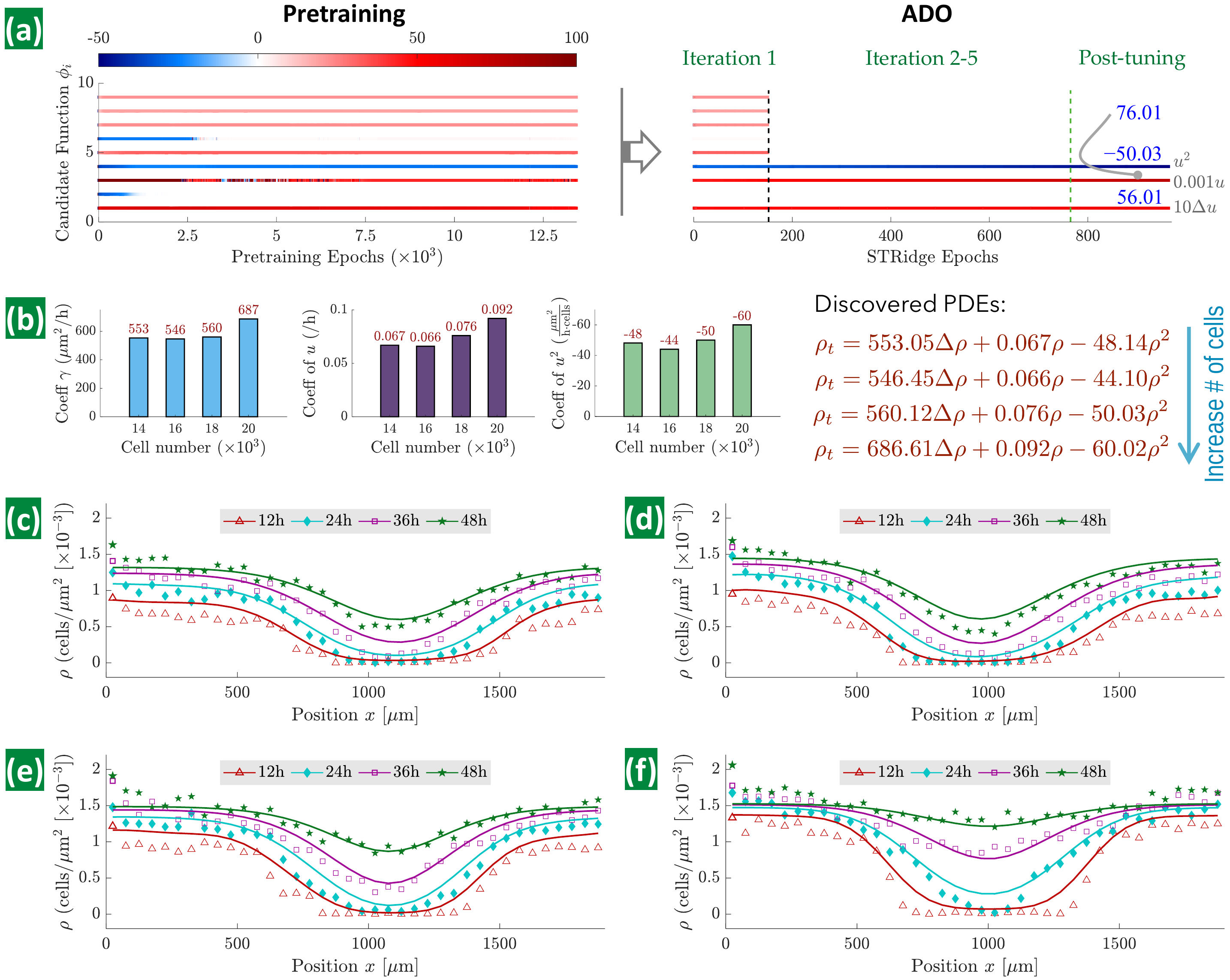}
\caption{Discovery result for cell migration and proliferation. (a) Example evolution of the sparse coefficients $\boldsymbol{\Lambda}\in\mathbb{R}^{9\times1}$ for 9 candidate functions used to construct the underlying PDE for the case of 18,000 cells. The diffusion and reaction coefficients for $\Delta u$ and $u$ are re-scaled for visualization purpose. (b) Discovered active terms $\{\Delta \rho, \rho, \rho^2\}$, their coefficients and the corresponding PDEs for 14,000, 16,000, 18,000 and 20,000 cells, respectively. (c)-(f) Simulated cell densities at different time instants based on the discovered PDEs for 14,000, 16,000, 18,000 and 20,000 cells, respectively, where the measurement at 0h is used as the initial condition while $\rho_x(x=0, t)=\rho_x(x=1900, t)=0$ is employed as the Neumann boundary condition. The simulation result is represented by solid curves while the markers denote the measurement data.}
\label{fig:Cell_result}
\end{figure}

Fig. \ref{fig:Cell_result}a shows the evolution of 9 coefficients for the example case of 18,000 cells, where redundant candidate terms are pruned right after the first ADO iteration via hard thresholding of the corresponding coefficients to zero. The next ADO iterations followed by post-tuning refine the coefficients of active terms for final reconstruction of the PDE. Fig. \ref{fig:Cell_result}b depicts the identified active term coefficients and the corresponding PDEs for different quantities of cells, sharing a unified form of $\rho_t = \gamma\rho_{xx} + \lambda_1\rho + \lambda_2\rho^2$ which exactly matches the famous Fisher-Kolmogorov model \cite{maini2004traveling}. The rates of migration (diffusion) and proliferation (reaction) generally increase along with the number of cells, as seen from the identified coefficients in Fig. \ref{fig:Cell_result}b. With the discovered PDEs, we simulate/predict the evolution of cell densities at different time instants (12h, 24h, 36h and 48h) presented in Fig. \ref{fig:Cell_result}c-f, where the measurement at 0h is used as the initial condition while $\rho_x(x=0, t)=\rho_x(x=1900, t)=0$ is employed as the Neumann boundary condition. The satisfactory agreement between the prediction and the measurement provides a clear validation of our discovered PDEs. It is noted that the extremely scarce and noisy experimental datasets unfortunately pose intractable challenge for any other existing methods (e.g., SINDy \cite{Brunton3932, Rudye1602614}) to produce a reasonable discovery. This experimental example further demonstrates the strength and capacity of the proposed methodology in regard to handling high level of data scarcity and noise for PDE discovery.

\section*{DISCUSSION}
In summary, we have presented a novel interpretable deep learning method for discovering physical laws, in particular parsimonious closed-form PDE(s), from scarce and noisy data (commonly seen in scientific investigations and real-world applications) for multi-dimensional nonlinear spatiotemporal systems. This approach combines the strengths of DNNs for rich representation learning of nonlinear functions, automatic differentiation for accurate derivative calculation as well as $\ell_0$ sparse regression to tackle the fundamental limitation faced by existing sparsity-promoting methods that scale poorly with respect to data noise and scarcity. The use of collocation points (having no correlation with the measurement data) can render the proposed framework tolerable to scarce and noisy measurements, making the DNN for PDE solution approximation generalizable (see \textcolor{blue}{\text{Supplementary} Note C2}). The special network architecture design is able to account for multiple independent datasets sampled under different initial/boundary conditions. An alternating direction optimization strategy is proposed to simultaneously train the DNN and determine the optimal sparse coefficients of selected candidate terms for reconstructing the PDE(s). The synergy of interpretable DNN and sparse PDE representation results in the following outcome: the DNN provides accurate modeling of the solution and its derivatives as a basis for constructing the governing equation(s), while the sparsely represented PDE(s) in turn informs and constraints the DNN which makes it generalizable and further enhances the discovery. The overall approach is rooted in a comprehensive integration of bottom-up (data-driven) and top-down (physics-informed) processes for scientific discovery, with fusion of physics-informed deep learning, sparse regression and optimization. We demonstrate this method on a number of dynamical systems exhibiting nonlinear spatiotemporal behaviors (e.g., chaotic, shock, propagating front, etc.) governed by multi-dimensional PDEs based on either single or multiple datasets, numerically or experimentally. Results highlight that the approach is capable of accurately discovering the exact form of the governing equation(s), even in an information-poor space where the multi-dimensional measurements are scarce and noisy. 

There still remain some potential limitations associated with the present PiDL framework for physical law discovery. For example, although the fully connected DNN used in this work has advantage of analytical approximation of the PDE derivatives via automatic differentiation, directly applying it to model the solution of higher dimensional systems (such as long/short-term response evolution in a 3D domain) results in computational bottleneck and optimization challenges, e.g., due to the need for a vast number of collocation points to maintain satisfactory accuracy. Advances in discrete DNNs with spatiotemporal discretization (e.g., the convolutional long short-term memory network (ConvLSTM) \cite{Shi2015} or similar) have the potential to help resolve this challenge, which will be demonstrated in our future work. In addition, the ``root-branch'' scheme might suffer from scalability issues when a large number of independent datasets sampled under various IBCs are available, resulting in many branches of the network for PDE solution approximation. The number of DNN trainable variables, the requirement of collocation points for retaining solution accuracy, and thus the computing memory, will grow in general linearly with the number of independent datasets (e.g., $\mathcal{O}(r)$). Nevertheless, this issue can be potentially well resolved by multi-GPU parallelization. Ideally, if the IBCs are known \textit{a priori} and can be parameterized, a parametric DNN learning scheme could be developed into the proposed PiDL for parametric PDE solution approximation that accounts for different IBCs \cite{SUN2020112732}. Several other aspects, such as the design of library of candidate functions and discovery with unknown source terms, are further discussed in \textcolor{blue}{\text{Supplementary} Note C1, C3, C4}.

\section*{METHOD}
The innovations of this work are built upon seamless integration of the strengths of deep neural networks for rich representation learning, physics embedding, automatic differentiation and sparse regression to (1) approximate the solution of system variables, (2) compute essential derivatives, as well as (3) identify the key derivative terms and parameters that form the structure and explicit expression of the PDE(s). The resulting approach is able to deal with scarce/sparse and highly noisy measurement data while accounting for different initial/boundary conditions. The key method components are discussed below.

\subsection*{Network Architecture}
The proposed network architectures of PiDL with sparse regression are shown in Figs. \ref{fig:PINN}a and \ref{fig:PINN}b that respectively deal with single-IBC dataset and multiple-IBC ($r$) independent datasets. The latent solution $\mathbf{u}$ is interpreted by a dense (fully connected) DNN shown in Fig. \ref{fig:PINN}a, namely, $\mathbf{u}^\theta = \mathbf{u}(\mathbf{x}, t; \boldsymbol{\theta})$, for the case of single dataset, while a ``root-branch'' dense DNN depicted in Fig. \ref{fig:PINN}b is designed to approximate the latent solutions $\mathbf{u}_i$ ($i=1, ..., r$) corresponding to different IBCs, viz., $\mathbf{u}^\theta_i =\mathbf{u}\big(\mathbf{x}, t; \boldsymbol{\theta}^{(0)}, \boldsymbol{\theta}^{(i)}\big)$, for multiple independent datasets. Here, $\boldsymbol{\theta}$'s denote the DNN trainable parameters. The DNNs take the spatiotemporal domain coordinates $\{\mathbf{x}, t\}$ as input followed by multiple fully-connected feedforward hidden layers (each layer has dozens of nodes). We use the hyperbolic tangent (tanh) or sine (sin) as the universal activation function thanks to their strength for high-order differentiation and unbiased estimation for both positive and negative values. The sin function is used when the system response exhibits periodic patterns. The output later is based on linear activation for universal magnitude mapping. When multiple datasets are available, e.g. sampled from different IBCs, domain coordinates are input to the ``root'' net (shared hidden layers), followed by $r$ ``branch'' nets (individual hidden layers) that predict system response corresponding to each IBC/dataset. The ``root'' learns the common patterns across all datasets while the ``branches'' learn specific details determined by each IBC for each independent dataset. Such an architecture integrates information from different measurements at the expense of larger computational efforts and produces solution approximations satisfying a unified physics (e.g., governing PDE(s)). The DNNs essentially play a role as a nonlinear functional to approximate the latent solution.

The DNN is connected to the physical law (reconstruction of PDE(s)) through a graph-based automatic differentiator where derivatives on $\mathbf{u}$'s are evaluated at machine precision. The library of candidate functions $\boldsymbol{\phi}^\theta$ can be computed from the DNNs. For the case of multiple independent datasets, the libraries $\boldsymbol{\phi}^{(i)}$ resulted from the ``branch'' nets are concatenated to build one unified $\boldsymbol{\phi}^\theta$. If there is unknown source input, the candidate functions for $\mathbf{p}$ can also be incorporated into the library for discovery. The sparse representation of the reconstructed PDE(s) is then expressed in a residual form: $\boldsymbol{\mathcal{R}}^{\theta}:=\mathbf{u}_t^{\theta} - \boldsymbol{\phi}^{\theta}\boldsymbol{\Lambda} \rightarrow \mathbf{0} \text{~s.t.~} \boldsymbol{\Lambda}\in\mathcal{S}$ , where $\boldsymbol{\mathcal{R}}^{\theta}\in \mathbb{R}^{1\times n}$ denotes the PDE residuals, $\mathcal{S}$ represents the sparsity constraint set, and $n$ is the dimension of the system variable (e.g., $\mathbf{u}\in\mathbb{R}^{1\times n}$). Thus, the overall network architecture consists of heterogeneous trainable variables, namely, DNN parameters $\boldsymbol{\theta}\in\mathbb{R}^{n_\theta\times1}$ and PDE coefficients $\boldsymbol{\Lambda}\in\mathcal{S}\subset\mathbb{R}^{s\times n}$, where $n_\theta$ denotes the number of DNN trainable parameters and $n_\theta\gg sn$.

\subsection*{Physics-constrained Sparsity-regularized Loss Function}
The physics-constrained sparsity-regularized loss function, expressed in Eq. \eref{eq:totalloss}, is composed of three components, the data loss $\mathcal{L}_d$, the residual physics loss $\mathcal{L}_p$ and a sparsity regularization term imposed on $\boldsymbol{\Lambda}$. The data loss function reads
\begin{equation}\label{eq:dataloss}
    \mathcal{L}_d(\boldsymbol{\theta}; \mathcal{D}_u) = \frac{1}{N_m} \big \|\mathbf{u}^\theta - \mathbf{u}^m \big\|_2^2
\end{equation}
where $\mathbf{u}^m$ is the measurement data, $\mathbf{u}^\theta$ is the corresponding DNN-approximated solution, $N_m$ is the total number of data points, and $\|\cdot\|_2$ denotes the Frobenius norm. The responses are stacked when multiple datasets are available, e.g., $\mathbf{u}^m = \{\mathbf{u}^m_1, ..., \mathbf{u}^m_r\}$ and $\mathbf{u}^\theta = \{\mathbf{u}^\theta_1, ..., \mathbf{u}^\theta_r\}$, where $r\geq2$, as shown in Fig. \ref{fig:PINN}b. The PDE residuals $\boldsymbol{\mathcal{R}}^{\theta}$ are evaluated on a large number of randomly sampled collocation points $\mathcal{D}_c$, and used to form the residual physics loss function given by
\begin{equation}\label{eq:residualloss}
    \mathcal{L}_p(\boldsymbol{\theta}, \boldsymbol{\Lambda}; \mathcal{D}_c) = \frac{1}{N_c} \left \|\dot{\mathbf{U}}(\boldsymbol{\theta}) - \boldsymbol{\Phi}(\boldsymbol{\theta})\boldsymbol{\Lambda} \right\|_2^2
\end{equation}
where $\dot{\mathbf{U}}$ and $\boldsymbol{\Phi}$ denote respectively the discretization of the first-order time derivative term and the library of candidate functions evaluated on the collocation points; $N_c$ is the total number of spatiotemporal collocation points. For the case of multiple datasets, $\dot{\mathbf{U}}$ and $\boldsymbol{\Phi}$ are concatenated over the index of different IBCs to ensure the identical physical law (in particular, the governing PDE(s)) is imposed, as depicted in Fig. \ref{fig:PINN}b. Note that $\mathcal{L}_d$ ensures that the DNN accurately interpret the latent solution of the PDE(s) via fitting the data, while $\mathcal{L}_p$ generalizes and provides constraints for the DNN through reconstructing the closed form of the PDE(s). The $\ell_0$ regularization term in Eq. \eref{eq:totalloss} promotes the sparsity of the coefficients $\boldsymbol{\Lambda}$ for sparse representation of the PDE(s).

\subsection*{Alternating Direction Optimization}
The total loss function in Eq. \eref{eq:totalloss} has an implicit complex form, and thus, directly solving the optimization problem is highly intractable since the $\ell_0$ regularization makes this problem $np$-hard. Though relaxation of the $\ell_0$ term by the less rigorous $\ell_1$ regularization improves the well-posedness and enables the optimization in a continuous space, false positive identification occurs \cite{BERG2019239, Both2019}. To address this challenge, we present an alternating direction optimization (ADO) algorithm that divides the overall optimization problem into a set of tractable subproblems to sequentially optimize $\boldsymbol{\theta}$ and $\boldsymbol{\Lambda}$ within a few alternating iterations (denoted by $k$), namely,
\begin{subequations}
\begin{align}
\boldsymbol{\Lambda}^\star_{k+1} &:= \operatorname*{arg\,min}_{\mathbf{\Lambda}} \left[ \left \|\dot{\mathbf{U}}(\boldsymbol{\theta}^\star_k) - \boldsymbol{\Phi}(\boldsymbol{\theta}^\star_k)\boldsymbol{\Lambda} \right\|_2^2 + \beta{\|\boldsymbol{\Lambda}\|}_{0} \right] \label{eq:ADO1}\\
\boldsymbol{\theta}^\star_{k+1} &:= \operatorname*{arg\,min}_{\boldsymbol{\theta}} \left[\mathcal{L}_d(\boldsymbol{\theta}; \mathcal{D}_u) + \alpha\mathcal{L}_p(\boldsymbol{\theta}, \boldsymbol{\Lambda}^\star_{k+1}; \mathcal{D}_c)\right]\label{eq:ADO2} 
\end{align}
\end{subequations}
The fundamental concept of the ADO algorithm shares similarity with the alternating direction methods of multipliers \cite{boyd2011}. In each alternating iteration $k+1$, the sparse PDE coefficients $\boldsymbol{\Lambda}$ in Eq. \eref{eq:ADO1} are updated (denoted by $\boldsymbol{\Lambda}^\star_{k+1}$) via STRidge (a sequential thresholding regression process that serves as a proxy for $\ell_0$ regularization \cite{Brunton3932, Rudye1602614}), based on the DNN parameters from the previous iteration (e.g., $\boldsymbol{\theta}^\star_k$). The DNN parameters $\boldsymbol{\theta}$ in the current iteration are then updated (denoted by $\boldsymbol{\theta}^\star_{k+1}$) through a standard neural network training algorithm (in particular, the combined Adam \cite{adam2014} + L-BFGS \cite{Byrd1995} optimizer), taking $\boldsymbol{\Lambda}^\star_{k+1}$ as known. The alternations between the sub-optimal solutions will lead to a high-quality optimization solution. It is noteworthy that the Adam optimizer plays a role for global search while the L-BFGS optimizer takes responsibility of fine tuning in a local solution region. The learning rate of Adam ranges from $10^{-5}$ to $10^{-3}$ in the test examples. The algorithm design of ADO, the choice of hyperparameters (e.g., the relative weighting of the loss functions, $\alpha$ and $\beta$), as well as the implementation details and specifications are given in \textcolor{blue}{\text{Supplementary} Algorithm 1} and \textcolor{blue}{Algorithm 2}.

Pre-training of PiDL is conducted before running the ADO algorithm for discovery, by simply replacing $\|\boldsymbol{\Lambda}\|_0$ in Eq. \eref{eq:totalloss} with $\|\boldsymbol{\Lambda}\|_1$ where brute-force gradient-based optimization (e.g., Adam + L-BFGS) for both $\boldsymbol{\theta}$ and $\boldsymbol{\Lambda}$ becomes applicable, namely,
\begin{equation}
\{\boldsymbol{\theta}^\star, \boldsymbol{\Lambda}^\star\} = \mathop{\arg\min}_{\{\boldsymbol{\theta}, \mathbf{\Lambda}\}} \left\{\mathcal{L}_d(\boldsymbol{\theta}; \mathcal{D}_u) + \alpha\mathcal{L}_p(\boldsymbol{\theta}, \boldsymbol{\Lambda}; \mathcal{D}_c) + \beta\|\boldsymbol{\Lambda}\|_1 \right\}
\end{equation}
The $\ell_1$-regularized pre-training can accelerate the convergence of ADO by providing an admissible ``initial guess''. Post-training (or post-tuning) is also applicable, which can be applied after the closed form of the PDE(s) is uncovered. This can be done by training the DNN along with the identification of the discovered non-zero coefficients, viz., 
\begin{equation}
\{\boldsymbol{\theta}^\star, \boldsymbol{\Lambda}^\star\} = \mathop{\arg\min}_{\{\boldsymbol{\theta}, \mathbf{\Lambda}\}} \left\{\mathcal{L}_d(\boldsymbol{\theta}; \mathcal{D}_u) + \alpha\mathcal{L}_p(\boldsymbol{\theta}, \boldsymbol{\Lambda}; \mathcal{D}_c) \right\}
\end{equation}
where the initialization of the unknown parameters $\{\boldsymbol{\theta}, \mathbf{\Lambda}\}$ can be inherited from the ADO result. The post-training step is completely optional since the ADO method can already provides a high-quality solution as shown in the test examples. Nevertheless, the post-training could add additional discovery accuracy through fine tuning.

\section*{Data availability} 
All the used datasets in this study are available on GitHub at \url{https://github.com/isds-neu/EQDiscovery} upon final publication.

\section*{Code availability} 
All the source codes to reproduce the results in this study are available on GitHub at \url{https://github.com/isds-neu/EQDiscovery} upon final publication.

\bibliographystyle{unsrt}
\bibliography{references}

\begin{thebibliography}{10}

\bibitem{Bongard9943}
Josh Bongard and Hod Lipson.
\newblock Automated reverse engineering of nonlinear dynamical systems.
\newblock {\em Proceedings of the National Academy of Sciences},
  104(24):9943--9948, 2007.

\bibitem{Schmidt2009}
Michael~D. Schmidt and Hod Lipson.
\newblock Distilling free-form natural laws from experimental data.
\newblock {\em Science}, 324 5923:81--5, 2009.

\bibitem{Schaeffer6634}
Hayden Schaeffer, Russel Caflisch, Cory~D. Hauck, and Stanley Osher.
\newblock Sparse dynamics for partial differential equations.
\newblock {\em Proceedings of the National Academy of Sciences},
  110(17):6634--6639, 2013.

\bibitem{Daniels2015}
Bryan~C. Daniels and Ilya Nemenman.
\newblock Automated adaptive inference of phenomenological dynamical models.
\newblock {\em Nature Communications}, 6:8133, 2015.

\bibitem{Brunton3932}
Steven~L. Brunton, Joshua~L. Proctor, and J.~Nathan Kutz.
\newblock Discovering governing equations from data by sparse identification of
  nonlinear dynamical systems.
\newblock {\em Proceedings of the National Academy of Sciences},
  113(15):3932--3937, 2016.

\bibitem{Rudye1602614}
Samuel~H. Rudy, Steven~L. Brunton, Joshua~L. Proctor, and J.~Nathan Kutz.
\newblock Data-driven discovery of partial differential equations.
\newblock {\em Science Advances}, 3(4):e1602614, 2017.

\bibitem{Schaeffer2017}
Hayden Schaeffer.
\newblock Learning partial differential equations via data discovery and sparse
  optimization.
\newblock {\em Proceedings of the Royal Society A: Mathematical, Physical and
  Engineering Sciences}, 473(2197):20160446, 2017.

\bibitem{lusch2018deep}
Bethany Lusch, J~Nathan Kutz, and Steven~L Brunton.
\newblock Deep learning for universal linear embeddings of nonlinear dynamics.
\newblock {\em Nature Communications}, 9(1):1--10, 2018.

\bibitem{WANG201944}
Z.~Wang, X.~Huan, and K.~Garikipati.
\newblock Variational system identification of the partial differential
  equations governing the physics of pattern-formation: Inference under varying
  fidelity and noise.
\newblock {\em Computer Methods in Applied Mechanics and Engineering}, 356:44
  -- 74, 2019.

\bibitem{Champion22445}
Kathleen Champion, Bethany Lusch, J.~Nathan Kutz, and Steven~L. Brunton.
\newblock Data-driven discovery of coordinates and governing equations.
\newblock {\em Proceedings of the National Academy of Sciences},
  116(45):22445--22451, 2019.

\bibitem{Pfister25405}
Niklas Pfister, Stefan Bauer, and Jonas Peters.
\newblock Learning stable and predictive structures in kinetic systems.
\newblock {\em Proceedings of the National Academy of Sciences},
  116(51):25405--25411, 2019.

\bibitem{HUANG2020JMPS}
Zhilong Huang, Yanping Tian, Chunjiang Li, Guang Lin, Lingling Wu, Yong Wang,
  and Hanqing Jiang.
\newblock Data-driven automated discovery of variational laws hidden in
  physical systems.
\newblock {\em Journal of the Mechanics and Physics of Solids}, 137:103871,
  2020.

\bibitem{loiseau2018constrained}
Jean-Christophe Loiseau and Steven~L Brunton.
\newblock Constrained sparse galerkin regression.
\newblock {\em Journal of Fluid Mechanics}, 838:42--67, 2018.

\bibitem{loiseau2018sparse}
Jean-Christophe Loiseau, Bernd~R Noack, and Steven~L Brunton.
\newblock Sparse reduced-order modelling: sensor-based dynamics to full-state
  estimation.
\newblock {\em Journal of Fluid Mechanics}, 844:459--490, 2018.

\bibitem{LAI2019813}
Zhilu Lai and Satish Nagarajaiah.
\newblock Sparse structural system identification method for nonlinear dynamic
  systems with hysteresis/inelastic behavior.
\newblock {\em Mechanical Systems and Signal Processing}, 117:813 -- 842, 2019.

\bibitem{Li-PRE2220}
Shanwu Li, Eurika Kaiser, Shujin Laima, Hui Li, Steven~L. Brunton, and
  J.~Nathan Kutz.
\newblock Discovering time-varying aerodynamics of a prototype bridge by sparse
  identification of nonlinear dynamical systems.
\newblock {\em Physics Review E}, 100:022220, Aug 2019.

\bibitem{Mangan2016}
N.~M. {Mangan}, S.~L. {Brunton}, J.~L. {Proctor}, and J.~N. {Kutz}.
\newblock Inferring biological networks by sparse identification of nonlinear
  dynamics.
\newblock {\em IEEE Transactions on Molecular, Biological and Multi-Scale
  Communications}, 2(1):52--63, 2016.

\bibitem{hoffmann2019reactive}
Moritz Hoffmann, Christoph Fr{\"o}hner, and Frank No{\'e}.
\newblock Reactive {SINDy: Discovering} governing reactions from concentration
  data.
\newblock {\em The Journal of chemical physics}, 150(2):025101, 2019.

\bibitem{BHADRIRAJU2019372}
Bhavana Bhadriraju, Abhinav Narasingam, and Joseph Sang-Il Kwon.
\newblock Machine learning-based adaptive model identification of systems:
  Application to a chemical process.
\newblock {\em Chemical Engineering Research and Design}, 152:372--383, 2019.

\bibitem{cichos2020machine}
Frank Cichos, Kristian Gustavsson, Bernhard Mehlig, and Giovanni Volpe.
\newblock Machine learning for active matter.
\newblock {\em Nature Machine Intelligence}, 2(2):94--103, 2020.

\bibitem{Kaiser2018}
E.~Kaiser, J.~N. Kutz, and S.~L. Brunton.
\newblock Sparse identification of nonlinear dynamics for model predictive
  control in the low-data limit.
\newblock {\em Proceedings of the Royal Society A: Mathematical, Physical and
  Engineering Sciences}, 474(2219):20180335, 2018.

\bibitem{Champion2019SIAM}
Kathleen~P. Champion, Steven~L. Brunton, and J.~Nathan Kutz.
\newblock Discovery of nonlinear multiscale systems: Sampling strategies and
  embeddings.
\newblock {\em SIAM Journal on Applied Dynamical Systems}, 18(1):312--333,
  2019.

\bibitem{Dam2017}
Magnus Dam, Morten Br{\o}ns, Jens Juul~Rasmussen, Volker Naulin, and Jan~S.
  Hesthaven.
\newblock Sparse identification of a predator-prey system from simulation data
  of a convection model.
\newblock {\em Physics of Plasmas}, 24(2):022310, 2017.

\bibitem{Boninsegna2018}
Lorenzo Boninsegna, Feliks Nuske, and Cecilia Clementi.
\newblock Sparse learning of stochastic dynamical equations.
\newblock {\em The Journal of Chemical Physics}, 148(24):241723, 2018.

\bibitem{Kaheman2020Implicit}
Kadierdan Kaheman, J~Nathan Kutz, and Steven~L Brunton.
\newblock {SINDy-PI: A} robust algorithm for parallel implicit sparse
  identification of nonlinear dynamics.
\newblock {\em arXiv preprint arXiv:2004.02322}, 2020.

\bibitem{schaeffer2018extracting}
Hayden Schaeffer, Giang Tran, and Rachel Ward.
\newblock Extracting sparse high-dimensional dynamics from limited data.
\newblock {\em SIAM Journal on Applied Mathematics}, 78(6):3279--3295, 2018.

\bibitem{zhang2019convergence}
Linan Zhang and Hayden Schaeffer.
\newblock On the convergence of the {SINDy} algorithm.
\newblock {\em Multiscale Modeling \& Simulation}, 17(3):948--972, 2019.

\bibitem{Rudy2019SIAM}
Samuel Rudy, Alessandro Alla, Steven~L. Brunton, and J.~Nathan Kutz.
\newblock Data-driven identification of parametric partial differential
  equations.
\newblock {\em SIAM Journal on Applied Dynamical Systems}, 18(2):643--660,
  2019.

\bibitem{ZhangPRS2018}
Sheng Zhang and Guang Lin.
\newblock Robust data-driven discovery of governing physical laws with error
  bars.
\newblock {\em Proceedings of the Royal Society A: Mathematical, Physical and
  Engineering Sciences}, 474(2217):20180305, 2018.

\bibitem{Vaddireddy2020}
Harsha Vaddireddy, Adil Rasheed, Anne~E. Staples, and Omer San.
\newblock Feature engineering and symbolic regression methods for detecting
  hidden physics from sparse sensor observation data.
\newblock {\em Physics of Fluids}, 32(1):015113, 2020.

\bibitem{zhang_ma_2020}
Jun Zhang and Wenjun Ma.
\newblock Data-driven discovery of governing equations for fluid dynamics based
  on molecular simulation.
\newblock {\em Journal of Fluid Mechanics}, 892:A5, 2020.

\bibitem{lagergren2020learning}
John~H Lagergren, John~T Nardini, G~Michael~Lavigne, Erica~M Rutter, and
  Kevin~B Flores.
\newblock Learning partial differential equations for biological transport
  models from noisy spatio-temporal data.
\newblock {\em Proceedings of the Royal Society A}, 476(2234):20190800, 2020.

\bibitem{Gurevich2019}
Daniel~R. Gurevich, Patrick A.~K. Reinbold, and Roman~O. Grigoriev.
\newblock Robust and optimal sparse regression for nonlinear {PDE} models.
\newblock {\em Chaos: An Interdisciplinary Journal of Nonlinear Science},
  29(10):103113, 2019.

\bibitem{baydin2017automatic}
At{\i}l{\i}m~G{\"u}nes Baydin, Barak~A Pearlmutter, Alexey~Andreyevich Radul,
  and Jeffrey~Mark Siskind.
\newblock Automatic differentiation in machine learning: a survey.
\newblock {\em The Journal of Machine Learning Research}, 18(1):5595--5637,
  2017.

\bibitem{Han8505}
Jiequn Han, Arnulf Jentzen, and Weinan E.
\newblock Solving high-dimensional partial differential equations using deep
  learning.
\newblock {\em Proceedings of the National Academy of Sciences},
  115(34):8505--8510, 2018.

\bibitem{SIRIGNANO20181339}
Justin Sirignano and Konstantinos Spiliopoulos.
\newblock {DGM}: A deep learning algorithm for solving partial differential
  equations.
\newblock {\em Journal of Computational Physics}, 375:1339--1364, 2018.

\bibitem{RAISSI2019686}
M.~Raissi, P.~Perdikaris, and G.E. Karniadakis.
\newblock Physics-informed neural networks: A deep learning framework for
  solving forward and inverse problems involving nonlinear partial differential
  equations.
\newblock {\em Journal of Computational Physics}, 378:686--707, 2019.

\bibitem{YANG2019136}
Yibo Yang and Paris Perdikaris.
\newblock Adversarial uncertainty quantification in physics-informed neural
  networks.
\newblock {\em Journal of Computational Physics}, 394:136--152, 2019.

\bibitem{Bar-Sinai15344}
Yohai Bar-Sinai, Stephan Hoyer, Jason Hickey, and Michael~P. Brenner.
\newblock Learning data-driven discretizations for partial differential
  equations.
\newblock {\em Proceedings of the National Academy of Sciences},
  116(31):15344--15349, 2019.

\bibitem{SUN2020112732}
Luning Sun, Han Gao, Shaowu Pan, and Jian-Xun Wang.
\newblock Surrogate modeling for fluid flows based on physics-constrained deep
  learning without simulation data.
\newblock {\em Computer Methods in Applied Mechanics and Engineering},
  361:112732, 2020.

\bibitem{Raissi1026}
Maziar Raissi, Alireza Yazdani, and George~Em Karniadakis.
\newblock Hidden fluid mechanics: Learning velocity and pressure fields from
  flow visualizations.
\newblock {\em Science}, 367(6481):1026--1030, 2020.

\bibitem{BERG2019239}
Jens Berg and Kaj Nystr\.
\newblock Data-driven discovery of pdes in complex datasets.
\newblock {\em Journal of Computational Physics}, 384:239--252, 2019.

\bibitem{Both2019}
Gert-Jan Both, Subham Choudhury, Pierre Sens, and Remy Kusters.
\newblock Deepmod: Deep learning for model discovery in noisy data.
\newblock {\em Journal of Computational Physics}, page 109985, 2020.

\bibitem{jin2016reproducibility}
Wang Jin, Esha~T Shah, Catherine~J Penington, Scott~W McCue, Lisa~K Chopin, and
  Matthew~J Simpson.
\newblock Reproducibility of scratch assays is affected by the initial degree
  of confluence: experiments, modelling and model selection.
\newblock {\em Journal of Theoretical Biology}, 390:136--145, 2016.

\bibitem{maini2004traveling}
Philip~K Maini, DL~Sean McElwain, and David~I Leavesley.
\newblock Traveling wave model to interpret a wound-healing cell migration
  assay for human peritoneal mesothelial cells.
\newblock {\em Tissue Engineering}, 10(3-4):475--482, 2004.

\bibitem{Shi2015}
Shi Xingjian, Zhourong Chen, Hao Wang, Dit-Yan Yeung, Wai-Kin Wong, and
  Wang-chun Woo.
\newblock Convolutional lstm network: A machine learning approach for
  precipitation nowcasting.
\newblock In {\em Advances in Neural Information Processing Systems}, pages
  802--810, 2015.

\bibitem{boyd2011}
Stephen Boyd, Neal Parikh, Eric Chu, Borja Peleato, and Jonathan Eckstein.
\newblock Distributed optimization and statistical learning via the alternating
  direction method of multipliers.
\newblock {\em Foundations and Trends{\textregistered} in Machine learning},
  3(1):1--122, 2011.

\bibitem{adam2014}
Diederik~P. Kingma and Jimmy Ba.
\newblock Adam: {A} method for stochastic optimization.
\newblock In {\em 3rd International Conference on Learning Representations
  ({ICLR})}, 2015.

\bibitem{Byrd1995}
R.~Byrd, P.~Lu, J.~Nocedal, and C.~Zhu.
\newblock A limited memory algorithm for bound constrained optimization.
\newblock {\em SIAM Journal on Scientific Computing}, 16(5):1190--1208, 1995.

\end{thebibliography}


\begin{thebibliography}{10}

\bibitem{Brunton3932}
Steven~L. Brunton, Joshua~L. Proctor, and J.~Nathan Kutz.
\newblock Discovering governing equations from data by sparse identification of
  nonlinear dynamical systems.
\newblock {\em Proceedings of the National Academy of Sciences},
  113(15):3932--3937, 2016.

\bibitem{Rudye1602614}
Samuel~H. Rudy, Steven~L. Brunton, Joshua~L. Proctor, and J.~Nathan Kutz.
\newblock Data-driven discovery of partial differential equations.
\newblock {\em Science Advances}, 3(4):e1602614, 2017.

\bibitem{Sobol}
I.M Sobol.
\newblock On the distribution of points in a cube and the approximate
  evaluation of integrals.
\newblock {\em USSR Computational Mathematics and Mathematical Physics},
  7(4):86--112, 1967.

\bibitem{Schaeffer2017}
Hayden Schaeffer.
\newblock Learning partial differential equations via data discovery and sparse
  optimization.
\newblock {\em Proceedings of the Royal Society A: Mathematical, Physical and
  Engineering Sciences}, 473(2197):20160446, 2017.

\bibitem{TAIRA20072118}
Kunihiko Taira and Tim Colonius.
\newblock The immersed boundary method: A projection approach.
\newblock {\em Journal of Computational Physics}, 225(2):2118--2137, 2007.

\bibitem{GUPTA2009399}
Ankur Gupta and Saikat Chakraborty.
\newblock Linear stability analysis of high- and low-dimensional models for
  describing mixing-limited pattern formation in homogeneous autocatalytic
  reactors.
\newblock {\em Chemical Engineering Journal}, 145(3):399--411, 2009.

\bibitem{harrison1994kinetic}
Lionel~G Harrison.
\newblock Kinetic theory of living pattern.
\newblock {\em Endeavour}, 18(4):130--136, 1994.

\bibitem{holmes1994partial}
Elizabeth~E Holmes, Mark~A Lewis, JE~Banks, and RR~Veit.
\newblock Partial differential equations in ecology: spatial interactions and
  population dynamics.
\newblock {\em Ecology}, 75(1):17--29, 1994.

\bibitem{McKay}
M.~D. McKay, R.~J. Beckman, and W.~J. Conover.
\newblock A comparison of three methods for selecting values of input variables
  in the analysis of output from a computer code.
\newblock {\em Technometrics}, 21(2):239--245, 1979.

\bibitem{fitzhugh1961impulses}
Richard FitzHugh.
\newblock Impulses and physiological states in theoretical models of nerve
  membrane.
\newblock {\em Biophysical journal}, 1(6):445, 1961.

\bibitem{nagumo1962active}
Jinichi Nagumo, Suguru Arimoto, and Shuji Yoshizawa.
\newblock An active pulse transmission line simulating nerve axon.
\newblock {\em Proceedings of the IRE}, 50(10):2061--2070, 1962.

\bibitem{jin2016reproducibility}
Wang Jin, Esha~T Shah, Catherine~J Penington, Scott~W McCue, Lisa~K Chopin, and
  Matthew~J Simpson.
\newblock Reproducibility of scratch assays is affected by the initial degree
  of confluence: experiments, modelling and model selection.
\newblock {\em Journal of theoretical biology}, 390:136--145, 2016.

\bibitem{fisher1937wave}
Ronald~Aylmer Fisher.
\newblock The wave of advance of advantageous genes.
\newblock {\em Annals of Eugenics}, 7(4):355--369, 1937.

\bibitem{maini2004traveling}
Philip~K Maini, DL~Sean McElwain, and David~I Leavesley.
\newblock Traveling wave model to interpret a wound-healing cell migration
  assay for human peritoneal mesothelial cells.
\newblock {\em Tissue Engineering}, 10(3-4):475--482, 2004.

\bibitem{WU2019200}
Kailiang Wu and Dongbin Xiu.
\newblock Numerical aspects for approximating governing equations using data.
\newblock {\em Journal of Computational Physics}, 384:200--221, 2019.

\bibitem{Daniels2015}
Bryan~C. Daniels and Ilya Nemenman.
\newblock Automated adaptive inference of phenomenological dynamical models.
\newblock {\em Nature Communications}, 6:8133, 2015.

\bibitem{Manohar2018}
Krithika {Manohar}, J.~Nathan {Kutz}, and Steven~L. {Brunton}.
\newblock {Optimal Sensor and Actuator Selection using Balanced Model
  Reduction}.
\newblock {\em arXiv e-prints}, page arXiv:1812.01574, December 2018.

\bibitem{Shi2015}
Shi Xingjian, Zhourong Chen, Hao Wang, Dit-Yan Yeung, Wai-Kin Wong, and
  Wang-chun Woo.
\newblock Convolutional lstm network: A machine learning approach for
  precipitation nowcasting.
\newblock In {\em Advances in neural information processing systems}, pages
  802--810, 2015.

\end{thebibliography}

\vspace{36pt}
\noindent\textbf{Acknowledgement:}
We acknowledge the support by the Engineering for Civil Infrastructure program at National Science Foundation under grant CMMI-2013067, the research award from MathWorks, and the Tier 1 Seed Grant Program at Northeastern University.\\

\noindent\textbf{Competing interests:}
The authors declare no competing interests.\\

\noindent\textbf{Supplementary information:}
The supplementary information is attached.

\clearpage
\setcounter{figure}{0}
\renewcommand{\figurename}{Extended Data Fig.}

\begin{figure}[htbp]
\centering
\includegraphics[width = 0.75\textwidth]{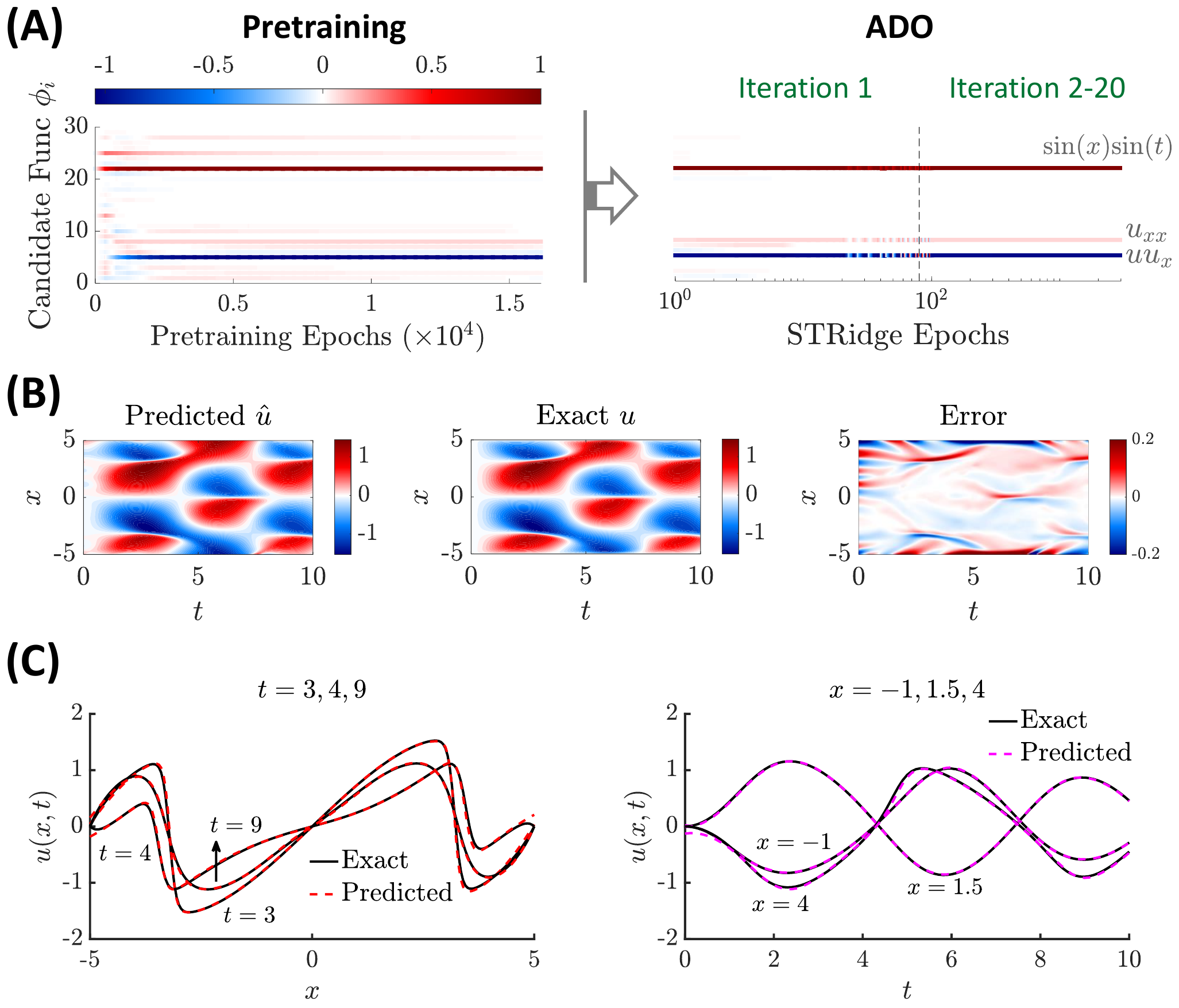}
\caption{Discovered Burgers' equation and source term for measurement data with 10\% noise. (a) Evolution of the sparse coefficients $\boldsymbol{\Lambda}\in\mathbb{R}^{30\times1}$ for 30 candidate functions $\boldsymbol{\phi}\in\mathbb{R}^{1\times30}$ used to form the PDE and the unknown source term, where the color represents the coefficient value. (b) The predicted response in comparison with the exact solution with the prediction error. The relative full-field $\ell_2$ error of the prediction is 13.8\%. The major errors are mostly distributed close to the boundaries due to scarce training data.}
\label{Fig:Source}
\end{figure}

\clearpage
\begin{figure}[htbp]
\centering
\includegraphics[width=0.995\linewidth]{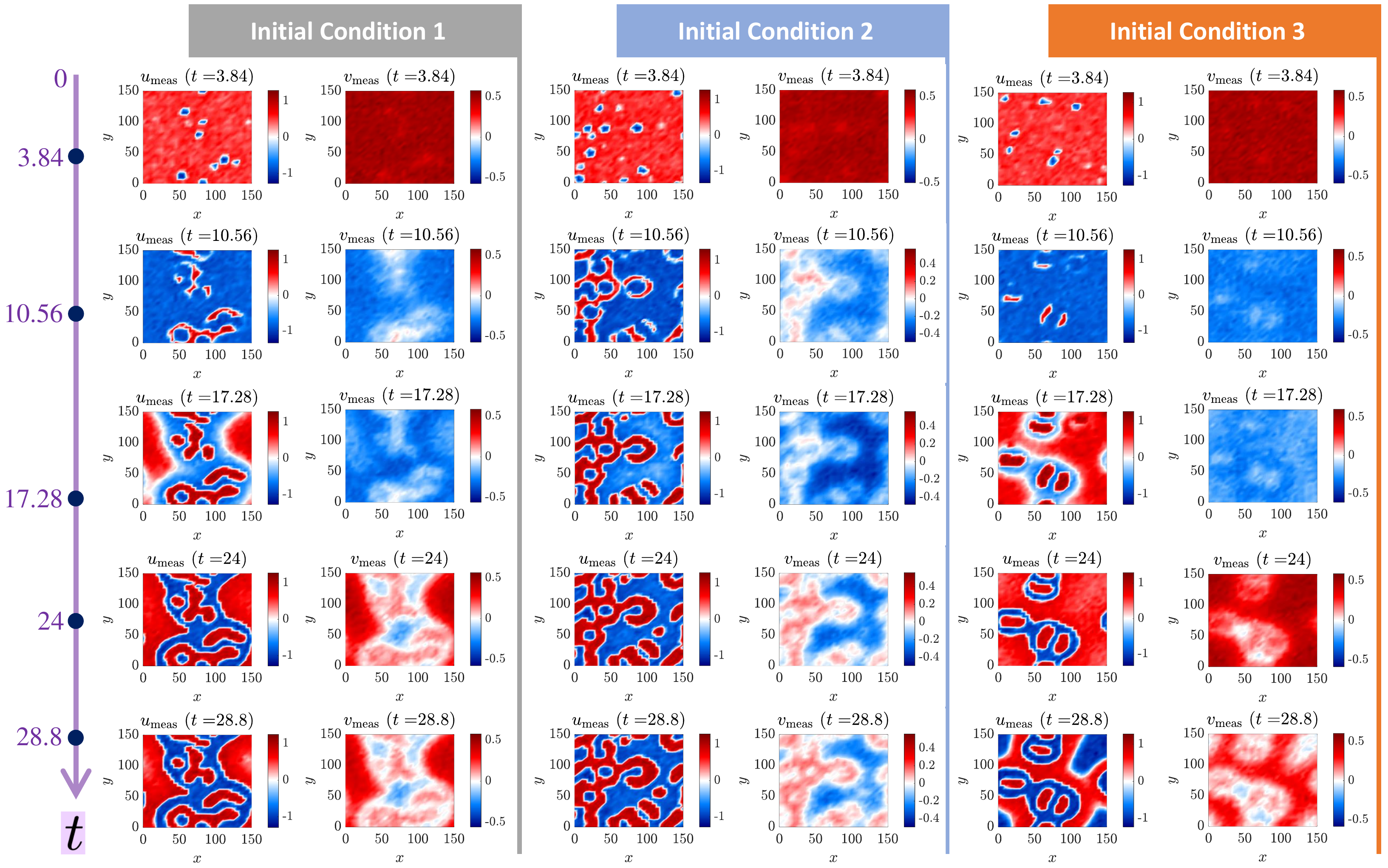}
\caption{A few typical snapshots of low-resolution noisy measurements (10\% noise) sampled from the system response under three different initial conditions (ICs) for discovering Fitzhugh-Nagumo equations. Note that the measurement data consists of 31 low-resolution noisy snapshots (with a grid size of $31\times31$) for each IC uniformly sampled within the time range of [0 28.8].}
\label{fig:FN_MultiIC_Meas}
\end{figure}

\clearpage
\begin{figure}[htbp]
\centering
\includegraphics[width=0.995\linewidth]{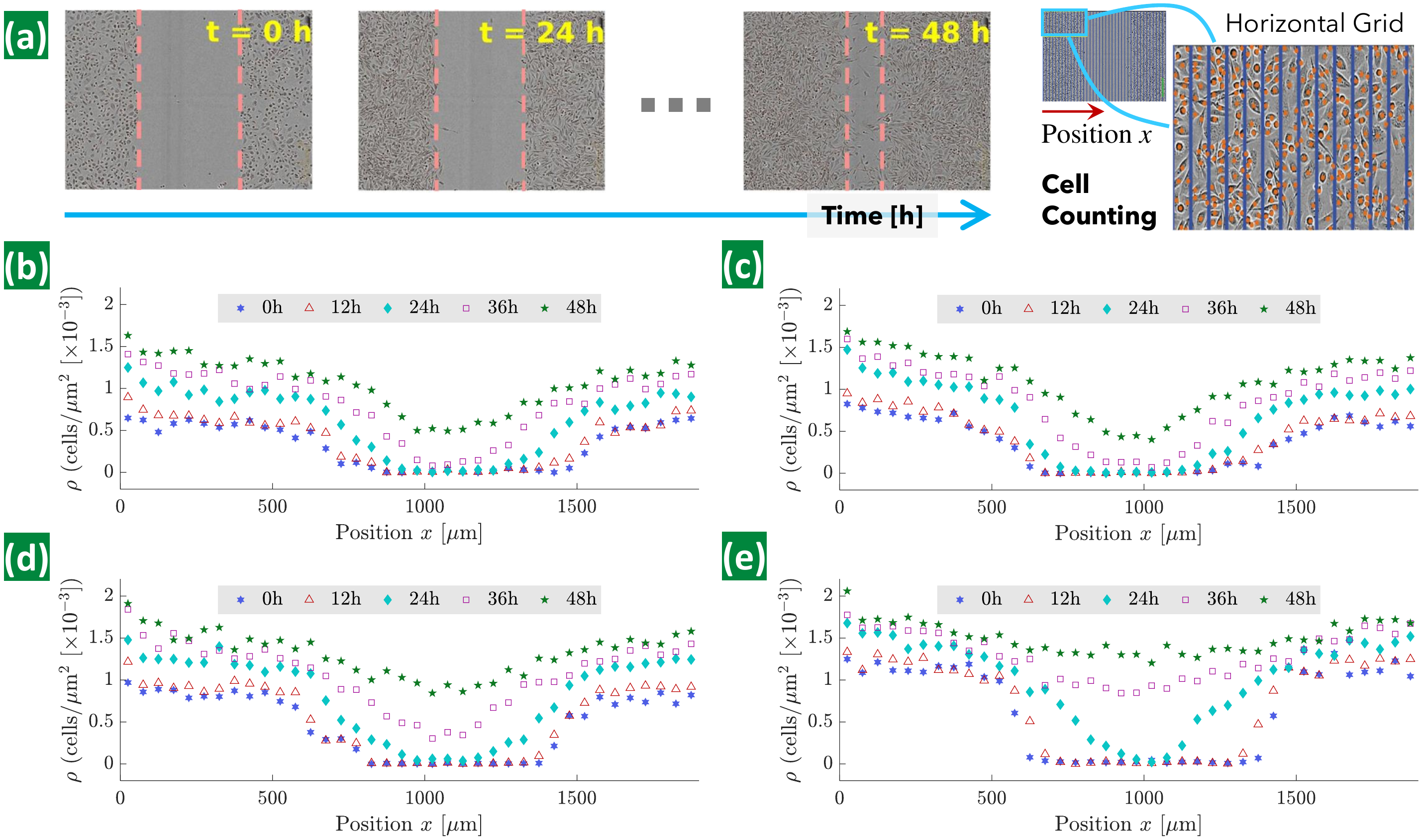}
\caption{Measurement datasets of cell densities, $\rho$, based on scratch assays \cite{jin2016reproducibility}. (a) Example scratch assay imaging of 16,000 cells in the test well with a width of 1,900 $\mu$m (the images are reproduced from Jin \textit{et al.} \cite{jin2016reproducibility}). The images are taken at different time instants (0h, 12h, 24h, 36h, 48h). The dashed lines show the approximate location of the positions of the leading edge. These images are then evenly divided into 38 segments (50 $\mu$m each) horizontally, where the cells are counted in each segment to determine the horizontal cell densities. (b)-(e) the cell densities at different time instants for 14,000, 16,000, 18,000 and 20,000 cells, respectively.}
\label{fig:Cell_meas}
\end{figure}

\end{document}


\title{\textbf{Supplementary Information} for: \\ Physics-informed learning of governing equations from scarce data}

\author[1]{Zhao Chen}
\author[2,*]{Yang Liu}
\author[1,3,$\ddag$]{Hao Sun}

\affil[1]{\small Department of Civil and Environmental Engineering, Northeastern University, Boston, MA 02115, USA}
\affil[2]{Department of Mechanical and Industrial Engineering, Northeastern University, Boston, MA 02115, USA}
\affil[3]{Department of Civil and Environmental Engineering, MIT, Cambridge, MA 02139, USA \vspace{12pt}} 
\affil[*]{Corresponding author. E-mail: yang1.liu@northeastern.edu}
\affil[$\ddag$]{Corresponding author. E-mail: h.sun@northeastern.edu}

\date{}

\maketitle

\vspace{-12pt}

{\small
\tableofcontents
}

\vspace{36pt}

\noindent This supplementary document provides a detailed description of the proposed algorithm, examples, and discussion of technical challenges for discovering closed-form partial differential equations (PDEs) from scarce and noisy data.

\section{Alternating Direction Optimization (ADO): Algorithm}
The proposed ADO algorithm for training the network of PiDL with sparse regression is outlined in Algorithm \ref{alg:PiDL}, where the STRidge sub-function (a sequential thresholding regression process that serves as a proxy for $\ell_0$ regularization \cite{Brunton3932, Rudye1602614}) is given in Algorithm \ref{alg:STR}.

\begin{algorithm}[htbp]
	\caption{\small{The proposed ADO for network training: $[\boldsymbol{\theta}_\text{best}, \boldsymbol{\Lambda}_\text{best}] = \texttt{ADO}\left( \mathcal{D}_u, \mathcal{D}_c, \Delta\delta, n_\text{max}, n_\text{str} \right)$}}
	\label{alg:PiDL}
    \footnotesize
    \begin{algorithmic}[1]
		
		\STATE {\bfseries Input:} Measurement data $\mathcal{D}_u$, collocation points $\mathcal{D}_c=\{\mathbf{x}_i, t_i\}_{i=1, 2, ..., N_c}$, relative weighting of loss functions $\alpha$ and $\beta$, threshold tolerance increment $\Delta\delta$ for STRidge, maximum number of alternating iterations $n_\text{max}$, and maximum number of STRidge iterations $n_\text{str}$. \par \noindent \textcolor{gray}{\# we take a 2D system in a 2D domain as an example: $\mathbf{u}=\{u, v\}$ and $\mathbf{x}=\{x, y\}$}  \par \noindent \textcolor{gray}{\# $\alpha$ typically takes 1 (or 10 if the magnitude of $\mathcal{L}_p$ is at least one order lower compared with the data loss)} \par \noindent \textcolor{gray}{\# $\beta$ takes a small positive value, e.g., $10^{-6}\sigma_{u}$ where $\sigma_{u}$ is the standard variation of the measurement data} \vspace{2pt}
		
		\STATE Split measurement data $\mathcal{D}_u$ into training-validation sets ($n_\text{tr}/n_\text{va}=80/20$): $\mathcal{D}_u^\text{tr}\in\mathbb{R}^{n_\text{tr}\times 2}$ and $\mathcal{D}_u^\text{va}\in\mathbb{R}^{n_\text{va}\times 2}$. ~~~\textcolor{gray}{\# $N_m=n_\text{tr}+n_\text{va}$ }\vspace{2pt}
		
		\STATE Split collocation points $\mathcal{D}_c$ into training-validation sets ($m_\text{tr}/m_\text{va}=80/20$): $\mathcal{D}_c^\text{tr}\in\mathbb{R}^{m_\text{tr}\times 3}$ and $\mathcal{D}_c^\text{va}\in\mathbb{R}^{m_\text{va}\times 3}.$ ~~~\textcolor{gray}{\# $N_c=m_\text{tr}+m_\text{va}$ }	\vspace{1pt}

		\STATE Initialize the \textit{Tensor Graph} for the entire network. \vspace{2pt}
		
		\STATE Pre-train the network via combined Adam and L-BFGS with $\{\mathcal{D}_u^\text{tr}, \mathcal{D}_c^\text{tr}\}$, and validate the trained model with $\{\mathcal{D}_u^\text{va}, \mathcal{D}_c^\text{va}\}$, namely, \par \vspace{3pt} {\begin{flushright} $\{\hat{\boldsymbol{\theta}}_0, \hat{\boldsymbol{\Lambda}}_0\} = \mathop{\arg\min}_{\{\boldsymbol{\theta}, \mathbf{\Lambda}\}} \big\{\mathcal{L}_d(\boldsymbol{\theta}; \mathcal{D}_u) + \alpha \mathcal{L}_p(\boldsymbol{\theta}, \boldsymbol{\Lambda}; \mathcal{D}_c) + \beta\|\boldsymbol{\Lambda}\|_1 \big\}.$  ~~~ \textcolor{gray}{\# pre-train the network; $\hat{\boldsymbol{\Lambda}}_0=\big\{\hat{\boldsymbol{\lambda}}_0^u, \hat{\boldsymbol{\lambda}}_0^v\big\}$} \end{flushright}} \vspace{3pt} 
		
		\FOR{$k = 1,2,...,n_\text{max}$} \vspace{1pt}
		
		\STATE Assemble the system states over the collocation points $\mathcal{D}_c^\text{tr}$ and $\mathcal{D}_c^\text{va}$:  \par \vspace{3pt} {\begin{center} \(
        \dot{\mathbf{U}}_u^\text{tr} = \bigcup_{i=1}^{N_c^\text{tr}} u_t\big(\hat{\boldsymbol{\theta}}_{k-1}; \mathbf{x}_i^\text{tr}, t_i^\text{tr}\big) ~~~\text{and}~~~ \dot{\mathbf{U}}_u^\text{va} = \bigcup_{i=1}^{N_c^\text{tr}}u_t\big(\hat{\boldsymbol{\theta}}_{k-1}; \mathbf{x}_i^\text{va}, t_i^\text{va}\big)\) \par \vspace{3pt}
        \(\dot{\mathbf{U}}_v^\text{tr} = \bigcup_{i=1}^{N_c^\text{va}}v_t\big(\hat{\boldsymbol{\theta}}_{k-1}; \mathbf{x}_i^\text{tr}, t_i^\text{tr}\big) ~~~\text{and}~~~ \dot{\mathbf{U}}_v^\text{va} = \bigcup_{i=1}^{N_c^\text{tr}} v_t\big(\hat{\boldsymbol{\theta}}_{k-1}; \mathbf{x}_i^\text{va}, t_i^\text{va}\big). \)  \end{center}} \vspace{3pt}
		
		\STATE Assemble the candidate library matrices over the collocation points $\mathcal{D}_c$, $\mathcal{D}_c^\text{tr}$ and $\mathcal{D}_c^\text{va}$:  \par \vspace{3pt} {\begin{center}\(\tilde{\boldsymbol{\Phi}} = \bigcup_{i=1}^{N_c} \boldsymbol{\phi}\big(\hat{\boldsymbol{\theta}}_{k-1}; \mathbf{x}_i, t_i\big)\),~~ \(\tilde{\boldsymbol{\Phi}}^\text{tr} = \bigcup_{i=1}^{N_c^\text{tr}} \boldsymbol{\phi}\big(\hat{\boldsymbol{\theta}}_{k-1}; \mathbf{x}_i^\text{tr}, t_i^\text{tr}\big)\) ~~and~~ \(\tilde{\boldsymbol{\Phi}}^\text{va} = \bigcup_{i=1}^{N_c^\text{va}} \boldsymbol{\phi}\big(\hat{\boldsymbol{\theta}}_{k-1}; \mathbf{x}_i^\text{va}, t_i^\text{va}\big).\)  \end{center}} \vspace{3pt} 
		
		\STATE Normalize candidate library matrices $\tilde{\boldsymbol{\Phi}}, \tilde{\boldsymbol{\Phi}}^\text{tr}$ and $\tilde{\boldsymbol{\Phi}}^\text{va}$ column-wisely ($j = 1, ..., s$) to improve matrix condition: \par \vspace{3pt} 
		{\begin{center}\(\boldsymbol{\Phi}_{:, j} = \tilde{\boldsymbol{\Phi}}_{:, j}\big/\norm{\tilde{\boldsymbol{\Phi}}_{:, j}}_2,~~
		\boldsymbol{\Phi}^\text{tr}_{:, j} = \tilde{\boldsymbol{\Phi}}^\text{tr}_{:, j}\big/\norm{\tilde{\boldsymbol{\Phi}}^\text{tr}_{:, j}}_2 ~~\text{and}~~
		\boldsymbol{\Phi}^\text{va}_{:, j} = \tilde{\boldsymbol{\Phi}}^\text{tr}_{:, j}\big/\norm{\tilde{\boldsymbol{\Phi}}^\text{tr}_{:, j}}_2. \) \end{center}} \vspace{3pt} 
		
		\STATE Determine $\ell_0$ regularization parameter $\gamma = 0.001\kappa(\boldsymbol{\Phi})$. ~~~~ \textcolor{gray}{\# $\kappa(\cdot)$ denotes the matrix condition number;} \vspace{3pt} 
		
		\STATE Initialize the error indices: {$\hat{\epsilon}^u= \big\|\boldsymbol{\Phi}^\text{va}\hat{\boldsymbol{\lambda}}_{k-1}^u - \dot{\mathbf{U}}_u^\text{va} \big\|_2^2 + \gamma\big\|\hat{\boldsymbol{\lambda}}_{k-1}^u\big\|_0 $ ~~and~~ $\hat{\epsilon}^v= \big\|\boldsymbol{\Phi}^\text{va}\hat{\boldsymbol{\lambda}}_{k-1}^v - \dot{\mathbf{U}}_v^\text{va} \big\|_2^2 + \gamma\big\|\hat{\boldsymbol{\lambda}}_{k-1}^v\big\|_0 $.} \vspace{3pt}
		
		\STATE Set the initial threshold tolerance $\delta_1 = \Delta\delta$. \vspace{1pt}
		
		\FOR{$iter = 1,2,...,n_\text{str}$} \vspace{1pt}
		
		    \STATE Run STRidge as shown in Algorithm \ref{alg:STR} to determine: \par \vspace{3pt} {\begin{center} $ \tilde{\boldsymbol{\lambda}}^u = \texttt{STRidge}\big(\dot{\mathbf{U}}_u^\text{tr}, \boldsymbol{\Phi}^\text{tr},  \delta_{iter}\big)$ ~~ and ~~ $ \tilde{\boldsymbol{\lambda}}^v = \texttt{STRidge}\big(\dot{\mathbf{U}}_v^\text{tr}, \boldsymbol{\Phi}^\text{tr}, \delta_{iter}\big)$. \end{center}}  \vspace{3pt}

		    \STATE Update the error indices: $\epsilon^u= \big\|\boldsymbol{\Phi}^\text{va}\tilde{\boldsymbol{\lambda}}^u - \dot{\mathbf{U}}_u^\text{va} \big\|_2^2 + \gamma\big\|\tilde{\boldsymbol{\lambda}}^u\big\|_0 $ ~~ and ~~ $\epsilon^v= \big\|\boldsymbol{\Phi}^\text{va}\tilde{\boldsymbol{\lambda}}^v - \dot{\mathbf{U}}_v^\text{va} \big\|_2^2 + \gamma\big\|\tilde{\boldsymbol{\lambda}}^v\big\|_0 $.  \vspace{4pt}
		
		    \IF{$\epsilon^u \leq \hat{\epsilon}^u \mathrm{~or~} \epsilon^v \leq \hat{\epsilon}^v $ (run in parallel)} \vspace{1pt}
		        \STATE Increase threshold tolerance with increment: $\delta_{iter + 1} = \delta_{iter}+\Delta\delta$. \vspace{1pt}
		    \ELSE \vspace{1pt}
		        \STATE Decrease threshold tolerance increment $\Delta\delta=\Delta\delta/1.618$. \vspace{1pt}
		        \STATE Update threshold tolerance with the new increment $\delta_{iter + 1} = \max \{ \delta_{iter}-2\Delta\delta, 0\} + \Delta\delta $. \vspace{1pt}
		    \ENDIF \vspace{1pt}

		\ENDFOR \vspace{1pt}
		
		\STATE Return and re-scale the current best solution from STRidge iterations: $\hat{\boldsymbol{\Lambda}}_{k} = \big\{\tilde{\boldsymbol{\lambda}}^u, \tilde{\boldsymbol{\lambda}}^v\big\}$. ~~~ \textcolor{gray}{\# re-scaling due to normalization of $\boldsymbol{\Phi}$} \vspace{3pt} 
		
		\STATE Train the DNN via combined Adam and L-BFGS with $\{\mathcal{D}_u^\text{tr}, \mathcal{D}_c^\text{tr}\}$, and validate the trained model with $\{\mathcal{D}_u^\text{va}, \mathcal{D}_c^\text{va}\}$, namely, \par \vspace{3pt} {\begin{center}$\hat{\boldsymbol{\theta}}_k = \mathop{\arg\min}_{\boldsymbol{\theta}} \big\{\mathcal{L}_d(\boldsymbol{\theta}; \mathcal{D}_u) + \alpha\mathcal{L}_p(\boldsymbol{\theta}, \hat{\boldsymbol{\Lambda}}_{k}; \mathcal{D}_c) \big\}$.  ~~~ \textcolor{gray}{\# train DNN given $\hat{\boldsymbol{\Lambda}}_{k}$ as known} \end{center}} \vspace{3pt} 
		
		\ENDFOR

		\STATE {\bfseries Output:} the best solution $\boldsymbol{\theta}_\text{best} = \hat{\boldsymbol{\theta}}_{n_\text{max}}$ and $\boldsymbol{\Lambda}_\text{best} = \hat{\boldsymbol{\Lambda}}_{n_\text{max}}$
	
	\end{algorithmic} 
\end{algorithm}

\begin{algorithm}[t!]
	\caption{\small{Sequential threshold ridge regression (STRidge): $\hat{\boldsymbol{\lambda}} = \texttt{STRidge}\big(\dot{\mathbf{U}}, \boldsymbol{\Phi}, \delta\big)$}}
	\label{alg:STR}
    \footnotesize
    \begin{algorithmic}[1]
		
		\STATE {\bfseries Input:} Time derivative vector $\dot{\mathbf{U}}$, candidate function library matrix $\boldsymbol{\Phi}$, and threshold tolerance $\delta$. \vspace{1pt} 
		
		\STATE Inherit coefficients $\hat{\boldsymbol{\lambda}}$ from the DNN pre-training or the previous update. \vspace{1pt}      
		
		\REPEAT \vspace{1pt} 
		
		\STATE Determine indices of coefficients in $\hat{\boldsymbol{\lambda}}$ falling below or above the sparsity threshold $\delta$: \par \vspace{1pt} {\begin{center} $\mathcal{I} = \{i\in\mathcal{I}: |\hat{\lambda}_i| < \delta\}$ and $\mathcal{J} = \{j\in\mathcal{J}: |\hat{\lambda}_j| \geq \delta\}$. \end{center}} \vspace{1pt} 
		
		\STATE Enforce sparsity to small values by setting them to zero: $\hat{\boldsymbol{\lambda}}_\mathcal{I} = \mathbf{0}$. \vspace{1pt} 
		
		\STATE Update remaining non-zero values with ridge regression:
		\par \vspace{1pt} {\begin{flushright} $\hat{\boldsymbol{\lambda}}_\mathcal{J} = \operatorname*{arg\,min}_{\boldsymbol{\lambda}_\mathcal{J}} \big\{ \big\|\boldsymbol{\Phi}_ \mathcal{J}\boldsymbol{\lambda}_\mathcal{J} - \dot{\mathbf{U}} \big\|_2^2 + 1\times 10^{-5}\big\|\boldsymbol{\lambda}_\mathcal{J}\big\|^2_2\big\}$. ~~~ \textcolor{gray}{\# the parameter $1\times10^{-5}$ is tunable} \end{flushright}}
		
		\UNTIL{maximum number of iterations reached.} \vspace{2pt} 
		
		\STATE {\bfseries Output:} The best solution $\hat{\boldsymbol{\lambda}} = \hat{\boldsymbol{\lambda}}_\mathcal{I} \cup \hat{\boldsymbol{\lambda}}_\mathcal{J}$
	
	\end{algorithmic} 
\end{algorithm}

\section{Examples}
We observe the efficacy and robustness of our methodology on a group of canonical PDEs used to represent a wide range of physical systems with nonlinear, periodic and/or chaotic behaviors. In particular, we discover the closed forms of Burgers', Kuramoto-Sivashinsky (KS), nonlinear Schr\"{o}dinger, Navier-Stokes (NS), and $\lambda$-$\omega$ Reaction-Diffusion (RD) equations from scarce and noisy time-series measurements recorded by a number of sensors at fixed locations from a single IBC. Gaussian white noise is added to the synthetic response with the noise level defined as the root-mean-square ratio between the noise and the exact solution. To demonstrate the ``root-branch'' network for discovery of PDE(s) based on multiple independent datasets sampled under different IBCs, we consider (1) the 1D Burgers' equation with light viscosity that exhibits a shock behavior, and (2) a 2D Fitzhugh-Nagumo (FN) type reaction-diffusion system that describes activator-inhibitor neuron activities excited by external stimulus. At last, we test our framework on the experimental data of cell migration and proliferation. Our method is also compared with SINDy (the PDE-FIND approach presented in \cite{Rudye1602614}) which is also presented herein. The identification error is defined as the average relative error of the identified non-zero PDE coefficients with respect to the ground truth, which is used to evaluate the accuracy of the discovered PDEs for the following examples. Simulations in this paper are performed on a workstation with 28 Intel Core i9-7940X CPUs and 2 NVIDIA GTX 1080Ti GPU cards.

\subsection{Discovery of Benchmark PDEs with Single Dataset}
\subsubsection{Burgers' equation}\label{Burgers}
\begin{figure*}[t!]
\centering
\includegraphics[width=0.78\linewidth]{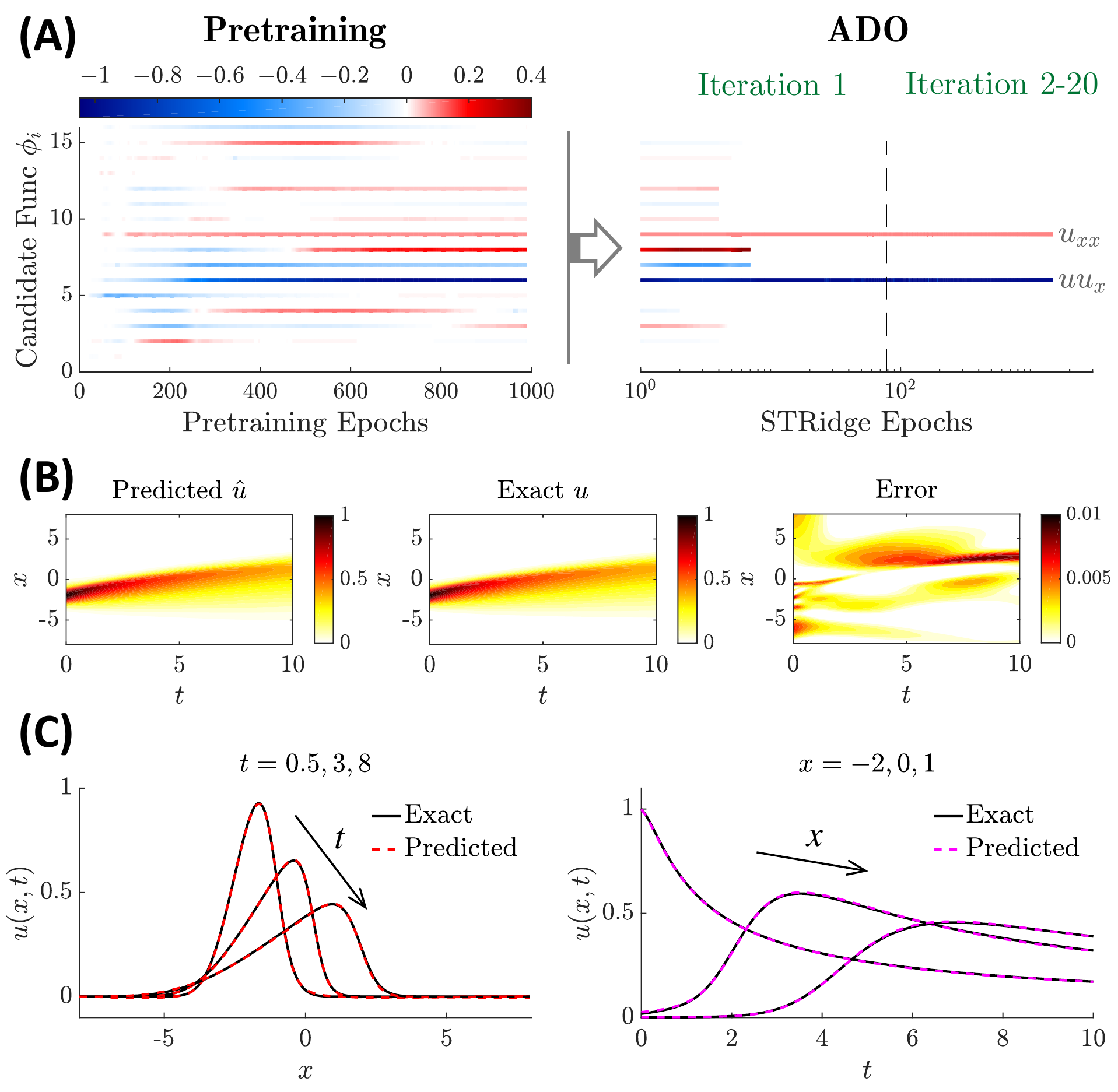}
\caption{Discovered Burgers' equation for data with 10\% noise. (\textit{A}) Evolution of the sparse coefficients $\boldsymbol{\Lambda}\in\mathbb{R}^{16\times1}$ for 16 candidate functions $\boldsymbol{\phi}\in\mathbb{R}^{1\times16}$ used to form the PDE, where the color represents the coefficient value. (\textit{B}) The predicted response in comparison with the exact solution with the prediction error. (\textit{C}) Comparison of spatial and temporal snapshots between the predicted and the exact solutions. The relative full-field $\ell_2$ error of the prediction is 2.02\%.}
\label{Sup_fig:Burgers}
\end{figure*}

We first consider a dissipative system with the dynamics governed by a 1D viscous Burgers' equation expressed as 
\begin{equation*} \label{eq:Burgers}
u_t = - uu_x + \nu u_{xx}
\end{equation*}
where $u$ is a field variable, $x$ and $t$ are the spatial and temporal coordinates, and $\nu$ denotes the diffusion coefficient. The equation describes the decaying stationary viscous shock of a system after a finite period of time, commonly found in simplified fluid mechanics, nonlinear acoustics, gas dynamics and traffic flow. In this work, solution for the Burgers' Equation is from an open dataset \cite{Rudye1602614}, in which the diffusion coefficient $\nu$ is assumed to be 0.1 and $u$ is discretized into 256 spatial grid points for 101 time steps with a Gaussian initial condition. In particular, 5 sensors are randomly placed at fixed locations among the 256 spatial grid points to record the wave response for 101 time steps, leading to 1.95\% of the dataset used in \cite{Rudye1602614}. A total number of $1.6\times10^5$ collocation points, e.g., in the pair of $\{x, t\}$, are sampled by the Sobol sequence \cite{Sobol}. A group of 16 candidate functions ($\boldsymbol{\phi}\in\mathbb{R}^{1\times 16}$) are used to reconstruct the PDE, consisting of polynomial terms ($u, u^2, u^3$), derivatives ($u_x, u_{xx}, u_{xxx}$) and their multiplications. The fully connected DNN has 8 hidden layers and a width of 20 neuron nodes in each layer. The training efforts are performed via $1\times10^3$ epochs of L-BFGS for pre-training and 20 ADO iterations. In each ADO iteration, we use the combination of 100 epochs of Adam and $1\times10^3$ (or less, depending on the relative loss decay) epochs of L-BFGS to train the DNN for alternation with STRidge. The discovered equation for the dataset with 10 \% noise reads:
\begin{equation*} \label{eq:Burgers_discovered}
u_t = -0.997uu_x + 0.098 u_{xx}
\end{equation*}
where the aggregated relative identification error for all non-zero elements in $\boldsymbol{\Lambda}$ is $1.15 \pm 1.20 \%$. The discovery result is summarized in Fig. \ref{Sup_fig:Burgers}. The evolution of the PDE coefficients for all candidate functions is illustrated in Fig. \ref{Sup_fig:Burgers}\textit{A}. While all coefficients are initialized as zeros, they vary evidently in pre-training process. In the ADO stage, the adaptive sparsity threshold in STRidge gradually prunes redundant components right after the first alternating iteration. Despite that only 1.95\% subsampled responses are measured, the PiDL approach can accurately extrapolate the full-field solution with a $\ell_2$ error of 2.02\% (see Fig. \ref{Sup_fig:Burgers}\textit{B}). Fig. \ref{Sup_fig:Burgers}\textit{C} shows the comparison of spatial and temporal snapshots between the predicted and the exact solutions which agree extremely well.

\begin{figure*}[t!]
\centering
\includegraphics[width=0.78\linewidth]{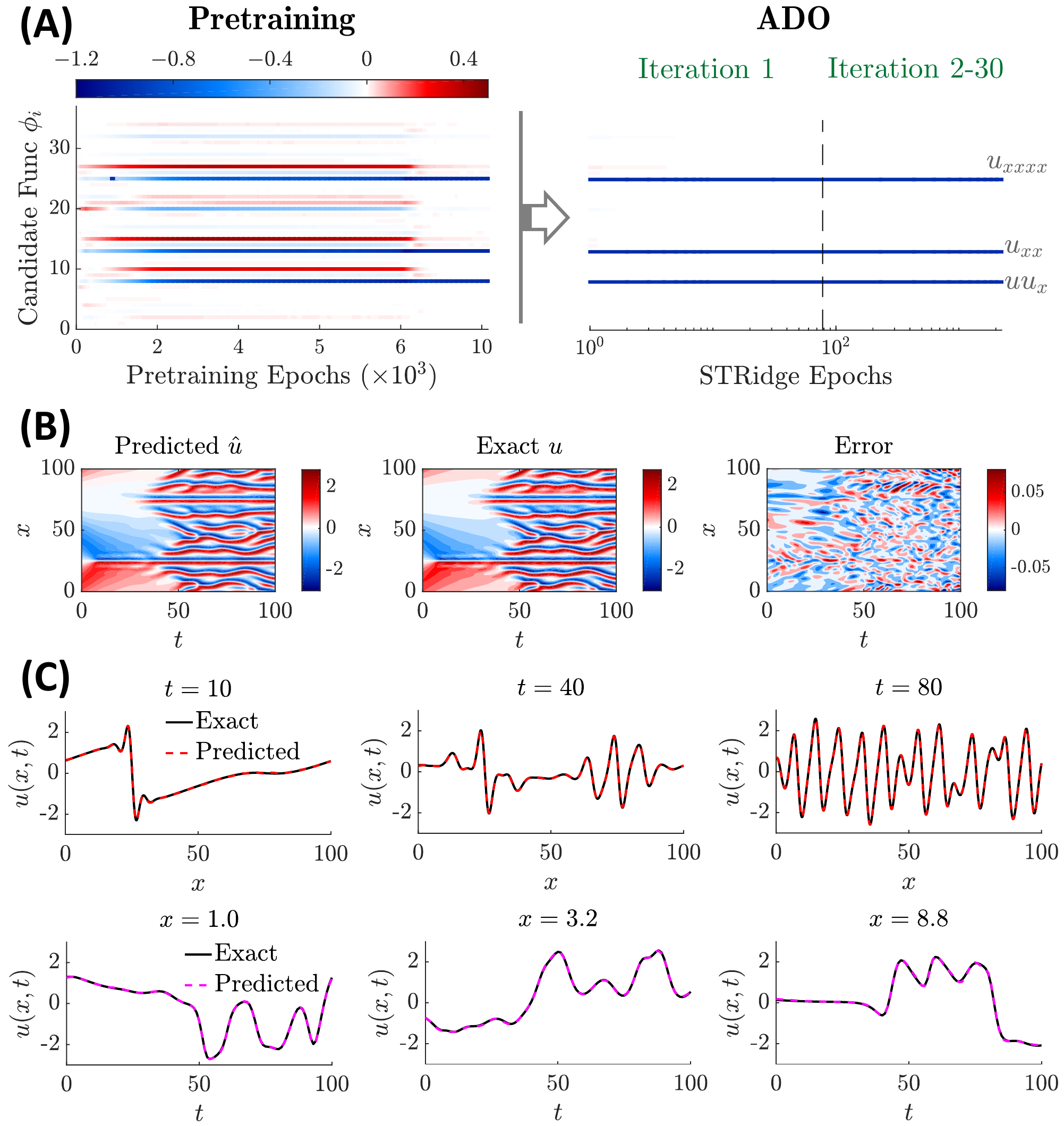}
\caption{Discovered the KS equation for data with 10\% noise. (\textit{A}) Evolution of the sparse coefficients $\boldsymbol{\Lambda}\in\mathbb{R}^{36\times1}$ for 36 candidate functions $\boldsymbol{\phi}\in\mathbb{R}^{1\times36}$ used to reconstruct the PDE. (\textit{B}) The predicted response compared with the exact solution. (\textit{C}) Comparison of spatial and temporal snapshots between the predicted and the exact solutions. The relative full-field $\ell_2$ error of the prediction is 1.87\%.}
\label{Sup_fig:Sup_KS}
\end{figure*}

\subsubsection{Kuramoto-Sivashinsky equation}
Another dissipative system with intrinsic instabilities is considered, governed by the 1D Kuramoto-Sivashinsky (KS) equation:
\begin{equation*} 
\label{eq:KS}
u_t = - uu_x - u_{xx} - u_{xxxx}
\end{equation*}
where the reverse diffusion term $-u_{xx}$ leads to the blowup behavior while the fourth-order derivative $u_{xxxx}$ introduces chaotic patterns as shown in Fig. \ref{Sup_fig:Sup_KS}\textit{B}, making an ideal test problem for equation discovery. Starting with a smooth initial condition, the KS system evolves to an unstable laminar status due to the highly nonlinear terms including the high-order derivative. The KS equation is widely used to model the instabilities in laminar flame fronts and dissipative trapped-ion modes among others. We subsample the open dataset \cite{Rudye1602614} by randomly choosing 320 points from the 1024 spatial grid nodes as fixed sensors and record the wave response for 101 time steps, occupying about 12.3\% of the original dataset. A set of $2\times10^4$ collocation points, sampled using the Sobol sequence in the spatiotemporal domain, are employed to evaluate the residual physics loss. A library of 36 candidate functions are used to construct the PDE, consisting of polynomials ($u, u^2, u^3, u^4, u^5$), derivatives ($u_x, u_{xx}, u_{xxx}, u_{xxxx}, u_{xxxxx}$) and their multiplications. The DNN architecture and hyperparameters are same as those in the Burgers’ equation example. 

It is notable that the chaotic behavior poses significant challenges in approximating the full-field spatiotemporal derivatives, especially the high-order $u_{xxxx}$, from poorly measured data for discovery of such a PDE. Existing methods (for example the family of SINDy methods \cite{Rudye1602614, Schaeffer2017}) eventually fail in this case given very coarse and noisy measurements. Nevertheless, the PiDL approach successfully distils the closed form of the KS equation from subsampled sparse data even with 10\% noise:
\begin{equation*}\label{eq:KS_discovered}
u_t = - 0.992uu_x - 0.990u_{xx} - 0.991u_{xxxx}
\end{equation*}
where the coefficients have an average relative error, for all non-zero elements in $\boldsymbol{\Lambda}$, of $0.71\pm0.06\%$. The evolution of the coefficients $\boldsymbol{\Lambda}\in\mathbb{R}^{36\times1}$ in Fig. \ref{Sup_fig:Sup_KS}\textit{A} illustrates that both the candidate terms and the corresponding coefficients are correctly identified (close to the original parameters) within a small number of ADO iterations. Although the available measurement data are sparsely sampled in the spatiotemporal domain under a high-level noise corruption, the predicted full-field wave by the trained PiDL also agrees well with the exact solution with a relative $\ell_2$ error of 1.87\% (Fig. \ref{Sup_fig:Sup_KS}\textit{B}). The spatial and temporal snapshots of the predicted response match seamlessly the ground truth as shown in  Fig. \ref{Sup_fig:Sup_KS}\textit{C}.

\subsubsection{Nonlinear Schr\"{o}dinger equation}
In the third example, we discover the nonlinear Schr\"{o}dinger equation, originated as a classical wave equation, given by 
\begin{equation*} 
\label{eq:NSE}
i u_t = -0.5u_{xx}-|u|^2u
\end{equation*}
where $u$ is a complex field variable. This well-known equation is widely used in modeling the propagation of light in nonlinear optical fibers, Bose-Einstein condensates, Langmuir waves in hot plasmas, and so on. The solution to this Schr\"{o}dinger equation is simulated based on a Gaussian initial condition with the problem domain meshed into 512 spatial points and 501 temporal steps, while the measurements are taken from 256 randomly chosen spatial ``sensors'' for 375 time instants, resulting in 37.5\% data used in \cite{Rudye1602614} for uncovering the closed form of the equation. A library of 40 candidate functions are used for constructing the PDE, varying among polynomial functions ($u, u^2, u^3$), absolute values ($|u|, |u|^2, |u|^3$), derivatives ($u_x, u_{xx}, u_{xxx}$) and their combination. Since the function is complex valued, we model separately the real part ($u_R$) and the imaginary part ($u_I$) of the solution in the output of the DNN, assemble them to obtain the complex solution $u = u_R+iu_I$, and construct the complex valued candidate functions for discovery. To avoid complex gradients in optimization, we use the modulus (magnitude, $|u|$), instead of the $\ell_2$ norm, for the residual physics loss $\mathcal{L}_p$. The fully connected DNN has 8 hidden layers and a width of 40 neuron nodes in each layer. The pre-training takes $1.6\times10^5$ epochs of Adam (with additional L-BFGS tuning) followed by 30 ADO iterations. In each ADO iteration, we use $1\times10^3$ Adam epochs and up to $4\times10^3$ (depending on the relative loss decay) L-BFGS epochs to train the DNN for alternation with STRidge.

\begin{figure*}[t!]
\centering
\includegraphics[width=0.78\linewidth]{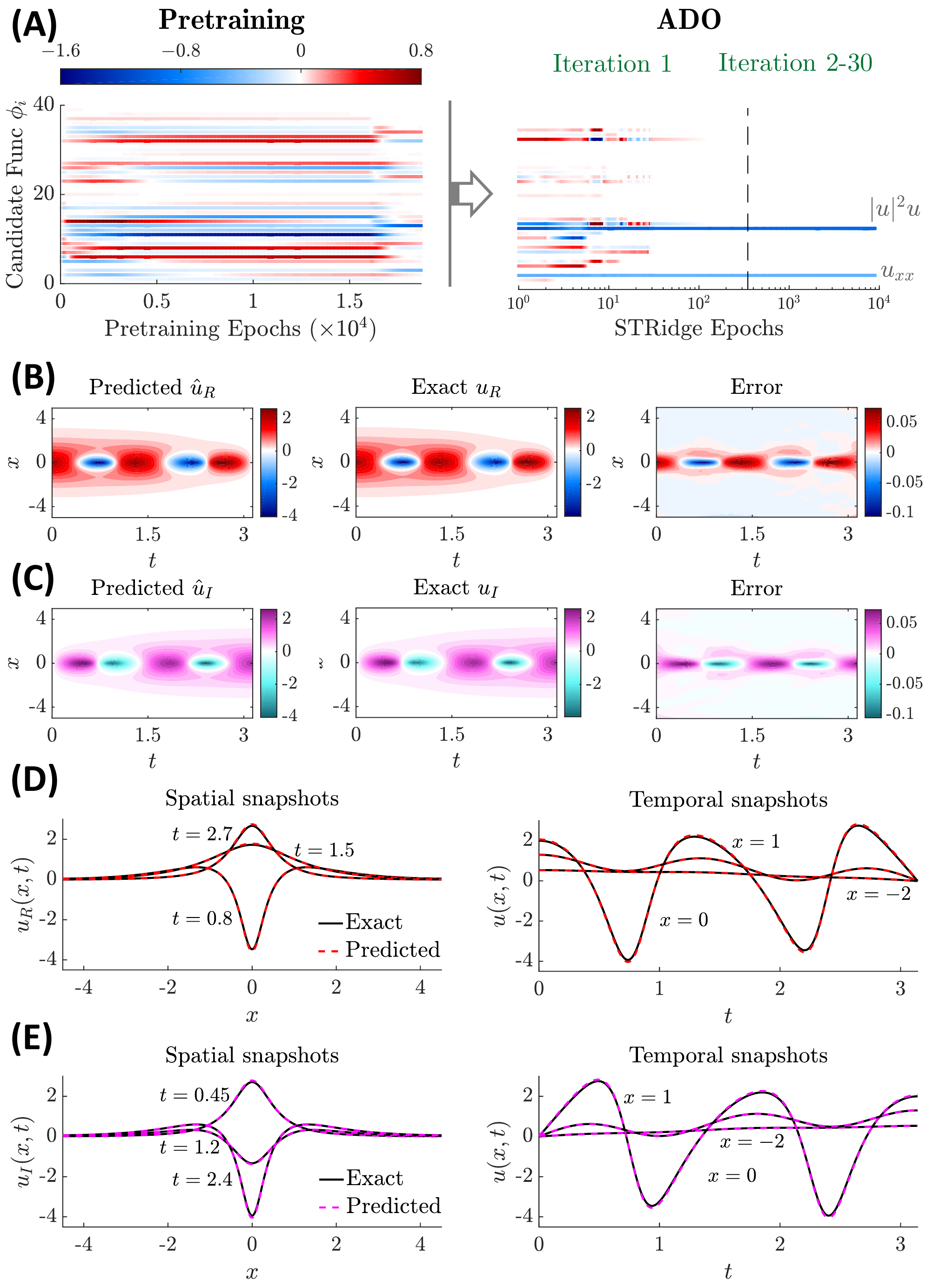}
\caption{Discovered nonlinear Schr\"{o}dinger equation for a dataset with 10\% noise. (\textit{A}) Evolution of the sparse coefficients $\boldsymbol{\Lambda}\in\mathbb{R}^{40\times1}$ for the candidate functions $\boldsymbol{\phi}\in\mathbb{R}^{1\times40}$ used to reconstruct the PDE. (\textit{B} and \textit{C}) The predicted real-part (\textit{B}) and imaginary-part (\textit{C}) responses compared with the exact solution. (\textit{D} and \textit{E}) Comparison of spatial and temporal snapshots between the predicted and the exact solutions for the real part (\textit{D}) and imaginary part (\textit{E}). The relative full-field $\ell_2$ error of the prediction is about 1\%.}
\label{Sup_fig:Sup_NSE}
\end{figure*}

The discovered equation under 10 \% noise is written as
\begin{equation*} 
\label{eq:NSE_discovered}
iu_t = -0.490u_{xx} - 0.974|u|^2u
\end{equation*}
where the average relative error for non-zero coefficients is $2.31\pm0.28\%$. The evolution history of the sparse coefficients $\boldsymbol{\Lambda}\in\mathbb{R}^{40\times1}$ clearly shows the convergence to the actual values (Fig. \ref{Sup_fig:Sup_NSE}\textit{A}) resulting in accurate closed-form identification of the PDE. Even though the evolution of $\boldsymbol{\Lambda}$ is quite intense in the  pre-training stage, the most dominant components remain after the first few ADO iterations. The predicted full-field response, for both real and imaginary parts, matches well the exact solution with a slight relative $\ell_2$ error of about 1\% (Fig. \ref{Sup_fig:Sup_NSE}\textit{B} and \textit{C}). The comparison of spatiotemporal snapshots between the predicted and the exact solutions for the real part (Fig. \ref{Sup_fig:Sup_NSE}\textit{D}) and imaginary part (Fig. \ref{Sup_fig:Sup_NSE}\textit{E}) also shows almost perfect agreement.

\begin{figure}[t!]
\centering
\includegraphics[width = 0.5\textwidth]{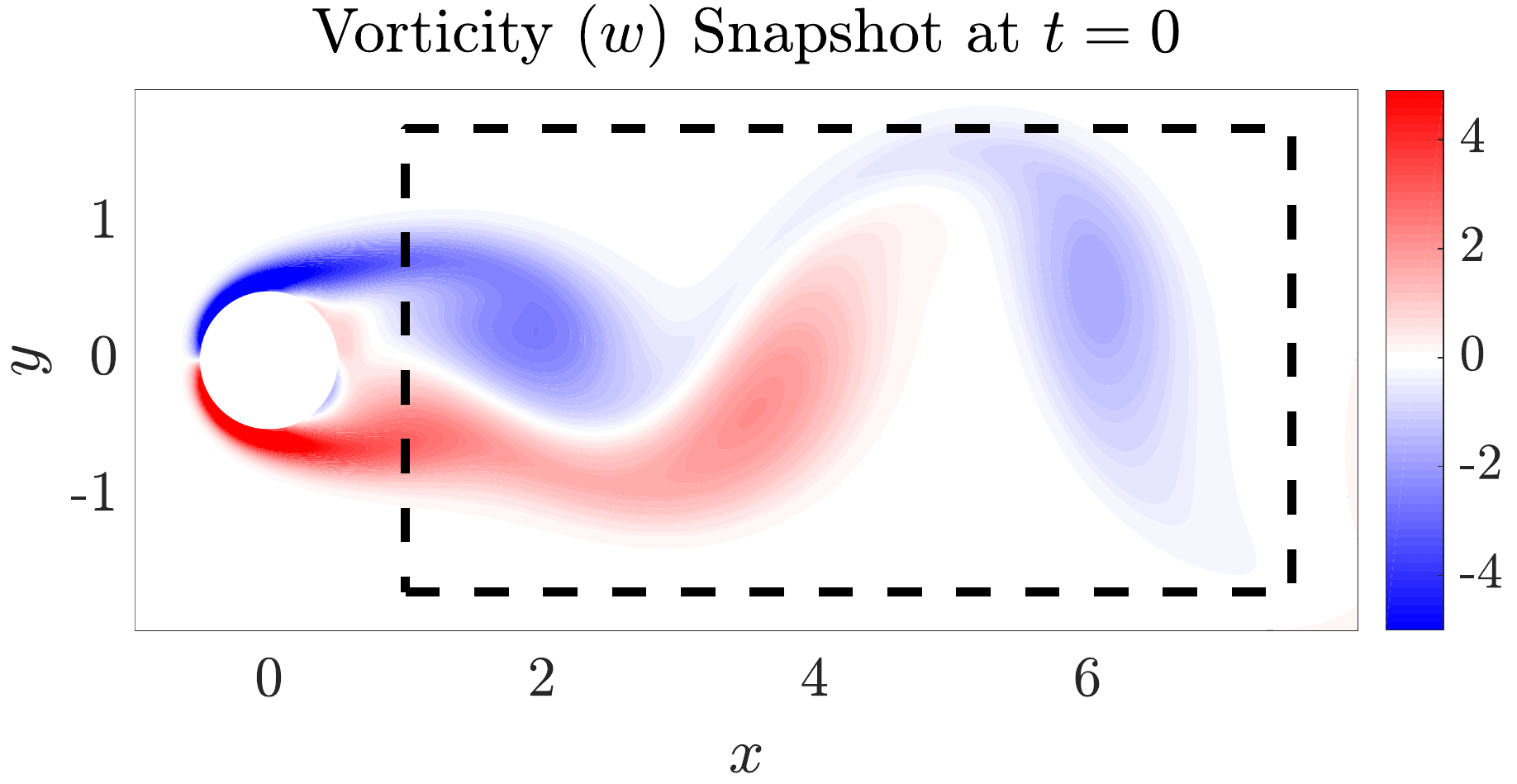}
\caption{Vorticity field $w(\mathbf{x}, t)$ at $t = 0$ for a steady flow passing a cylinder. Measurements are sampled from the the boxed area surrounded by the dashed line.}
\label{Sup_Fig:Vorticity}
\end{figure}

\subsubsection{Navier-Stokes equation}
We consider a 2D fluid flow passing a circular cylinder with the local rotation dynamics (see Fig. \ref{Sup_Fig:Vorticity}). For incompressible and isotropic fluids which also have conservative body forces, the well-known Navier-Stokes vorticity equation reads
\begin{equation*} 
\label{eq:Vorticity}
w_t = -uw_x - vw_y + 0.01w_{xx} + 0.01w_{yy}
\end{equation*}
where $w$ is the spatiotemporally variant vorticity, $\mathbf{u}=\{u, v\}$ denotes the fluid velocities at Reynolds number 100, $\nabla$ is the gradient, and $\nabla^2$ is the Laplace operator. The full-field solution to the NS vorticity equation is obtained using the immersed boundary projection method \cite{TAIRA20072118}. The dimensionless domain is discretized into a 499$\times$199 spatial grid and 151 time steps. The cylinder has a unit diameter and the input flow from the left side has a unit velocity. Measurements of velocities $\{u,v\}$ and vorticity $w$ are taken at 500 random spatial locations lasting 60 time steps in the boxed area behind the cylinder as shown in Fig. \ref{Sup_Fig:Vorticity}, namely 0.22\% subsamples from the total dataset and 1/10 of the data used in \cite{Rudye1602614}. The residual physics loss is evaluated on $6\times10^4$ collocation points randomly sampled in the spatiotemporal doamin using the Sobol sequence \cite{Sobol}. The library of candidate functions consists of 60 components including polynomial terms ($u, v, w, uv, uw, vw, u^2, v^2, w^2$), derivatives ($w_x, w_y, w_{xx}, w_{xy}, w_{yy}$) and their combination. The latent output in the DNN contains $u$, $v$ and $w$. The DNN has 8 fully connected hidden layers and a width of 60 nodes in each layer. The pre-training takes $5\times10^3$ epochs of Adam (with additional L-BFGS tuning up to $1\times10^4$ epochs) followed by 6 ADO iterations. In each ADO iteration, we use the Adam optimizer with 500 epochs and the L-BFGS with up to $1\times10^3$ epochs to train the DNN for each alternation within STRidge.

\begin{figure*}[t!]
\centering
\includegraphics[width=1\linewidth]{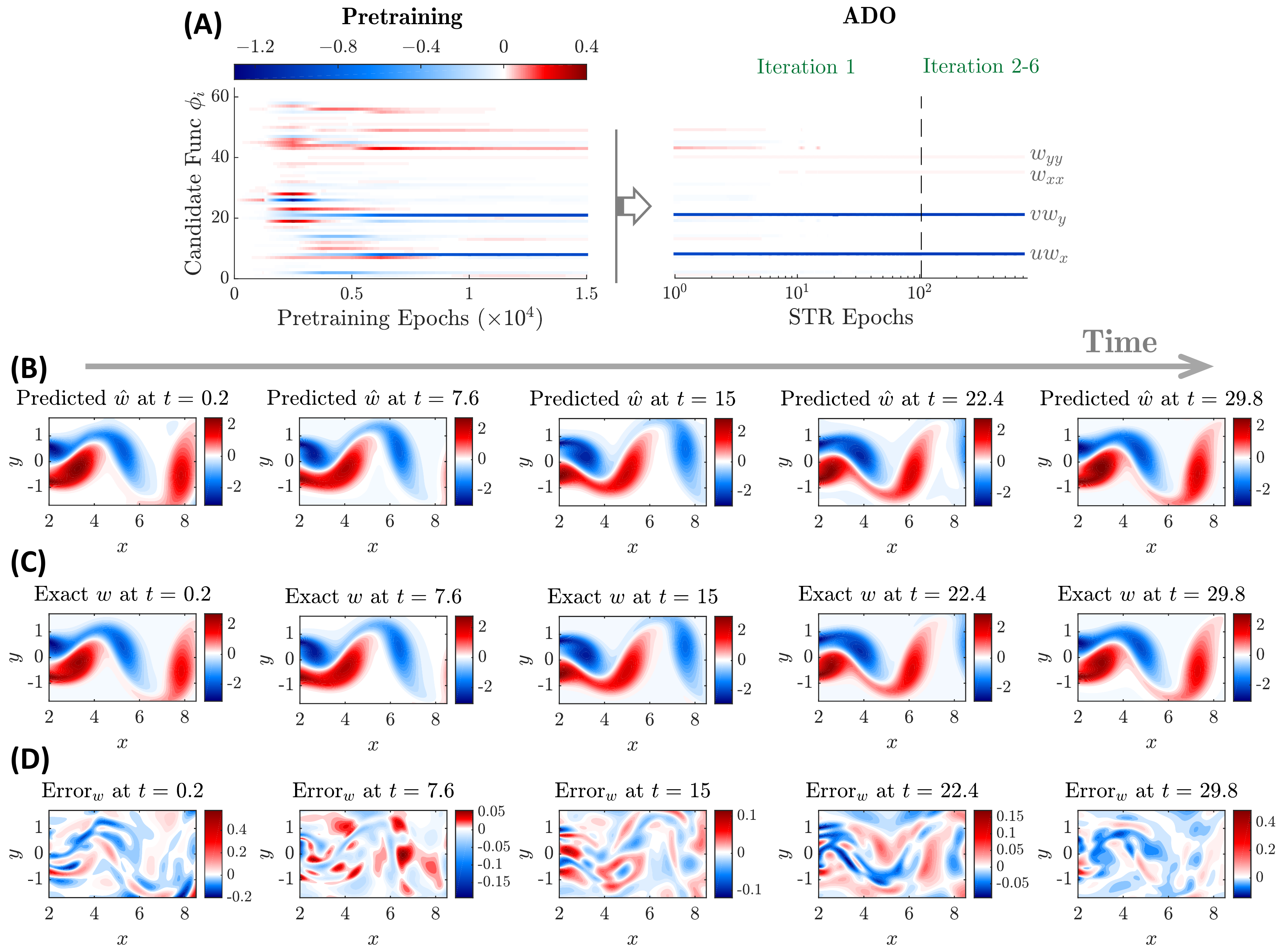}
\caption{Discovered NS equation for data with 10\% noise. (\textit{A}) Evolution of the sparse coefficients $\boldsymbol{\Lambda}\in\mathbb{R}^{60\times1}$ for 60 candidate functions $\boldsymbol{\phi}\in\mathbb{R}^{1\times60}$ used to form the vorticity equation. (\textit{B}-\textit{D}) Vorticity snapshots at different time instants ($t=0.2, 7.6, 15, 22.4, 29.8$) for the prediction (\textit{B}), the exact solution (\textit{C}) and the prediction error (\textit{D}). Note that response at these time instants are not included in dataset for training the PiDL model and equation discovery. The relative full-field $\ell_2$ error of the prediction is about 2.57\%.}
\label{Sup_fig:Sup_NS}
\end{figure*}

The discovered NS vorticity equation for the case of 10\% noise is given as follows
\begin{equation*} 
\label{eq:Vorticity_discovered}
w_t = -0.999uw_x - 0.994vw_y + 0.010w_{xx} + 0.010w_{yy}
\end{equation*}
where the aggregated relative identification error for all non-zero elements in $\boldsymbol{\Lambda}$ is $1.40\pm1.83\%$. It is encouraging that the uncovered vorticity equation is almost identical to the ground truth, for both the derivative terms and their coefficients, even under 10\% noise corruption. The coefficients $\boldsymbol{\Lambda}\in\mathbb{R}^{60\times1}$, corresponding to 60 candidate functions $\boldsymbol{\phi}\in\mathbb{R}^{1\times60}$, converge very quickly to the correct values with precise sparsity right after the first ADO iteration (Fig. \ref{Sup_fig:Sup_NS}\textit{A}). The vorticity patterns and magnitudes are also well predicted as indicated by multiple spatial snapshots at different time instants ($t=0.2, 7.6, 15, 22.4, 29.8$) shown in Fig. \ref{Sup_fig:Sup_NS}\textit{B} in comparison with the exact solution (Fig. \ref{Sup_fig:Sup_NS}\textit{C}, with small errors as depicted in Fig. \ref{Sup_fig:Sup_NS}\textit{D}). Note that the response in these snapshots is not used in training the net work. The $\ell_2$ error of the predicted full-field vorticity response is about 2.57\%. This example provides a compelling test case for the proposed PiDL approach which is capable of discovering the closed-form NS equation with scarce and noisy data.

\begin{figure}[htbp]
\centering
\includegraphics[width=0.95\linewidth]{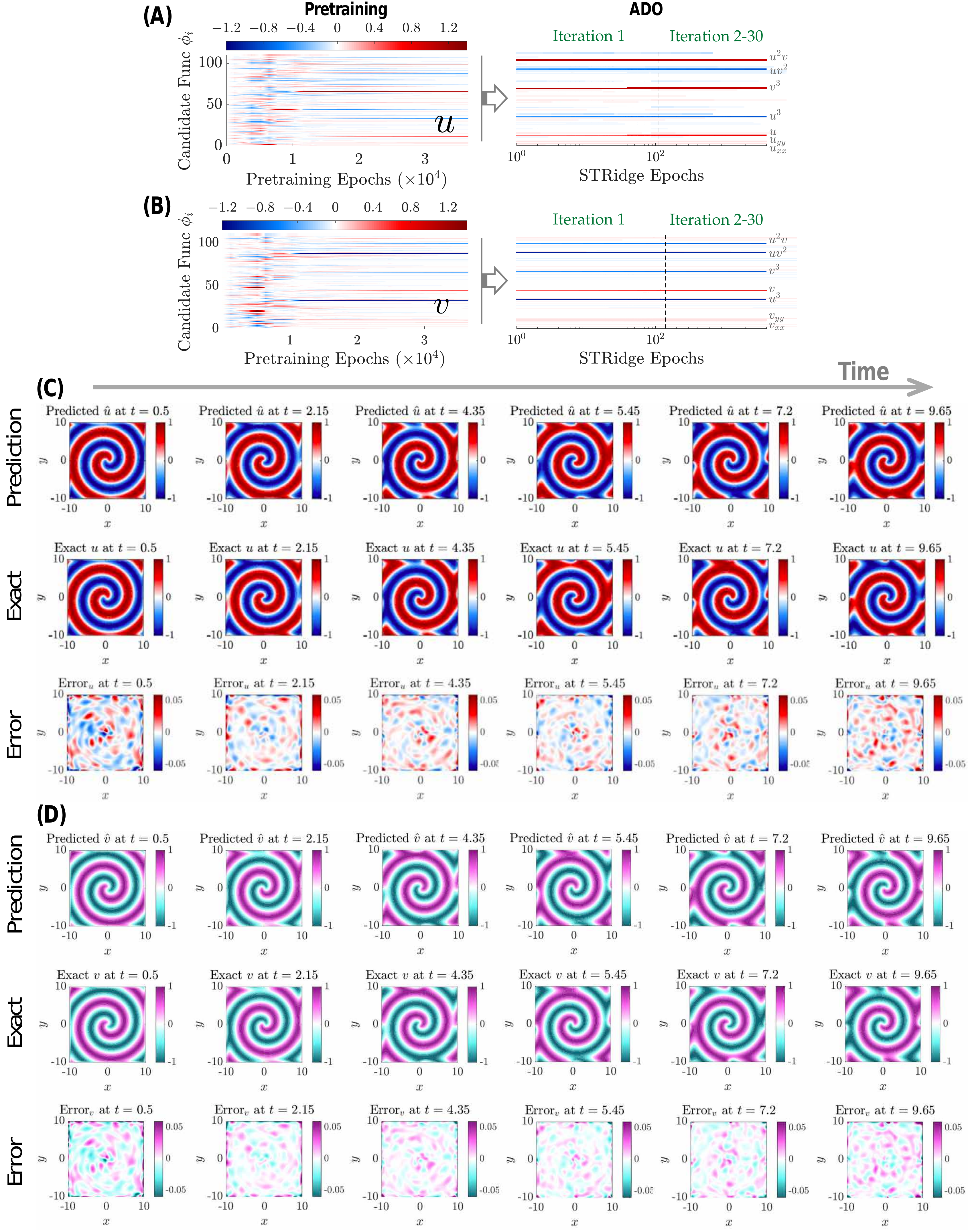}
\caption{Discovered $\lambda$-$\omega$ equations for a dataset with 10\% noise. (\textit{A} and \textit{B}) Evolution of the sparse coefficients $\boldsymbol{\lambda}^u\in\mathbb{R}^{110\times1}$ (\textit{A}) and $\boldsymbol{\lambda}^v\in\mathbb{R}^{110\times1}$ (\textit{B}) for 110 candidate functions $\boldsymbol{\phi}\in\mathbb{R}^{1\times110}$ used to reconstruct the $u$-equation and the $u$-equation, respectively. (\textit{C} and \textit{D}) The response snapshots $u$ (\textit{C}) and $v$ (\textit{D}) at different time instants ($t=0.5, 2.15, 4.35, 5.45, 7.2, 9.65$), showing the predictions and the exact solutions, as well as the prediction error maps. The relative full-field $\ell_2$ error of the prediction is about 2.21\%.}
\label{Sup_fig:Sup_RD}
\end{figure}

\subsubsection{$\lambda$-$\omega$ type Reaction-Diffusion equations}
The examples discussed previously are low-dimensional (1D) models with limited complexity. We herein consider a $\lambda$-$\omega$ reaction-diffusion (RD) system in a 2D domain with the pattern forming behavior governed by two coupled PDEs: 
\begin{equation*} 
\label{eq:RD}
\begin{cases}
u_t = 0.1u_{xx} + 0.1u_{yy} - uv^2 - u^3 + v^3 + u^2v +u \\
v_t = 0.1v_{xx} + 0.1v_{yy} - uv^2 - u^3 - v^3 - u^2v +v
\end{cases}
\end{equation*}
where $u$ and $v$ are two field variables. The $\lambda$-$\omega$ equations are typically used to describe the multi-scale phenomenon of local reactive transformation and the global diffusion in chemical reactions, with wide applications in pattern formation \cite{GUPTA2009399}, biological morphogenesis \cite{harrison1994kinetic}, and ecological invasions \cite{holmes1994partial}, among others. The $\lambda$-$\omega$ equations exhibit a wide range of behaviors including wave-like phenomena and self-organized patterns found in chemical and biological systems. The binomial system is also called an activator-inhibitor system because one state variable encourages the increase of both states while the other state component inhibits their growth. The particular $\lambda$-$\omega$ equations in this test example display spiral waves subjected to periodic boundary conditions. The domain for generating the solution is divided into 65,536 (256$\times$256) spatial points with 201 time steps. We take randomly 2,500 spatial points as fixed sensors recording the wave response for 15 randomly sampled time steps, leading to 1/4 of the subsampled dataset used in \cite{Rudye1602614} and 0.29\% of the total data. We sample $1\times10^5$  collocation points using the Sobol sequence \cite{Sobol} to evaluate the residual physics loss. A total of 110 candidate functions are employed, including polynomials up to the 3rd order ($u, v, u^2, v^2, uv, u^3, u^2v, uv^2, v^3$), derivatives up to the 2nd order ($u_x, u_y,v_x, v_y, u_{xx}, u_{xy}, u_{yy}, v_{xx}, v_{xy}, v_{yy}$) and their combination, for the sparse  discovery of the two PDEs. Since the system dimension is relatively high, we enhance the discovery by post-training (post-tuning) of the DNN and the uncovered non-zero PDE coefficients, after the ADO stage, resulting in refined/improved discovery. The DNN has 8 fully connected hidden layers and a width of 60 nodes in each layer. The pre-training takes $1\times10^4$ epochs of Adam (with additional L-BFGS tuning up to $1\times10^4$ epochs) followed by 30 ADO iterations. In each ADO iteration, we use $1\times10^3$ Adam epochs and up to $4\times10^3$ (depending on the relative loss decay) L-BFGS epochs to train the DNN for each alternation within STRidge.

The reconstructed equations for the case of 10\% noise are given by
\begin{equation*} 
\label{eq:RD_discovered}
\begin{cases}
u_t = 0.091u_{xx} + 0.092u_{yy} - 0.907uv^2 - 0.918u^3 + 0.997v^3 + 0.997u^2v + 0.917u \\
v_t = 0.099v_{xx} + 0.099v_{yy} - 1.006uv^2 - 1.003u^3 - 0.930v^3 - 0.935u^2v + 0.934v
\end{cases}
\end{equation*}
where the the average relative error for all non-zero coefficients is $4.78\pm3.66\%$. The evolution process illustrates that the sparse patterns are iteratively recovered out of a mixture of more than 100 candidate functions. At the end, both the sparse terms and the associated coefficients are precisely identified (as depicted in Fig. \ref{Sup_fig:Sup_RD}\textit{A} and \textit{B}). Due to the complexity of the PDEs and the high dimension, slightly more epochs are required in ADO to retain reliable convergence. The predicted response snapshots by the trained PiDL at different time instants, e.g., $t=\{0.5, 2.15, 4.35, 5.45, 7.2, 9.65\}$, are shown in in Fig. \ref{Sup_fig:Sup_RD}\textit{C} and \textit{D}, which are very close to the ground truth (the errors are distributed within a small range). This example shows especially the great ability and robustness of our method for discovering governing PDEs for high-dimensional systems from highly noisy data. The relative full-field $\ell_2$ error of the prediction is about 2.21\%.

\subsection{Comparison with SINDy}
We have performed the comparison study between the proposed PiDL approach and the state-of-the-art PDE-FIND method (an extended version of SINDy) \cite{Rudye1602614}, in the context of different levels of data size and noise. We test the five PDEs described previously and summarize the discovery errors for the sparse coefficients in Table \ref{Sup_discovery_result}. The error is defined as the average relative error of the identified non-zero PDE coefficients with respect to the ground truth. If the terms in the PDEs are discovered incorrectly, we mark it as ``NA'' (not applicable). It is seen from Table \ref{Sup_discovery_result} that the proposed PiDL approach is capable of correctly uncovering the closed-form PDEs for all cases, regardless of the varying levels of data size and noise, which demonstrates excellent robustness. Although PDE-FIND shows great success in PDE discovery with negligible error for large and clean (or approximately noise-free) measurement data, this method eventually fails when the level of data scarcity and/or noise increases. In general, PDE-FIND relies on the strict requirement of measurement quality and quantity. However, PiDL is able to alleviate and resolve this limitation thanks to the combination of the strengths of DNNs for rich representation learning of nonlinear functions, automatic differentiation for accurate derivative calculation as well as $\ell_0$ sparse regression. In addition, the use of collocation points introduces additional ``pseudo datasets'', compensates indirectly the scarcity of measurement data, and enriches the constraint for constructing the closed form of PDEs. Nonetheless, we have to mention that the proposal PiDL approach is much more computationally costly compared to PDE-PIND, primarily due to the training of DNNs. Fortunately, this issue can be well managed through parallel computing on a powerful GPU platform and remains a less important concern compared to the aim for successful discovery of correct underlying PDEs.

\begin{table*}[t!]
\footnotesize
\caption{Summary of the PiDL discovery results in comparison with PDE-FIND \cite{Rudye1602614} for canonical models.}
\begin{tabular}{llrrrr}
\toprule
PDE name & Method & Error (noise 0\%) & Error (noise 1\%) & Error (noise 10\%) & \# of Measurement points  \\
\midrule
Burgers' & PiDL & 0.01$\pm$0.01\% & 0.19$\pm$0.11\% & \colorbox{mygray}{1.15$\pm$1.20\%} & $\sim$505  \\
\cmidrule{2-6}
 & PDE-FIND & NA & NA & NA & $\sim$505 \\
 & & 0.15$\pm$0.06\% & 0.80$\pm$0.60\% & NA & $\sim$26K \\
\midrule
KS & PiDL & 0.07$\pm$0.01\% & 0.61$\pm$0.04\% & \colorbox{mygray}{0.71$\pm$0.06\%} & $\sim$32K \\
\cmidrule{2-6}
 & PDE-FIND & 35.75$\pm$16.30\% & NA & NA & $\sim$32K \\
 & & 1.30$\pm$1.30\% & 52.00$\pm$1.40\% & NA & $\sim$257K \\
\midrule
Schr\"{o}dinger & PiDL & 0.09$\pm$0.04\% & 0.65$\pm$0.29\% & \colorbox{mygray}{2.31$\pm$0.28\%} & $\sim$96K \\
\cmidrule{2-6}
 & PDE-FIND & NA & NA & NA & $\sim$96K \\
 & & 0.05$\pm$0.01\% & 3.00$\pm$1.00\% & NA & $\sim$257K \\
 \midrule
NS & PiDL & 0.66$\pm$0.72\% & 0.86$\pm$0.63\% & \colorbox{mygray}{1.40$\pm$1.83\%} & $\sim$30K \\
\cmidrule{2-6}
 & PDE-FIND & NA & NA & NA & $\sim$30K \\
 & & 1.00$\pm$0.20\% & 7.00$\pm$6.00\% & NA & $\sim$300K \\
\midrule
$\lambda$-$\omega$ RD & PiDL & 0.07$\pm$0.08\% & 0.25$\pm$0.30\% & \colorbox{mygray}{4.78$\pm$3.66\%} & $\sim$37.5K \\
\cmidrule{2-6}
 & PDE-FIND & NA & NA & NA & $\sim$37.5K \\
 & & 0.02$\pm$0.02\% & NA & NA & $\sim$150K \\ 
\bottomrule
\end{tabular}\label{Sup_discovery_result}
\begin{tablenotes}
{\scriptsize
\item Note: In the table, KS, NS and RD refer to the Kuramoto-Sivashinsky, Navier-Stokes and the $\lambda$-$\omega$  Reaction-Diffusion PDEs. Gaussian white noise is added to the synthetic response with the noise level defined as the root-mean-square ratio between the noise and exact solution. NA denotes ``not applicable'' (e.g., failure in correct identification of the sparse PDE coefficients). The identification error is defined as the average relative error of the identified non-zero PDE coefficients with respect to the ground truth.}
\end{tablenotes}
\end{table*}

\subsection{Discovery of PDEs with Multiple Independent Datasets}
\setcounter{subsubsection}{0}
\subsubsection{Burgers' equation with shock behavior}\label{Burgers_3IBCs}
\begin{figure*}[t!]
\centering
\includegraphics[width=0.8\linewidth]{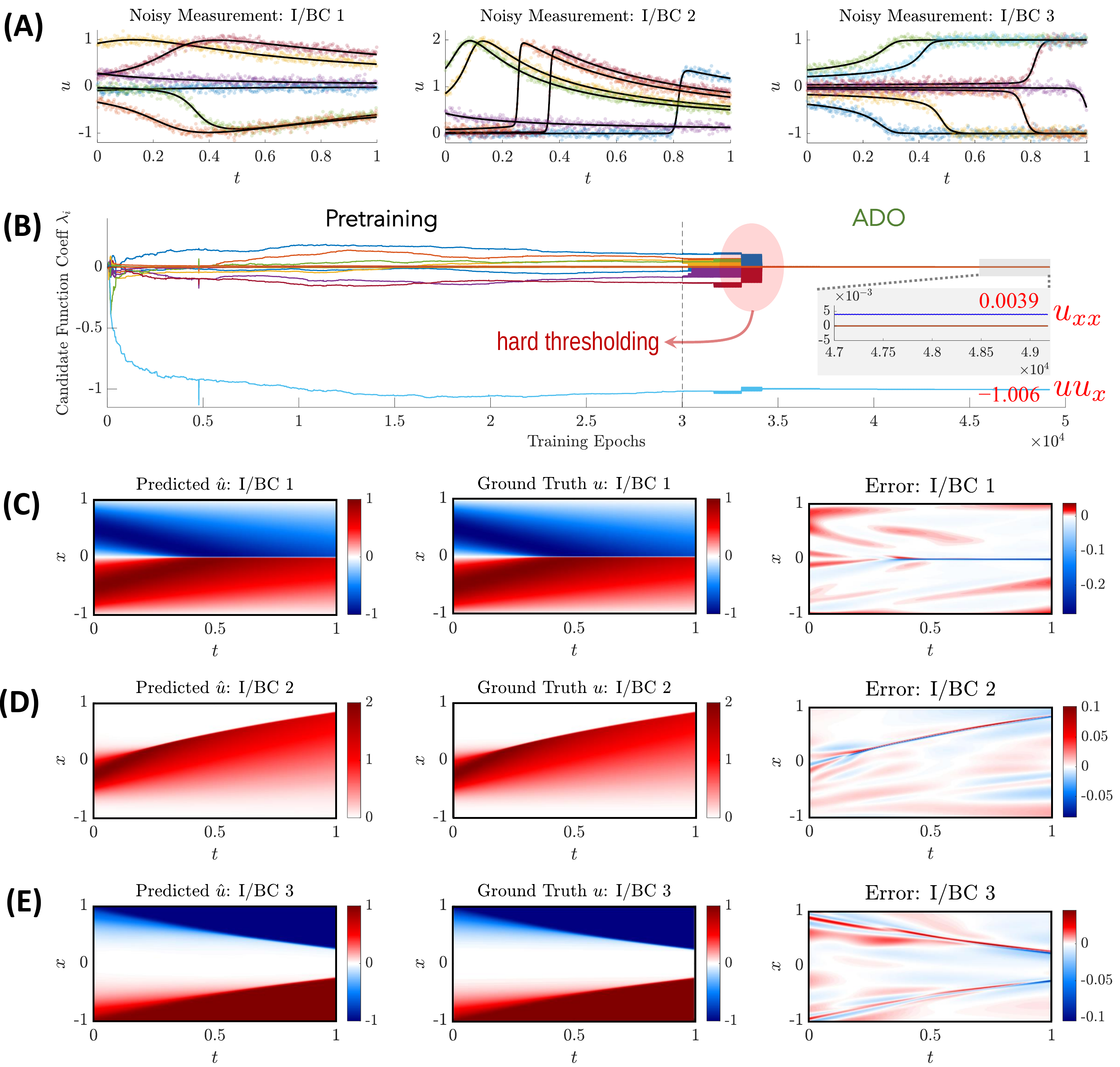}
\caption{Discovered Burgers' equation with small viscosity based on datasets sampled under three IBCs with 10\% noise. (\textit{A}) Visualization of noisy measurements for the three datasets. Note that there are 30 sensors and only a few are illustrated in this figure. (\textit{B}) Evolution of the sparse coefficients $\boldsymbol{\Lambda}\in\mathbb{R}^{16\times1}$ for 16 candidate functions $\boldsymbol{\phi}\in\mathbb{R}^{1\times16}$ used to construct the PDE, where the color represents the coefficient value. The correct terms ($uu_{x}$ and $u_{xx}$) and their coefficients are successfully identified while other redundant terms are eliminated by ADO. (\textit{C}-\textit{E}) The predicted response in comparison with the exact solution for three IBCs. The relative full-field $\ell_2$ error of all the stacked predictions is 2.24\%.}
\label{Sup_fig:Sup_Burgers_MultiIC}
\end{figure*}

We consider to discover the previously discussed Burgers' equation (see Section \ref{Burgers}) with a small diffusion/viscosity parameter, expressed as
\begin{equation*} 
u_t = -uu_x + \frac{0.01}{\pi} u_{xx}
\end{equation*}
based on datasets generated by imposing three different IBCs. The small diffusion coefficient $0.01/\pi\approx0.0032$ creates creates shock formation in a compact area with sharp gradient and poses notorious difficulty for many numerical methods to resolve, which could challenge the DNN's approximation ability and thus affect the discovery. The three IBCs used for data generation include:
\begin{align*}
&\text{IBC 1:~} u(x, 0) = -\sin(\pi x), u(-1, t) = u(1, t) = 0 \\
&\text{IBC 2:~} u(x, 0) = \mathcal{G}(x), u(-1, t) = u(1, t) = 0 \\
&\text{IBC 3:~} u(x, 0) = -x^3, u(-1, t) = 1, u(1, t) = -1
\end{align*}
where $\mathcal{G}$ denotes a Gaussian function. The ground truth solution is simulated by MATLAB function \texttt{pdede} in a spatiotemporal domain $\Omega \times [0, T] = [-1, 1]_{d=200} \times [0, 1]_{d=1000}$. For all IBCs, we assume that there are 30 sensors randomly deployed in space measuring the wave traveling (e.g., $u$) for 500 time instants (7.5\% of the total grid points). A denser sensor grid is needed herein, compared with the previous Burgers' example, in order to capture the shock behaviors. All measurements are polluted with 10\% Gaussian noise. The noisy measurements are depicted in Fig. \ref{Sup_fig:Sup_Burgers_MultiIC}\textit{A} for the three datasets. For visualization purpose, we only draw a handful of signals out of a total of 30 time series for each IBC. 

We design a ``root-branch'' DNN: the root takes the spatiotemporal coordinates $\{x, t\}$ as input followed by 4 hidden layers of 20 nodes, while each of the three branches is separately connected to the last hidden layer of the root followed by 4 hidden layers of 30 nodes before the output layer. The motivation for this design is that the branch nets can capture the solution difference due to different IBCs while the shared root net learns the common response patterns that obey the unique Burgers' equation. Note that although we have three distinctive solution approximations, we stack them into one candidate library followed by a unified form of PDE. Therefore, we can combine the information from three datasets to discover one physics equation. A group of $4.5 \times 10^4$ collocation points are generated by the Latin hypercube sampling strategy \cite{McKay} for determining the residual physics loss. The PDE library consists of 16 candidate functions, exactly the same as the Burgers' case in Section \ref{Burgers}. The training efforts include $3 \times 10^4$ epochs pretraining by L-BFGS followed by 6 ADO iterations. In each ADO iteration, we use $1 \times 10^3$ Adam epochs (with an initial learning rate of 1e-4 and drops by 50\% every $1 \times 10^3$ Adam epochs) and up to $3 \times 10^4$ L-BFGS epochs (depending on relative loss decay) to train the DNN in synergy with STRidge. The discovered PDE is given by
\begin{equation*} 
u_t = -1.006uu_x + 0.0039u_{xx}
\end{equation*}
which shows great agreement with the ground truth. Fig. \ref{Sup_fig:Sup_Burgers_MultiIC}\textit{B} depicts the evolution of the coefficients ($\boldsymbol{\Lambda}\in\mathbb{R}^{16\times1}$) of candidate functions, where the correct terms in the library ($uu_{x}$ and $u_{xx}$) are successfully distilled while other redundant terms are eliminated (e.g., hardly thresholded to zero) by ADO. The coefficients of the active terms are accurately identified as well (in particular the small viscosity parameter that leads to shock formation, e.g., 0.0039). While the $\ell_1$ penalty lays the foundation for sparsity in pretraining stage, while ADO finds the correct sparsity pattern in after the first iteration and refines the coefficients in the subsequent iterations. Fig. \ref{fig:Burgers_MultiIBC}c-d show the predicted responses and errors for three IBC cases, with a stacked full-field $\ell_2$ error of 2.24\%. The trained ``root-branch'' network can accurately reproduce distinctive system responses even with limited measurements under 10\% noise, giving a full-field $\ell_2$ error of 2.24\%, as shown in Fig. \ref{Sup_fig:Sup_Burgers_MultiIC}\textit{C-E}.

\subsubsection{Fitzhugh-Nagumo type of Reaction-Diffusion equations}
We consider the Fitzhugh-Nagumo (FN) type reaction-diffusion system, in a 2D domain $\Omega = [0, 150]\times [0, 150]$ with periodic boundary conditions, whose governing equations are expressed by two coupled PDEs \cite{fitzhugh1961impulses, nagumo1962active}:
\begin{align*} 
u_t &= \gamma_u \Delta u + u - u^3 - v + \alpha \\
v_t &= \gamma_v \Delta v + \beta(u - v)
\end{align*}
where $u$ and $v$ represent two interactive components/matters (e.g., biological), $\gamma_u = 1$ and $\gamma_v = 100$ are diffusion coefficients, $\alpha = 0.01$ and $\beta = 0.25$ are the coefficients for reaction terms, and $\Delta$ is the Laplacian operator. The FN equations are commonly used to describe biological neuron activities excited by external stimulus ($\alpha$), which exhibit an activator-inhibitor system because one equation boosts the production of both components while the other equation dissipates their new growth. The ground truth data is generated by the finite difference method ($dx=dy=0.5$ and $dt=0.0002$) for the time period of $[0, T]=[0, 36]$, with three random fields as initial conditions. Three measurement datasets are then generated, each of which consists of 31 low-resolution snapshots (projected into a $31\times31$ grid) uniformly down-sampled from full-field synthetic data during the period of $[7.18, 36]$ under a 10\% noise condition. SImilar to the previous example in Section \ref{Burgers_3IBCs}, we design a ``root-branch'' DNN with three branches: the root net has 2 hidden layers of 60 nodes while each of the three branch nets has 3 hidden layers of 60 nodes. We sample $5 \times 10^4$ spatiotemporal collocation points using Latin hypercube sampling \cite{McKay} to construct the physics residuals. 

\begin{figure*}[htbp]
\centering
\includegraphics[width=0.95\linewidth]{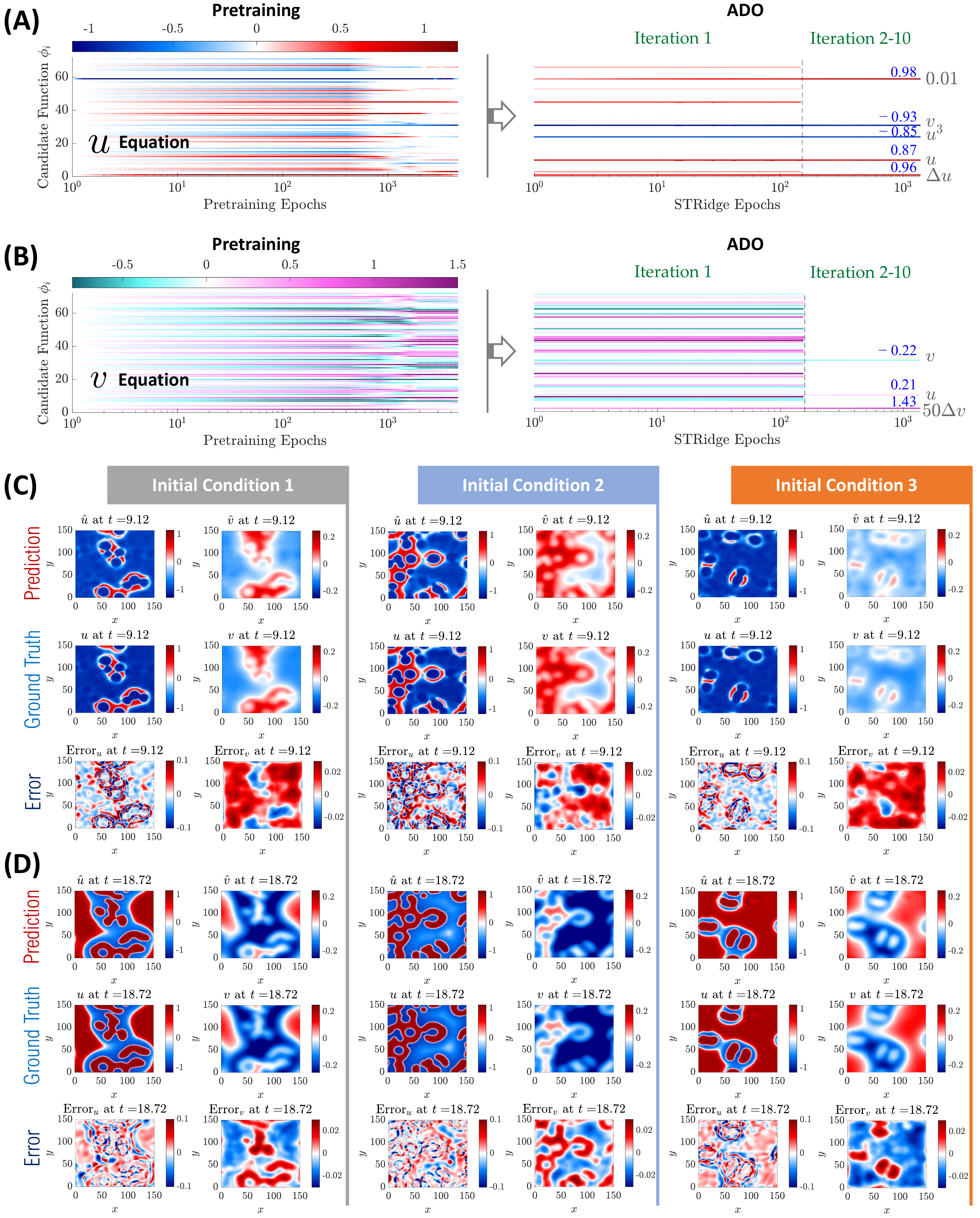}
\caption{Discovered Fitzhugh-Nagumo equations based on measurements sampled under three initial conditions (ICs) with 10\% noise. (\textit{A}) Evolution of the sparse coefficients $\boldsymbol{\lambda}_u\in\mathbb{R}^{72\times1}$ for 72 candidate functions used to construct the first PDE ($u$-equation), where the color represents the coefficient value. (\textit{B}) Evolution of the sparse coefficients $\boldsymbol{\lambda}_v\in\mathbb{R}^{72\times1}$ for the second PDE ($v$-equation). Note that, for visualization purpose, we re-scale the identified coefficients of the constant stimulus term ``1'' in the $u$-equation by multiplying 100 in \textit{A} and the diffusion term $\Delta v$ in the $v$-equation by dividing 50 in \textit{B}. (\textit{C-D}) Snapshots of predicted response, ground truth and error distributions for all three ICs at two unmeasured time instances ($t=9.12$ and $t=18.72$). The relative $\ell_2$ error for the predicted full-field response (stacked $u$ and $v$) is 5.04\%.}
\label{Sup_fig:Sup_FN_MultiIC}
\end{figure*}

We assume the diffusion terms ($\Delta u$ and $\Delta v$) are known in the PDEs, whose coefficients ($\gamma_u$ and $\gamma_v$) yet need to be identified. We employ the bounds to these two positive coefficients to speed up the convergence, namely, $\gamma_u\in[0, 5]$ and $\gamma_v\in[0, 150]$. We design 70 candidate functions, composed of up to third-order polynomials (including the constant term ``1'' as the zero order), derivatives $\{u_x, u_y, u_{xy}, v_x, v_y, v_{xy}\}$ and their mutual multiplication, to reconstruct the nonlinear reaction terms in the PDEs. Hence, the final library has 72 candidate terms. To account for the small stimulus term (e.g., 0.01 in the first equation), we increase the sensitivity of the constant candidate ``1'' in the library by down-scaling its magnitude to the order of $10^{-5}$ and $5\times10^{-4}$ for $u$ and $v$ equations respectively. The training efforts include the pretraining stage with $6 \times 10^3$ Adam epochs and $4 \times 10^4$ L-BFGS epochs, 10 ADO iterations, and an extra post-training with $1 \times 10^5$ Adam epochs. In each ADO iteration, we use $1 \times 10^4$ Adam epochs and up to $1 \times 10^4$ L-BFGS epochs in synergy with STRidge. To deal with the aforementioned bounds for $\gamma_u$ and $\gamma_v$ in an unconstrained optimization process, we set $\gamma_u = 5\sigma(\tilde{\gamma}_u)$ and $\gamma_v = 150\sigma(\tilde{\gamma}_v)$ and take $\{\tilde{\gamma}_u, \tilde{\gamma}_v\}$ as trainable variables, where $\sigma(\cdot)$ denotes the Sigmoid function. The discovered equations under 10 \% noise is
\begin{align*} 
u_t &= 0.962 \Delta u + 0.874u - 0.847u^3 - 0.931v + 0.0098 \\
v_t &= 71.515 \Delta v + 0.214u - 0.224v
\end{align*}
It is seen that the form of the PDEs is precisely uncovered with all correct active terms (including the unknown external stimulus in the first equation). The corresponding identified coefficients are generally close to the ground truth (error of non-zero coefficients: $11.72 \pm 8.34 \%$) except the diffusion coefficient for $v$ (i.e., $\gamma_v$) which seems to be a less sensitive parameter according to our test. It should be noted that, given very scarce and noisy measurement datasets in this example, the ``root-branch'' DNN is faced with challenges to accurately model the solutions with sharp propagating fronts (see Fig. \ref{Sup_fig:Sup_FN_MultiIC}\textit{C}-\textit{D}). The less accurate solution approximation by DNN then affects the discovery precision. This issue can be naturally alleviated by increasing the spatiotemporal measurement resolution (even still under fairly large noise pollution, e.g., 10\%). Nevertheless, the exact form of the PDEs is successfully discovered in this challenging example, which is deemed more important since the coefficients can be further tuned/calibrated when additional data arrives. The evolution of the PDE coefficients corresponding to 72 candidate functions for $\hat{u}$ and $\hat{v}$ is illustrated in Fig. \ref{Sup_fig:Sup_FN_MultiIC}\textit{A} and \textit{B}, respectively. Note that, for visualization purpose, we re-scale the identified coefficients of the constant stimulus term ``1'' in the $u$-equation by multiplying 100 in Fig. \ref{Sup_fig:Sup_FN_MultiIC}\textit{A} and the diffusion term $\Delta v$ in the $v$-equation by dividing 50 in Fig. \ref{Sup_fig:Sup_FN_MultiIC}\textit{B}. The trained network is finally used to predict the full-field responses under three IBCs (see the snapshots in Fig. \ref{Sup_fig:Sup_FN_MultiIC}\textit{C}-\textit{D} at two unmeasured time instants). The stacked full-field $\ell_2$ error is 5.04\%.

\subsection{Experimental Discovery of Cell Migration and Proliferation}
In this example, we consider to discover a biological system based on scratch assay experiments \cite{jin2016reproducibility} investigating the cell migration and proliferation process. The 1D cell density distributions at different time instants (0h, 12h, 24h, 36h, 48h) were extracted and simplified from high-resolution imaging via image segmentation and cell counting (see Extended Data Fig. {\color{blue} 3} in the {\color{blue}Main Text}). A series of assays were performed under different initial cell densities (e.g., the total number of cells spans from 10,000 to 20,000 following the designated initial distribution in the test well. More detailed description of the experiment setup and datasets can be found in \cite{jin2016reproducibility}. 

Our objective herein is to uncover a parsimonious PDE for modeling the dynamics of cell density $\rho(x, t)$. Here, we consider four scenarios with the initial number of cells ranging from 14,000, 16,000, 18,000 to 20,000. We take the mean of the test data from three identically-prepared experimental replicates for each scenario for PDE discovery. Each mean dataset has a total of $38 \times 5$ measurement points for five time instances. Given our prior knowledge that the cell dynamics can be described by a diffusion (migration) and reaction (proliferation) process, we assume the PDE holds the form of $\rho_t = \gamma\rho_{xx} + \mathcal{F}(\rho)$, where $\gamma$ is the unknown diffusion coefficient and $\mathcal{F}$ denotes the underlying nonlinear reaction functional. We use 8 additional candidate terms (e.g., $\{1, \rho, \rho^2, \rho^3, \rho_x, \rho\rho_x, \rho^2\rho_x, \rho^3\rho_x\}$) to reconstruct $\mathcal{F}$, whose coefficients are sparse. Hence, the total number of trainable coefficients remains 9 (e.g., $\boldsymbol{\Lambda}\in\mathbb{R}^{9\times1}$).

\begin{figure*}[t!]
\centering
\includegraphics[width=0.90\linewidth]{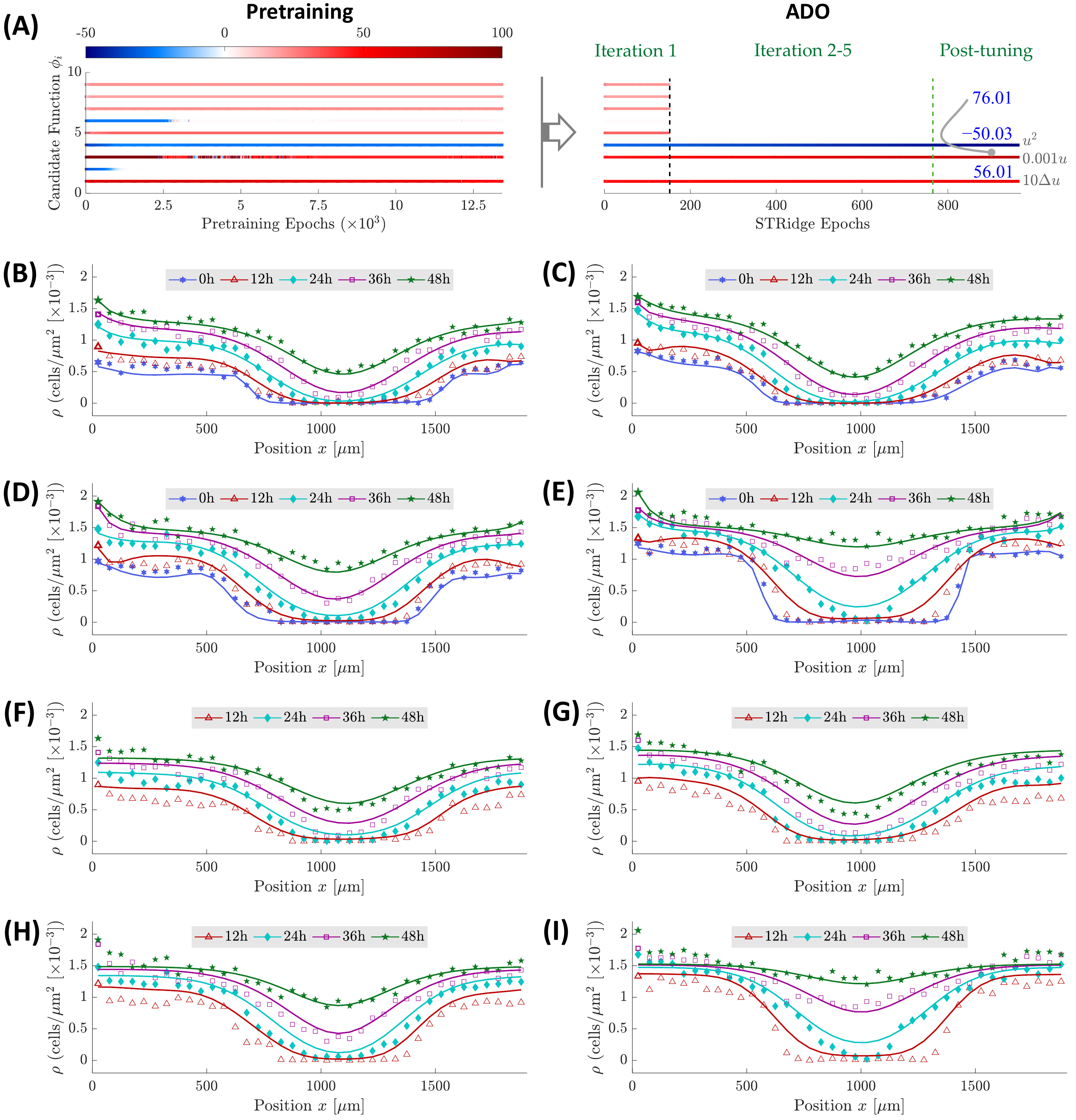}
\caption{Experimental discovery of cell migration and proliferation. (\textit{A}) Example evolution of the sparse coefficients $\boldsymbol{\Lambda}\in\mathbb{R}^{9\times1}$ for 9 candidate functions used to construct the underlying PDE for the case of 18,000 cells. The diffusion and reaction coefficients for $\Delta u$ and $u$ are re-scaled for visualization purpose. (\textit{B})-(\textit{E}) Predicted cell densities (represented by solid curves) by the trained DNNs in comparison with the measurement data (denoted by markers) for 14,000, 16,000, 18,000 and 20,000 cells, respectively. (\textit{F})-(\textit{I}) Simulated cell densities, represented by solid curves, at different time instants based on the discovered PDEs for 14,000, 16,000, 18,000 and 20,000 cells, respectively, where the measurement at 0h is used as the initial condition while $\rho_x(x=0, t)=\rho_x(x=1900, t)=0$ is employed as the Neumann boundary condition. The simulation result is represented by solid curves while the markers denote the measurement data.}
\label{Sup_fig:Supp_Cell_result}
\end{figure*}

We sample $1\times10^4$ collocation pairs using Latin hypercube sampling \cite{McKay} in the spatiotemporal domain of $\Omega \times [0, T] = [0, 1900]\mu \mathrm{m} \times [0, 48]\mathrm{h}$. The DNN has 3 hidden layers of 30 nodes activated by by the tanh function (see Fig. {\color{blue} 1} in the {\color{blue}Main Text}). Considering that the cell density is constantly positive, we impose a \textit{softplus} function (e.g., $\ln(1+\mathrm{e}^z)$) in the output layer to curb the final output of $\rho$. To account for potential large magnitude variation of the candidate term coefficients, we apply the sigmoid and tanh functions to squash magnitude gaps. Specifically, we set $\gamma  = 1000\text{sig}(\tilde{\gamma})$ and $\boldsymbol{\lambda} = 50\text{tanh}(\tilde{\boldsymbol{\lambda}})$, where $\tilde{\gamma}$ and $\tilde{\boldsymbol{\lambda}}$ are trainable ``proxies'' for diffusion coefficient $\gamma$ and other 8 coefficients $\boldsymbol{\lambda}$ (note: $\boldsymbol{\Lambda} = \{\gamma, \boldsymbol{\lambda}\}\in\mathbb{R}^{9\times1}$). The training efforts include the pretraining stage with $8 \times 10^3$ Adam epochs and $8 \times 10^3$ L-BFGS epochs, 5 ADO iterations, and extra post-training with $1 \times 10^5$ Adam epochs. In each ADO iteration, we use $2 \times 10^3$ Adam epochs in synergy with STRidge. Fig. \ref{Sup_fig:Supp_Cell_result}\textit{A} shows the evolution of 9 coefficients for the example case of 18,000 cells, where redundant candidate terms are pruned right after the first ADO iteration via hard thresholding of the corresponding coefficients to zero. The next ADO iterations followed by post-tuning refine the coefficients of active terms for final reconstruction of the PDE. The discovered underlying PDEs under different initial cell states are given as follows:
\begin{align*} 
&\text{14k cells:} \; \; \rho_t = 553.05\rho_{xx} + 0.067 \rho - 48.14 \rho^2\\ 
&\text{16k cells:} \; \; \rho_t = 546.45\rho_{xx} + 0.066 \rho - 44.10 \rho^2\\ 
&\text{18k cells:} \; \; \rho_t = 560.12\rho_{xx} + 0.076 \rho - 50.03 \rho^2\\ 
&\text{20k cells:} \; \; \rho_t = 686.61\rho_{xx} + 0.092 \rho - 60.02 \rho^2
\end{align*}
which share a unified form of $\rho_t = \gamma\rho_{xx} + \lambda_1\rho + \lambda_2\rho^2$ which exactly matches the famous Fisher-Kolmogorov model \cite{fisher1937wave,maini2004traveling}. The rates of migration (diffusion) and proliferation (reaction) generally increase along with the number of cells, as seen from the identified coefficients. Fig, \ref{Sup_fig:Supp_Cell_result}\textit{B}-\textit{E} depict the learned cell density profiles by the trained DNN, which capture the critical patterns of the measurement while showing little evidence of overfitting. With the discovered PDEs, we simulate/predict the evolution of cell densities at different time instants (12h, 24h, 36h and 48h) presented in Fig. \ref{Sup_fig:Supp_Cell_result}\textit{F}-\textit{I}, where the measurement at 0h is used as the initial condition while $\rho_x(x=0, t)=\rho_x(x=1900, t)=0$ is employed as the Neumann boundary condition. The satisfactory agreement between the prediction and the measurement provides a clear validation of our discovered PDEs.

\section{Discussion}
In this section, we discuss several other features, influence factors and limitations of the proposed PiDL method for data-driven discovery of PDEs, and highlight the potential future work. 

\subsection{Selection of candidate functions}
The library of candidate functions is a significant component in PiDL, similar to the SINDy framework. On one hand, we prefer to make the candidate library as diverse as possible. On the other hand, balancing the increasing theoretical and computational complexity is crucial for applications. We believe that a specialized library hinged by our domain-specific knowledge and statistical experience can constrain the search space and reduce the complexity of PDE discovery. Although the higher the dimension of the library is, the more likely the exact terms will be uncovered from data. Nevertheless, a highly large-scale library (e.g., the number of components on the order of magnitude of $\geq10^3$), essentially approximated by the DNN, is very likely to be rank deficient and have poor conditioning, in addition to the growing theoretical complexity and computational burden. Balancing these concerns and finding mathematical principles based on domain-specific knowledge to establish an efficient candidate library remain an open problem. Noteworthy, failing to include essential candidate functions will lead to false positive discovery of parsimonious closed form of PDEs, despite that a ``best-of-fit'' form can be found. Alternatively, we can first include rich polynomial terms and discover the governing PDEs in an approximate form, followed by Taylor series analysis \cite{WU2019200} and power-law classes \cite{Daniels2015} to infer a more parsimonious form. This will also help inform the redesign and enrichment of the library of candidate functions for potentially improved discovery.

\begin{figure}[t!]
\centering
\includegraphics[width=1\textwidth]{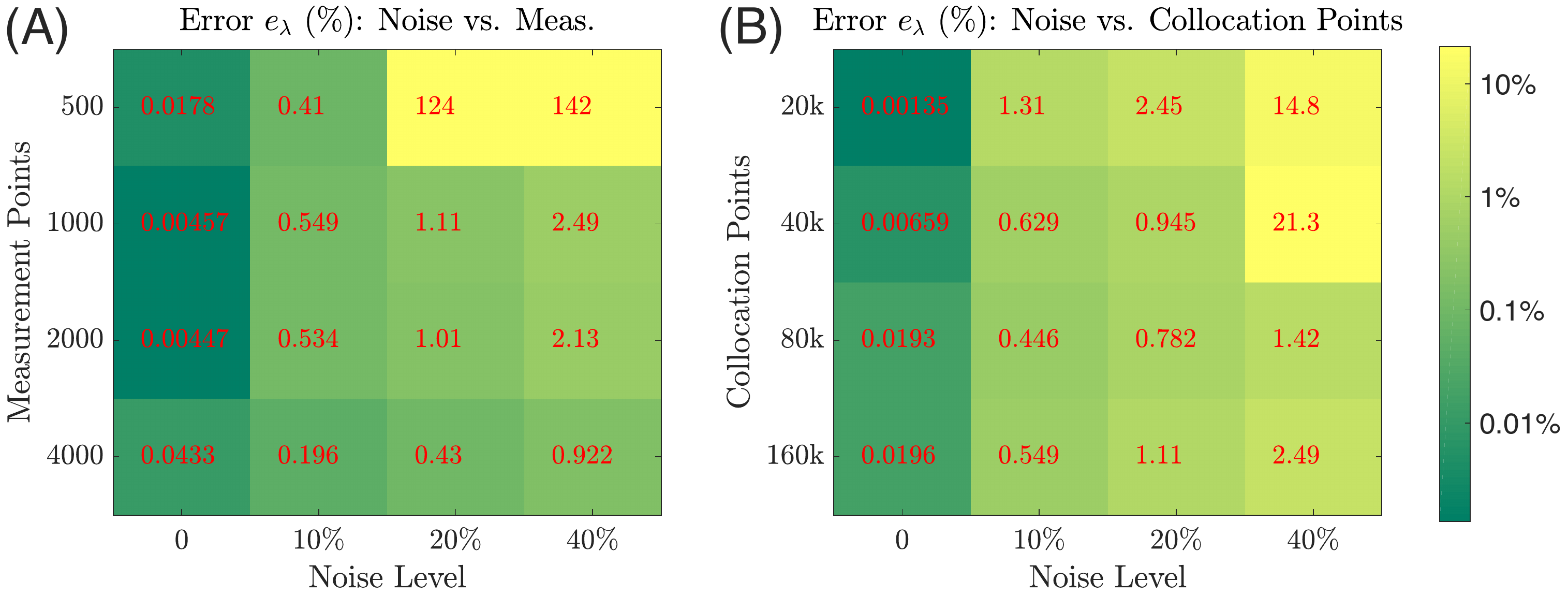}
\caption{Error $e_{\lambda}$ for discovery of Burgers' equation under different measurement points, collocation points and noise levels. Numbers in each cell denote the percentage error of $e_{\lambda}$ for a specific condition, which is the relative $\ell_2$ norm error between the identified coefficients $\boldsymbol{\Lambda}$ and the ground truth $\boldsymbol{\Lambda}_\text{true}$. The color also indicates the error level. The collocation points are fixed at $1.6\times10^5$ in (\textit{A}), while the measurement points are always $1\times10^3$ in (\textit{B}).}
\label{Sup_Fig:Meas_Collo_Noise}
\end{figure}

\subsection{Noisy measurements and collocation points}
The total loss function is evaluated on the measurement data (for $\mathcal{L}_d$) and the collocation points (for $\mathcal{L}_p$). Therefore, the availability of noisy measurement data and the number of collocation points sampled from the spatiotemporal space will affect the convergence of the PiDL model and thus the PDE discovery accuracy. We herein study the sensitivity of PiDL to these factors in the context of discovery accuracy based on the Burgers' equation example. In particular, we use the relative $\ell_2$-norm error to reflect the global accuracy of of the identified sparse coefficients, defined as $e_{\lambda} = ||\hat{\boldsymbol{\Lambda}}-\boldsymbol{\Lambda}_\text{true}||_2 / ||\boldsymbol{\Lambda}_\text{true}||_2$ where $\hat{\boldsymbol{\Lambda}}$ denotes the identified coefficients and $\boldsymbol{\Lambda}_\text{true}$ is the ground truth. Fig. \ref{Sup_Fig:Meas_Collo_Noise} shows the error metrics for discovering the Burgers' equation under different quantities of measurement points and collocation points and noise levels. Increasing the number of data points in the measurement (e.g., recorded by more sensors) can well compensate the noise effect as shown in Fig. \ref{Sup_Fig:Meas_Collo_Noise}\textit{A} (the number of collocation points is fixed at $1.6\times10^5$), which agrees with our common sense. Although optimal sensor placement might alleviate the need of large datasets \cite{Manohar2018}, this is out of the scope of this work. The use of more collocation points can mitigate the noise effect and improve the discovery accuracy as illustrated in Fig. \ref{Sup_Fig:Meas_Collo_Noise}\textit{B} (the number of measurement points is fixed at $1\times 10^3$). For this specific case, $2\times 10^4$ (or more) collocation points are able to maintain a satisfactory discovery accuracy for measurements under noise corruption at a realistic level (e.g., $\leq20\%$). When the data are sampled under a very noisy condition (e.g., 40\% noise level), the proposed method is still robust if a larger number of collocation points are used (e.g., $\geq8\times 10^4$).

It is noteworthy that the collocation points require no correlation with the measurement data. In particular, we use the Sobol sequence \cite{Sobol} (or Latin hypercube sampling \cite{McKay} which is also applicable) to simulate a finer uniform partitions of the problem domain, making the random sampling of collocation points more representative. Intuitively, the more the collocation points are used, the more generalizable the trained network will be and the more accurate the discovered PDE is. However, a large number of collocation points also impose heavy computational burden, limited by hardware resources. A fairly large amount of collocation samples (e.g., on the order of magnitude of $>10^4$), comparable to the complexity and dimension of the discovery problem, are suggested in practical applications meanwhile considering the memory of the computing machine.

\begin{figure}[t!]
\centering
\includegraphics[width = 0.65\textwidth]{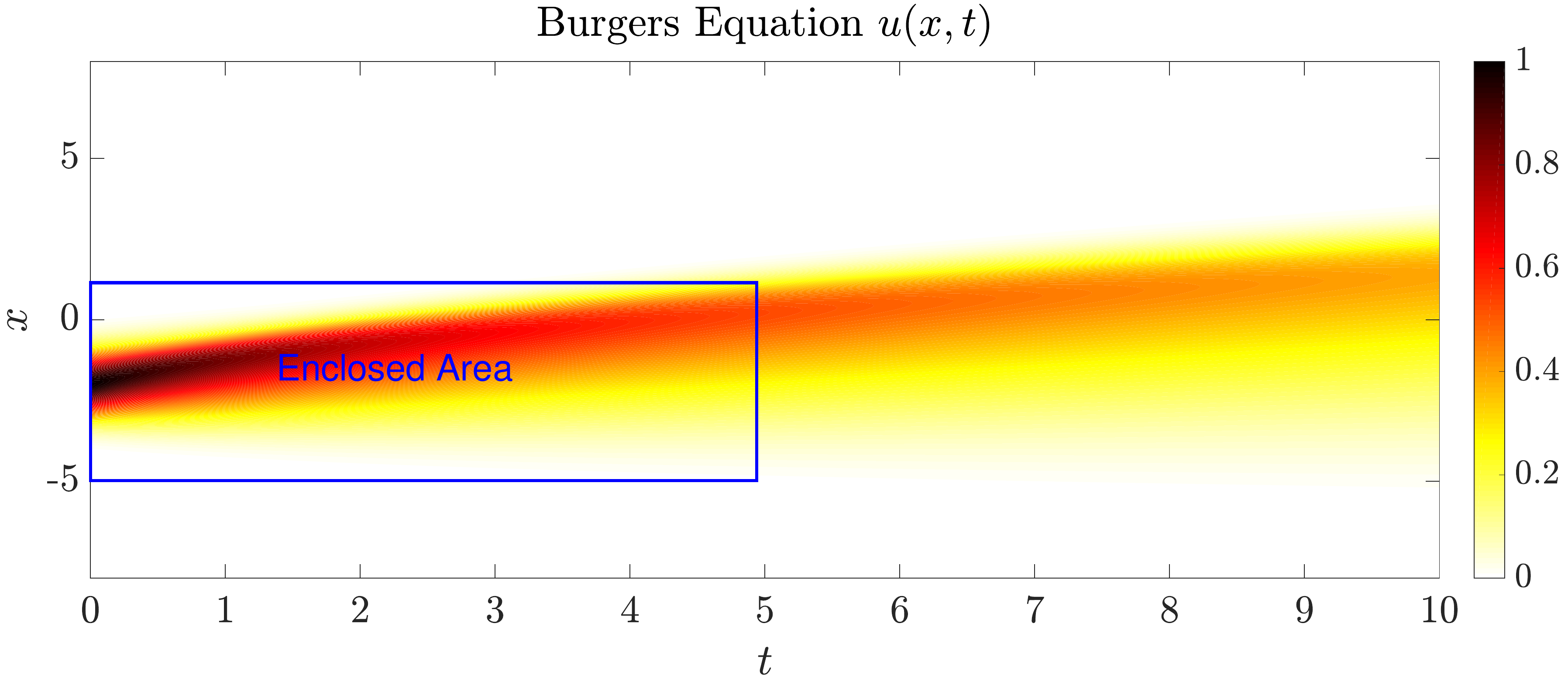}
\caption{Parametric study on the effect of collocation points for discovering the Burgers' equation. The measurements are only taken from the enclosed area, while the collocation points are sampled across the full field.}
\label{Sup_Fig:Burgers_EnclosedAreas}
\end{figure}

\begin{table}[t!]
\scriptsize
\caption{On the extrapolation (generalization) ability of PiDL}\label{table:EnclosedAreaStudy}
\begin{tabular}{l r r r r r r r}
\toprule
Case & Meas. points & Collocation points & Noise level & Training error & Validation error & Full-field error & $\ell_2$ error of $\boldsymbol{\Lambda}$\\
\midrule
1 & $1.5\times10^3$ & $8\times10^4$ & 0 & 0.03\% & 0.04\% & 2.41\% & 0.02\% \\

2 & $1.5\times10^3$ & $8\times10^4$ & 10\% & 5.73\% & 5.95\% & 4.50\% & 0.79\% \\

3 & $1.5\times10^3$ & 0 & 0 & 58.79\% & 60.19\% & 144.99\% & 142.66\% \\

4 & $3\times10^3$ & 0 & 0 & 0.10\% & 0.10\% & 14.36\% & 0.38\% \\
\bottomrule
\end{tabular}
\begin{tablenotes}
\item Note: The training error, validation error and full-field error are calculated in the form of $\|\hat{\mathbf{u}}-\mathbf{u}_\text{true}\|_2/\|\mathbf{u}_\text{true}\|_2$, where $\hat{\mathbf{u}}$ denotes the DNN-predicted response and $\mathbf{u}_\text{true}$ is the reference ground truth solution.
\end{tablenotes}
\end{table}

We further conduct a comparative study on the role of collocation points and seek for numerical understanding of how much they can help for extrapolation. Taking the Burgers' equation for instance, we define an enclosed area, part of the full-field response, as shown in Fig. \ref{Sup_Fig:Burgers_EnclosedAreas}, and sample the measurement data only within such an area. We intend to reconstruct the full-field response beyond the enclosed area and discover the PDE taking advantage of collocation points. More specifically, the enclosed area is meshed by $100\times50$ spatiotemporal points. We take 30 randomly selected spatial locations as fixed sensors recording the dynamic response of the system, resulting in $1.5\times10^3$ data points. Additionally, we sample $8\times10^4$ collocation points from the full spatiotemporal field for evaluating the residual physics loss during model training. Four cases are considered to demonstrate the function of collocation points with measurements sampled in the enclosed area (see Table \ref{table:EnclosedAreaStudy}). 

Provided with clean measurements from the enclosed area and global collocation points, PiDL does an impressive job on both full-field response reconstruction and sparse coefficients identification (see Case 1 in Table \ref{table:EnclosedAreaStudy}). When the measurements become noisy (e.g., 10\% noise level), despite the response prediction errors increase, the PDE can still be accurately discovered (see Case 2 in Table \ref{table:EnclosedAreaStudy}). If we consider removing all collocation points and only train the network with clean measurements, the response prediction errors (even during the training and validation stage) all remain over 50\%, meanwhile the PDE is also completely misidentified (see Case 3 in Table \ref{table:EnclosedAreaStudy}). Once we double the clean measurement points to $3\times10^3$, the trained DNN has strong interpolation and discovery abilities; however, the trained network does a poor job in extrapolating the full-field response (see Case 4 in Table \ref{table:EnclosedAreaStudy}). Concluding from this parametric test, we can see that the collocation points can render PiDL tolerable to scarce and noisy measurements, making the DNN generalizable.

\subsection{Simultaneous identification of unknown source term}
In practical applications, the physical system might be subjected to spatiotemporal source input ($\mathbf{p}$) which is unknown and can be only sparsely measured. When discovering the underlying governing equation for such a system, the source should be considered and reconstructed concurrently. In this case, we incorporate the source candidate functions into the library $\boldsymbol{\phi}$ for simultaneous discovery of the PDE and reconstruction of the unknown source. Thus, the sparse representation of the PDE(s) can be written as 
\begin{equation*}
    \mathbf{u}_t = [\boldsymbol{\phi}^u~\boldsymbol{\phi}^p][\boldsymbol{\Lambda}^u~\boldsymbol{\Lambda}^p]^T
\end{equation*}
where $\boldsymbol{\phi}^u$ and $\boldsymbol{\phi}^p$ denote the libraries of candidate functions, while $\boldsymbol{\Lambda}^u$ and $\boldsymbol{\Lambda}^p$ are the corresponding sparse coefficients, for the field variable $\mathbf{u}$ and the source $\mathbf{p}$, respectively. To demonstrate this concept, we test the Burgers' equation driven by a source term, expressed as
\begin{equation*} \label{eq:Burgers_source}
u_t + uu_x - 0.1 u_{xx} = \sin(x)\sin(t).
\end{equation*}
To generate the solution, the problem domain is meshed into 201 spatial grid points for $x\in[-5,5]$ and 101 time steps for $t\in[0, 10]$. We use 20 fixed sensors randomly selected from the spatial grid points to record the wave response ($u$) for 50 time steps, polluted with 10\% noise. Note that the source is not measured and regarded as unknown. 

\begin{figure}[t!]
\centering
\includegraphics[width = 0.75\textwidth]{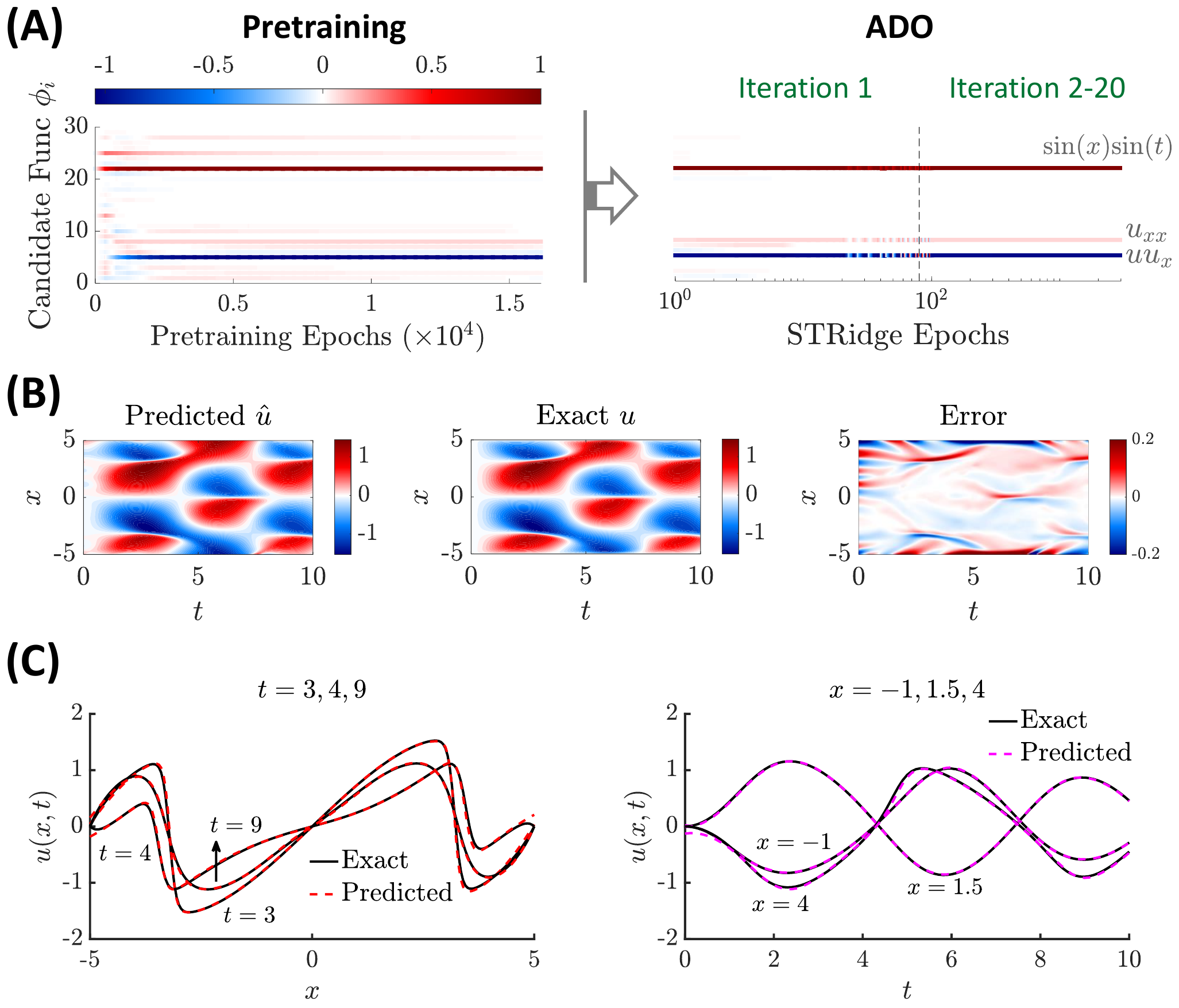}
\caption{Discovered Burgers' equation and source term for measurement data with 10\% noise. (\textit{A}) Evolution of the sparse coefficients $\boldsymbol{\Lambda}\in\mathbb{R}^{30\times1}$ for 30 candidate functions $[\boldsymbol{\phi}^u~\boldsymbol{\phi}^p]\in\mathbb{R}^{1\times30}$ used to form the PDE and the unknown source term, where the color represents the coefficient value. (\textit{B}) The predicted response in comparison with the exact solution with the prediction error. (\textit{C}) Comparison of spatial and temporal snapshots between the predicted and the exact solutions. The relative full-field $\ell_2$ error of the prediction is 13.8\%. The major errors are mostly distributed close to the boundaries due to scarce training data.}
\label{Sup_Fig:Source}
\end{figure}

The following libraries of candidate function are used to reconstruct the PDE and the source:
\begin{equation*}
\boldsymbol{\phi}^u = \{1, u, u^2, u^3, u_x, uu_x, u^2u_x, u^3u_x, u_{xx}, uu_{xx}, u^2u_{xx}, u^3u_{xx}, u_{xxx}, uu_{xxx}, u^2u_{xxx}, u^3u_{xxx}\}
\end{equation*}
\begin{equation*}
\boldsymbol{\phi}^p = \{a, b, c, d, a^2, b^2, c^2, d^2, ac, ab, ad, bc, bd, cd\}
\end{equation*}
where $a=\sin(t)$, $b=\sin(x)$, $c=\cos(t)$ and $d=\cos(x)$. The hyperparameters for the PiDL network are similar those used in the previous Burgers' example. The pre-training takes up to $15\times10^3$ epochs of Adam and about $1\times 10^3$ epochs of L-BFGS, followed by 20 ADO iterations. In each ADO iteration, we use the Adam optimizer with $1\times 10^3$ epochs and the L-BFGS with up to $1\times10^3$ epochs to train the DNN for each alternation within STRidge. The discovered PDE along with the uncovered source term is given by
\begin{equation*} \label{eq:Burgers_source_disc}
u_t + 1.002uu_x - 0.088u_{xx} = 0.995\sin(x)\sin(t).
\end{equation*}
It can be seen by comparing the above two equations that both the sparse terms and the corresponding coefficients are accurately identified, despite only scarce and noisy measurement of the system response is supplied. The discovery result is summarized in Fig. \ref{Sup_Fig:Source}. The evolution of the sparse coefficients for both the PDE and the source term shows robust and quick convergence to the ground truth (Fig. \ref{Sup_Fig:Source}\textit{A}), with the average relative error for non-zero coefficients of 4.39\% $\pm$ 7.03\%. Although only 4.9\% subsampled responses are measured while the source information is completely unknown, the PiDL approach can reasonably well extrapolate the full-field solution with a $\ell_2$ error of 13.8\% (see Fig. \ref{Sup_Fig:Source}\textit{B}). The major errors are mostly distributed close to the boundaries due to scarce training data. Fig. \ref{Sup_Fig:Source}\textit{C} shows the comparison of spatial and temporal snapshots between the predicted and the exact solutions which match well with each other. 

Nevertheless, if the source is very complex with its general expression or form completely unknown, distinct challenges arise when designing the source library of candidate functions $\boldsymbol{\phi}^p$. This may require an extraordinarily large-space library to retain diversifying representations, and thus pose additional computational complexity for accurate discovery of the PDEs. In some specific cases, the unknown source term can probably be approximated by the combination of continuous basis functions such the Fourier series, instead of finding its closed form. These open questions will be addressed in our future work.

\subsection{Other network architecture}
There still remain some potential limitations associated with the present PiDL framework for physical law discovery. For example, although the fully connected DNN used in this work has advantage of analytical approximation of the PDE derivatives via automatic differentiation, directly applying it to model the solution of higher dimensional systems (e.g., long-short term response evolution in a 3D domain) results in computational bottleneck and optimization challenges. Advances in discrete DNNs with spatiotemporal discretization (e.g., the convolutional long-short term memory network (ConvLSTM) \cite{Shi2015} or similar) have the potential to help resolve this challenge, which will be demonstrated in our future work. However, a careful design of the spatiotemporal differentiator is required for the discrete DNNs (e.g., high-order finite difference filter for derivative approximation, accounting for domain irregularity, etc.).

\vspace{12pt}

\bibliographystyle{unsrt}
\footnotesize
\bibliography{references_supp}